\documentclass{article}

\usepackage[preprint]{preprint}

\usepackage[utf8]{inputenc} 
\usepackage[T1]{fontenc}    
\usepackage{hyperref}       
\usepackage{url}            
\usepackage{booktabs}       
\usepackage{amsfonts}       
\usepackage{amsmath,amssymb,amsthm}
\usepackage{nicefrac}       
\usepackage[dvipsnames,table]{xcolor} 
\usepackage{listings}
\usepackage{graphicx}
\usepackage{subcaption}

\usepackage{booktabs,array,tabularx}
\usepackage{enumitem}

\ifdefined\DeclareUnicodeCharacter
\DeclareUnicodeCharacter{00B7}{\ensuremath{\cdot}}       
\DeclareUnicodeCharacter{00D7}{\ensuremath{\times}}      
\DeclareUnicodeCharacter{2190}{\ensuremath{\leftarrow}}  
\DeclareUnicodeCharacter{2192}{\ensuremath{\rightarrow}} 
\DeclareUnicodeCharacter{2264}{\ensuremath{\leq}}        
\DeclareUnicodeCharacter{2265}{\ensuremath{\geq}}        
\DeclareUnicodeCharacter{2260}{\ensuremath{\neq}}        
\DeclareUnicodeCharacter{2200}{\ensuremath{\forall}}     
\DeclareUnicodeCharacter{2014}{\textemdash}              
\DeclareUnicodeCharacter{27E8}{\ensuremath{\langle}}     
\DeclareUnicodeCharacter{27E9}{\ensuremath{\rangle}}     
\DeclareUnicodeCharacter{2227}{\ensuremath{\wedge}}      
\DeclareUnicodeCharacter{00AC}{\ensuremath{\neg}}        
\DeclareUnicodeCharacter{2203}{\ensuremath{\exists}}     
\DeclareUnicodeCharacter{2208}{\ensuremath{\in}}         
\DeclareUnicodeCharacter{22A2}{\ensuremath{\vdash}}      
\DeclareUnicodeCharacter{2194}{\ensuremath{\leftrightarrow}} 
\DeclareUnicodeCharacter{21A6}{\ensuremath{\mapsto}}     
\DeclareUnicodeCharacter{2115}{\ensuremath{\mathbb{N}}}  
\DeclareUnicodeCharacter{2124}{\ensuremath{\mathbb{Z}}}  
\DeclareUnicodeCharacter{03B1}{\ensuremath{\alpha}}      
\DeclareUnicodeCharacter{03B2}{\ensuremath{\beta}}       
\DeclareUnicodeCharacter{03C3}{\ensuremath{\sigma}}      
\DeclareUnicodeCharacter{03A3}{\ensuremath{\Sigma}}      
\DeclareUnicodeCharacter{2026}{\ensuremath{\ldots}}      
\DeclareUnicodeCharacter{2212}{\ensuremath{-}}           
\DeclareUnicodeCharacter{2080}{\ensuremath{{}_{0}}}      
\DeclareUnicodeCharacter{2081}{\ensuremath{{}_{1}}}      
\DeclareUnicodeCharacter{2082}{\ensuremath{{}_{2}}}      
\DeclareUnicodeCharacter{2098}{\ensuremath{{}_{m}}}      
\DeclareUnicodeCharacter{25B8}{\ensuremath{\blacktriangleright}} 
\DeclareUnicodeCharacter{0101}{\ensuremath{\bar{a}}}     
\DeclareUnicodeCharacter{2983}{\ensuremath{\{\!|}}       
\DeclareUnicodeCharacter{2984}{\ensuremath{|\!\}}}       
\fi

\newcommand{\li}[1]{\texttt{#1}}        
\newcommand{\aeneas}{\textsc{Aeneas}}   
\newcommand{\charon}{\textsc{Charon}}   
\newcommand{\fleet}{\textsc{InfiniLean}}   

\definecolor{keywordcolor}{HTML}{0033AA}
\definecolor{tacticcolor}{HTML}{7A3E9D}
\definecolor{sortcolor}{HTML}{0033AA}
\definecolor{attributecolor}{HTML}{7A3E9D}
\definecolor{symbolcolor}{HTML}{000000}
\definecolor{commentcolor}{HTML}{6A737D}

\lstdefinelanguage{Rust}{
  morekeywords={as,break,const,continue,crate,else,enum,false,fn,for,if,impl,
    in,let,loop,match,mod,move,mut,pub,ref,return,self,Self,static,struct,
    super,trait,true,type,unsafe,use,where,while,
    bool,char,i8,i16,i32,i64,i128,isize,u8,u16,u32,u64,u128,usize,f32,f64,
    str,String,Vec,Box,Option,Some,None},
  sensitive=true,
  morecomment=[l]{//},
  morecomment=[s]{/*}{*/},
  morestring=[b]",
  keywordstyle=\color{keywordcolor},
  commentstyle=\itshape\color{commentcolor},
}
\lstdefinestyle{code}{
  basicstyle=\ttfamily\scriptsize,
  commentstyle=\itshape\color{commentcolor},
  frame=single,
  framesep=2mm,
  breaklines=true,
  breakatwhitespace=false,
  tabsize=2,
}

\theoremstyle{plain}

\theoremstyle{definition}

\usepackage{mathtools}
\usepackage{wrapfig}

\hypersetup{
  colorlinks=true,
  linkcolor=BrickRed,
  citecolor=MidnightBlue,
  urlcolor=MidnightBlue,
}
\usepackage[capitalise, noabbrev, nameinlink]{cleveref}

\definecolor{oursrow}{HTML}{EAF1FB}            
\newcommand{\ours}{\rowcolor{oursrow}}
\newcommand{\best}[1]{\textbf{#1}}              
\newcommand{\second}[1]{\underline{#1}}         
\newcommand{\na}{\textcolor{black!30}{--}}      
\newcommand{\budget}[1]{\,{\scriptsize\textcolor{black!55}{#1}}}  

\newcommand{\sysname}{\textsc{Aegis}}
\newcommand{\harnessname}{\textsc{EasyProver}}

\title{Scaling Lean Theorem Proving for \\ Real-World Code Verification}

\author{%
  Debargha Ganguly$^{2}$\thanks{Corresponding author. Work done during an internship at Amazon Web Services.},
  Divya Raghunathan$^{1}$,
  Dejan Jovanovi\'c$^{1}$,
  Vikash Singh$^{2}$,\\
  \textbf{Soonho Kong}$^{1}$,
  \textbf{Aaron Tomb}$^{1}$,
  \textbf{Aman Goel}$^{1}$,
  \textbf{Ferhat Erata}$^{1}$,
  \textbf{Sam Bayless}$^{1}$\\
  $^{1}$Amazon Web Services
  $^{2}$Case Western Reserve University\\
  \texttt{debargha@case.edu}\\
  \texttt{divraghu@amazon.com}
}

\begin{document}

\maketitle

\begin{abstract}
We present \harnessname{}, a Lean proving system built from recursive agents with isolated contexts, asynchronous search, parallel tactic probing, and an elastic elaboration service. Using an off-the-shelf frontier model with no additional training, it proves $669/672$ PutnamBench problems and all six IMO~2026 formalizations and resolves $597/600$ NTP4VC verification conditions and all $189$ VERINA tasks by proof or refutation. Because these benchmarks are close to saturation, we also introduce \sysname{}, which converts competitive programming problems into much harder verification tasks over real implementations. The pipeline writes safe Rust solutions, translates them into Lean with \aeneas{}, generates specifications, and labels each frozen statement with a kernel-checked proof or refutation. It produces $2{,}045$ differentially validated Rust--Lean program pairs and $11{,}533$ candidate specifications. Unlike hand-written benchmarks, these tasks carry implementation-level obligations from fixed-width arithmetic, explicit failure, generated loops, casts, and functionalized mutation. On a stratified sample of $2{,}000$ statements, \harnessname{} proves $1{,}054$ ($52.7\%$) within a 30-minute budget per statement. The same tasks can also be used to benchmark other provers and as rewards for reinforcement learning.
\end{abstract}

\section{Introduction}
\label{sec:intro}

Large language models (LLMs) paired with proof assistants now produce machine-checked proofs of competition mathematics~\citep{seedprover2025,chung2026goedel} and of program correctness~\citep{verina2025,goedelcodeprover2026}. Building an effective agent for these tasks takes more than a frontier model and a proof checker. The agent must decompose goals, find or prove intermediate lemmas, and recover from failed approaches, all within a finite context window and compute budget. Verifier feedback is also local. Lean reports whether each step elaborates, but the cost of a poor decomposition or a missing lemma may only become visible many steps later. Mathematical proofs often depend on choosing the right abstraction and building auxiliary lemmas in layers. Program verification adds recursive program structure, data representations, invariants, and implementation-level proof obligations. In both settings, the agent must produce correct individual steps and also keep a long reasoning process coherent, revising it when an approach fails.

Evaluating such agents requires tasks they cannot already solve. Several established formal-reasoning benchmarks are now close to saturation (\cref{sec:results}), and existing verified-programming benchmarks are small. Each verification task needs a formal statement of the intended behavior, a model of the program, and meaningful proof obligations. Writing these requires expertise in programming, formal specification, and theorem proving, so these benchmarks contain only hundreds of tasks~\citep{verina2025,ntp4vc2026,loughridge2024dafnybench}. Small static benchmarks are also exposed to memorization and contamination~\citep{carlini2023quantifying,dong2024contamination}. Generating verification tasks automatically addresses both problems, provided the specifications capture the intended behavior and the proofs span a range of difficulty. Such tasks have a second use. Tests give only finite evidence of correctness~\citep{chen2021humaneval,jimenez2024swebench,liu2023evalplus}, and when used as training rewards they can encourage agents to exploit the test harness instead of implementing the intended behavior~\citep{baker2025monitoring,gamingverifiers2026}. A proof instead certifies correctness against a formal specification and is cheap to check~\citep{comparator}, so kernel-checked labels can also serve as rewards for reinforcement learning from verifiable rewards (RLVR)~\citep{lambert2024tulu3}.

We make two contributions.
\begin{enumerate}[leftmargin=*, itemsep=1pt, topsep=0pt]
\item \textbf{A proving system that nearly saturates current benchmarks.}
\harnessname{} combines recursive agents with isolated contexts, asynchronous search, parallel tactic probing, and \fleet{}, a stateless, elastic Lean elaboration service (\cref{sec:prover}). With an off-the-shelf frontier model and no additional training, it proves $669/672$ PutnamBench problems and all six IMO~2026 formalizations and resolves $597/600$ NTP4VC verification conditions and all $189$ VERINA tasks by proof or refutation (\cref{sec:results}). Every reported result is a kernel-checked proof or refutation.

\item \textbf{A much harder test set for verification.}
\sysname{} turns competitive programming problems into Lean verification tasks over translated implementations. It synthesizes safe Rust, translates it into Lean with \aeneas{}, generates specifications, and labels each frozen statement with a kernel-checked proof or refutation (\cref{sec:env}). The pipeline yields $2{,}045$ validated Rust--Lean programs and $11{,}533$ candidate specifications. \harnessname{} proves $1{,}054$ statements in a $2{,}000$-statement sample, so these tasks remain far from saturated for a prover that nearly saturates the established benchmarks. Because the pipeline needs no expert annotation, more tasks can be generated with additional compute, and the same tasks can be used to benchmark other provers or as RLVR rewards.
\end{enumerate}

\vspace{-1em}
\section{Related Work}
\label{sec:related}

\paragraph{Neural theorem proving and proof search.}
Most LLM-based provers are trained on curated proof corpora and guide search with a learned policy or value function. They generate either one proof step at a time~\citep{polu2020gpt} or whole proofs~\citep{first2023baldur}. Stepwise search typically relies on premise selection~\citep{leandojo2023} and on best-first procedures such as A* and MCTS~\citep{lample2022hypertree}. Goedel-Prover-V2~\citep{goedelproverv2}, DeepSeek-Prover-V2~\citep{ren2025deepseek}, and Seed-Prover~\citep{seedprover2025} improve accuracy through expert iteration~\citep{polu2022curriculum}, reinforcement learning, and large inference budgets. For code verification, Goedel-Code-Prover~\citep{goedelcodeprover2026} trains a policy to reduce verification goals to kernel-checked lemmas. In contrast, we do not train or fine-tune the model. Our search is loosely analogous to Monte Carlo tree search~\citep{kocsis2006bandit,silver2016alphago}. In UCT, a statistical rule selects which node to expand, and node values are updated by averaging the returns of random rollouts. In \harnessname{}, the parent agent is prompted with the proof state and the results its children have returned, and decides which branches to continue, revise, or expand. It reads each child's result directly, such as a completed proof or a counterexample, instead of maintaining averaged value estimates. Each agent runs a ReAct-style loop that alternates model calls with tool calls~\citep{yao2023react,wang2023voyager}, and the Lean tools are exposed through the Model Context Protocol~\citep{anthropic2024mcp}. Existing Lean interfaces, from LeanDojo~\citep{leandojo2023} to \li{lean-lsp-mcp}~\citep{lean-lsp-mcp}, run Lean on the same machine as the agent and refer to proofs by local file paths, so each prover is tied to a single Lean server; prompt libraries such as \li{lean4-skills}~\citep{lean4-skills} package Lean expertise for such agents. \fleet{} is instead a stateless service that accepts Lean source over the network, so any replica in a load-balanced pool can check any proof (\cref{sec:prover}).

\paragraph{Data, benchmarks, and contamination.}
Although \harnessname{} itself is not trained, the tasks that \sysname{} generates bear on the data from which neural provers learn. Neural provers are trained mostly on $(\text{proof-state}, \text{tactic})$ pairs mined from proof-assistant interaction~\citep{leandojo2023}. Several systems use a single model to both propose and solve tasks, including self-play provers~\citep{dong2025stp}, intrinsically motivated conjecturers~\citep{poesia2024minimo}, and reasoners whose tasks are graded by code execution~\citep{zhao2025absolutezero}. \sysname{} separates task generation from solving and labels each frozen statement with a kernel-checked proof of the statement or of its negation. Refutations provide negative labels, which proof-only and execution-graded loops do not produce. Once the statement and trust policy are fixed, anyone can recheck a label, which makes it suitable as an RLVR reward~\citep{lambert2024tulu3}. Existing benchmarks cover olympiad mathematics, such as miniF2F~\citep{zheng2021minif2f} and PutnamBench~\citep{putnambench2024}, and verified code, such as VERINA~\citep{verina2025}, CLEVER~\citep{clever2025}, AlgoVeri~\citep{zhao2026algoveri}, and the industrial NTP4VC suite~\citep{ntp4vc2026}. These benchmarks are static, which exposes them to contamination, and several are close to saturation; recent systems report $100\%$ on miniF2F~\citep{mistralLeanstral15}. Existing mitigations, such as splits with novel premises~\citep{leandojo2023}, $n$-gram overlap audits~\citep{verina2025}, and release-date cutoffs~\citep{livecodebench2024}, detect or limit contamination within a fixed task set but do not produce new tasks. \sysname{} instead generates tasks automatically. Its source problems are public, but the translated programs and their specifications are new, and more tasks can be generated with additional compute. The $52.7\%$ proof rate in \cref{sec:results} shows that these tasks are far from saturated. The pipeline also supports mutation testing~\citep{jia2011mutation}: small edits to a validated program produce plausible but incorrect variants, whose specifications yield further negative labels. We used mutation to generate additional tasks but do not evaluate them in this paper.

\paragraph{Autoformalization and Rust-to-Lean verification.}
\aeneas{}~\citep{aeneas2022}, built on the \charon{} LLBC representation~\citep{charon2024}, translates safe Rust into a functional Lean model in which faults and nontermination are explicit values. Leanstral combines this kind of translation with proving and attempts to prove the negation of properties it fails to prove~\citep{mistralLeanstral15}. DeepSeek-Prover filters statements by attempting disproofs in parallel~\citep{deepseekproverv1}. A zkVM pipeline discharges \aeneas{}-extracted obligations under a kernel check~\citep{klaus2026rustlean}. Autoformalization maps informal mathematical statements to formal ones~\citep{jiang2022draft,zhou2024lego}. The specification agent in \sysname{} performs a form of autoformalization: it turns a natural-language programming problem into Lean statements about the extracted program. We use \aeneas{} to translate Rust into Lean, and every statement refers to the extracted Lean program itself. Each label is therefore correct for that program even if the translation were imperfect; differential testing (\cref{sec:env}) separately provides evidence that the Lean program behaves like the Rust source. Agents that can edit proof files may game the checker instead of solving the task, for example by weakening the theorem statement, leaving a \li{sorry}, or introducing an axiom. This is a form of reward hacking~\citep{baker2025monitoring,gamingverifiers2026}. \harnessname{} blocks each of these: the statement prefix is frozen, and Lean Comparator requires the final file to compile without \li{sorry} and to use only allowlisted axioms (\cref{sec:prover}). In \sysname{}, a separate agent writes each statement before the prover sees it, so the prover cannot choose an easier statement to prove.

\section{Methodology}

\subsection{\harnessname{}: A Task-Parallel Proving Runtime}
\label{sec:prover}

\begin{figure}[!htbp]
    \centering
    \includegraphics[width=\linewidth]{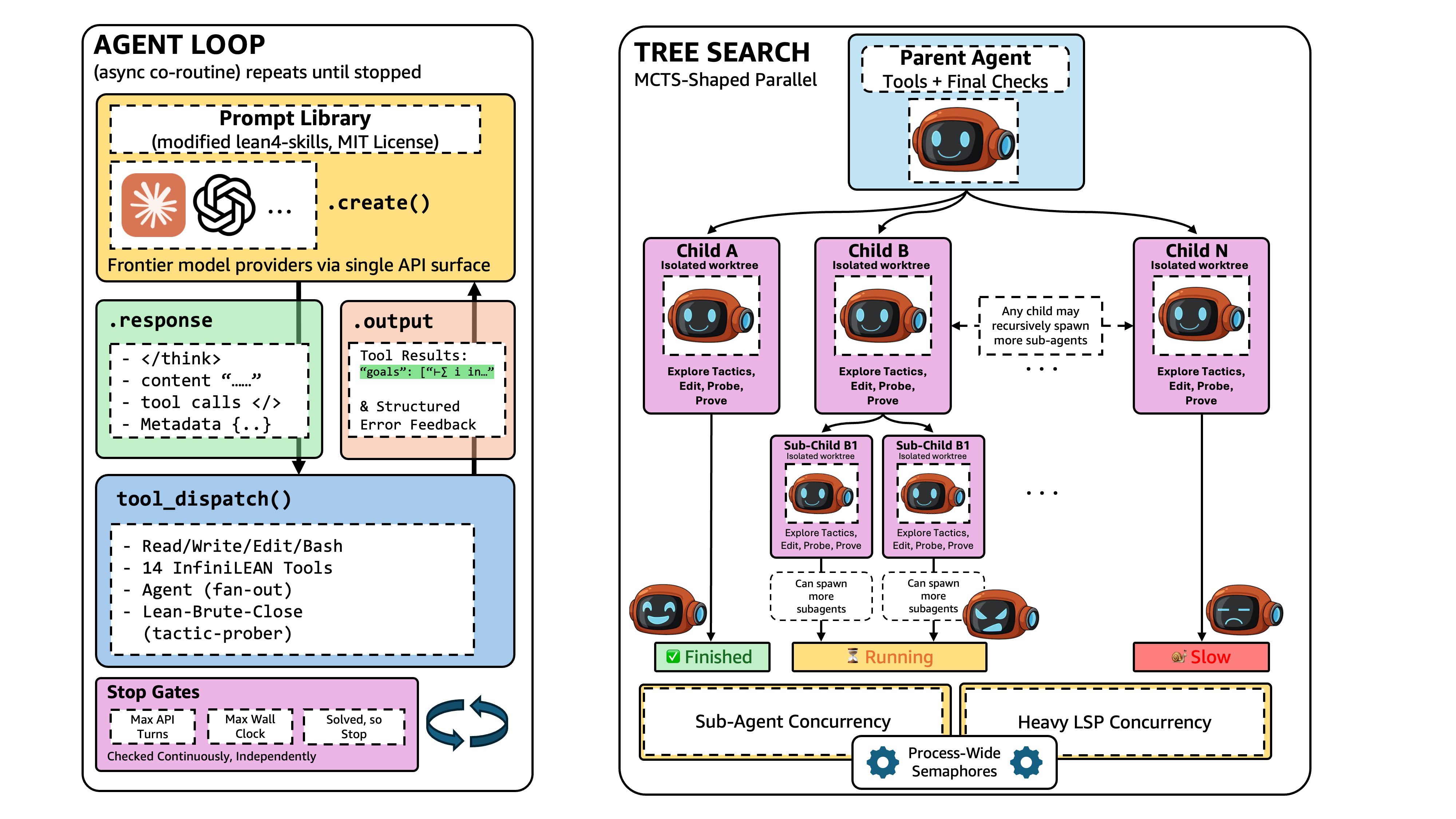}
    \caption{\textbf{\harnessname{} architecture.}
    \textbf{Left:} An asynchronous agent loop alternates model
    responses with coding and Lean tool calls.
    \textbf{Right:} Agents recursively delegate work to children
    with separate contexts and workspaces. Completed results return
    to the parent without waiting for unfinished siblings, which keep
    running; the search ends when the first candidate proof is accepted.}
    \label{fig:architecture}
    \vspace{-1em}
\end{figure}

\begin{figure}[!h]
    \centering
    \includegraphics[width=1.0\linewidth]{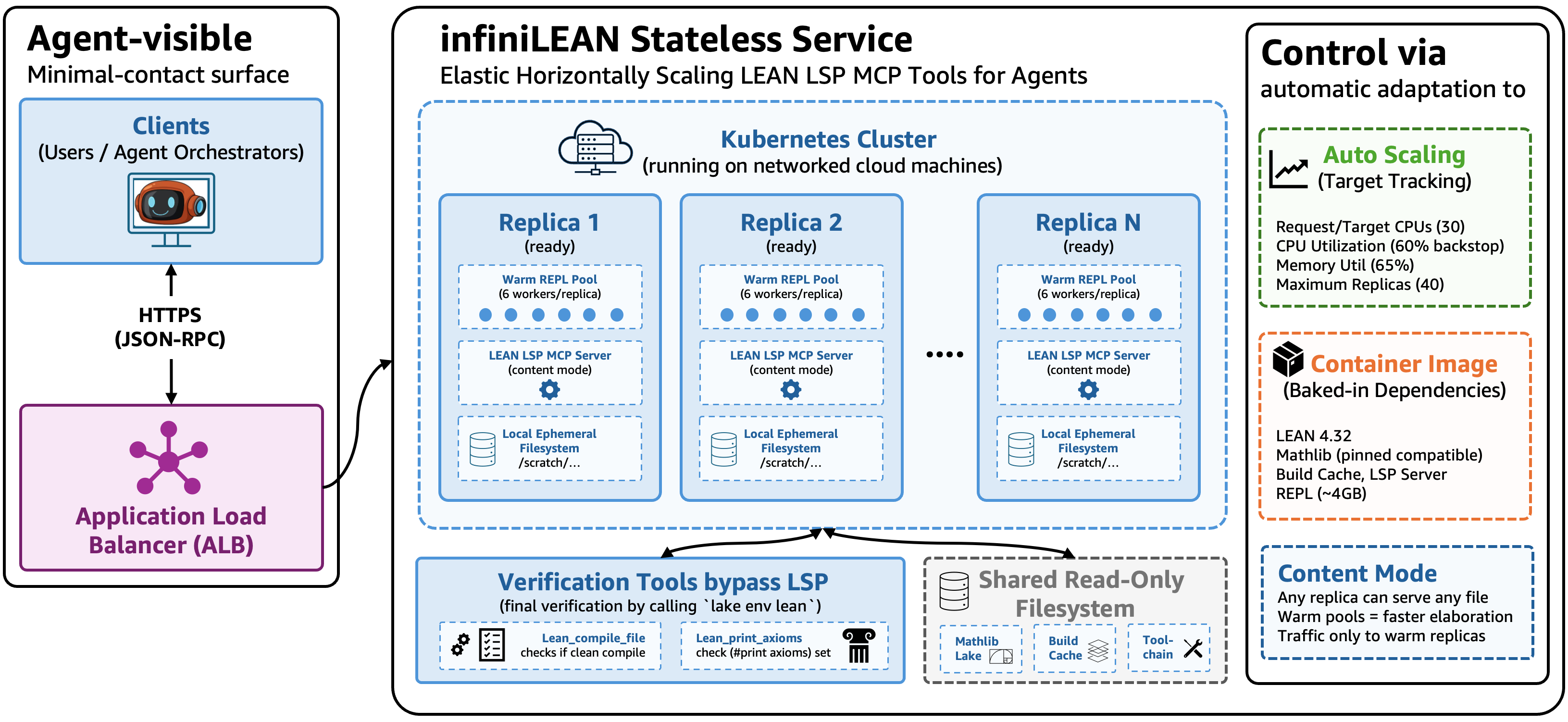}
    \caption{\textbf{\fleet{} architecture.}
    \fleet{} is a stateless service: Lean source requests are load-balanced across Kubernetes
    replicas with ephemeral workspaces and Mathlib-preloaded
    workers. Replicas serve interactive elaboration during search;
    final acceptance is checked separately by Lean Comparator.}
    \label{fig:infinilean}
    \vspace{-1em}

\end{figure}

\paragraph{Architecture.}
\harnessname{} separates proof search (\cref{fig:architecture}), interactive Lean elaboration (\cref{fig:infinilean}), and final validation of the proof. Agents explore proof strategies through recursive delegation and parallel tactic probing, while \fleet{} serves elaboration requests independently of agent execution.

\paragraph{Problem formulation and acceptance.}
Each task fixes a Lean~4 environment $\Gamma$ and a proposition $P$~\citep{demoura2021lean4}. For both mathematical and program-verification tasks, the goal is a kernel-checked proof of either $P$ or $\neg P$. The file prefix up to and including the target theorem signature is frozen, so only the proof body can change. A completed proof is accepted only if it passes Lean Comparator~\citep{comparator}, which runs \li{lake env lean} and \li{\#print axioms} on the generated file with a shared, read-only toolchain, independently of the language server used during search. Neither an agent's report of success nor a successful local tactic probe counts as acceptance. Anyone with the same environment and axiom policy can recheck an accepted proof; its guarantee rests on the frozen statement, the permitted axioms, and Lean's kernel.

\paragraph{Asynchronous agent runtime.}
The harness uses the off-the-shelf Anthropic Fable-5 model with no fine-tuning, no separately trained policy or value network, and no additional proprietary proof corpus. At each turn, the model receives its system prompt, the conversation history, and the available tools. The runtime dispatches the requested tool calls, appends their results, and continues. The tools cover file and sandboxed shell operations, inspection of Lean proof states and diagnostics, premise retrieval, parallel tactic probing, and recursive delegation through \li{Agent}. Parent and child agents run the same loop. Limits on API turns and a wall-clock deadline bound execution, and prompt caching and compact tool responses reduce token and communication costs.

\paragraph{Recursive, model-directed search.}
\label{sec:search}
Search grows during execution instead of running a fixed batch of independent restarts. A parent launches children to pursue alternative proof strategies or auxiliary lemmas. Each child receives a fresh context, a private copy of the project, and the same tools as its parent, including \li{Agent}, so delegation can recurse~\citep{zhang2025recursive}. Children return completed proofs, counterexamples, or partial findings, and the parent uses them to decide which approaches to continue, revise, or expand. The structure resembles Monte Carlo tree search, but the language model makes these decisions from the content of the returned results, not from visit counts or rollout-based value estimates.

\begin{wrapfigure}{r}{0.52\textwidth}
    \centering
    \vspace{-2em}
    \includegraphics[width=0.48\textwidth]{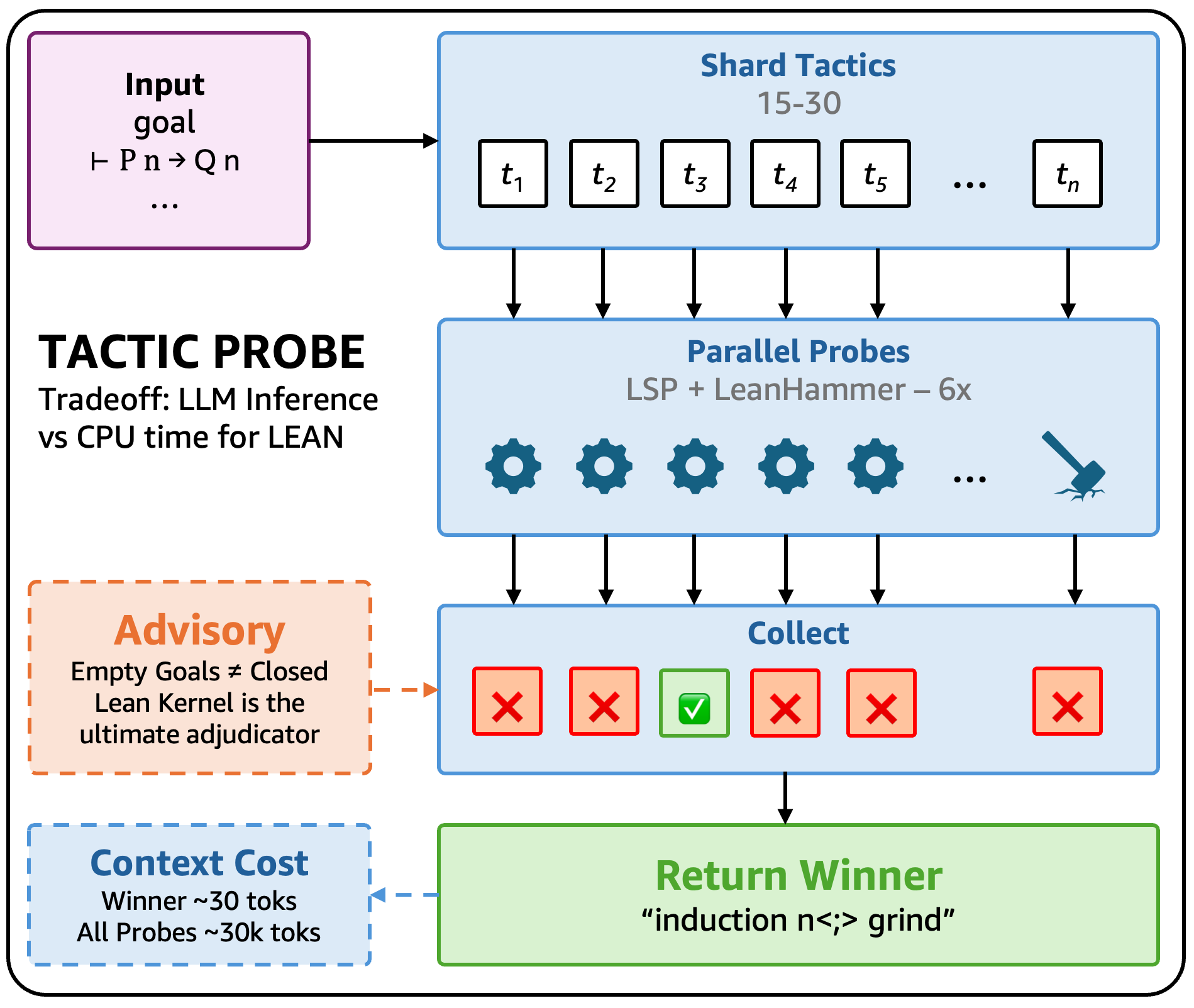}
    \caption{\textbf{Parallel tactic probing.}
    Candidate tactics are evaluated concurrently, and a successful
    tactic is returned to the agent. In this example, the returned
    tactic is about $30$ tokens, while the probe traces total about
    $30{,}000$ tokens.}
    \vspace{-3em}
    \label{fig:tactic-probe}
\end{wrapfigure}

\paragraph{Context isolation and asynchronous returns.}
A child may spend dozens of tool calls deriving a helper lemma and return only the lemma. The parent can build on it without importing the failed attempts and diagnostics behind it. Context isolation thus decouples the total exploration across the search tree from the history that any single agent must keep. Completed results reach the parent at its next turn, without waiting for unfinished siblings, so an early finding can redirect search while other branches are still running.

\paragraph{Parallel tactic probing.}
Within a proof attempt, the tactic prober runs several candidate tactics and automated reasoning tools~\citep{zhu2025leanhammer} on the current proof state concurrently, through Lean language-server and automation workers (\cref{fig:tactic-probe}). When a candidate succeeds, the prober returns that tactic and discards the diagnostics of the failed probes. An agent can therefore try many local continuations without adding every probe trace to its context.

\paragraph{Resource admission control.}
Recursive delegation can create more concurrent sub-agents and Lean queries than a harness process can run at once. Two process-wide semaphores cap active sub-agents at $64$ and concurrent expensive Lean queries at $16$, so each harness process limits agent execution and Lean load independently.

\paragraph{Decoupled Lean elaboration.}
The hardest proofs used more than $300$ sub-agents in a single run. A single host lacks the Lean capacity for search this wide, so \fleet{} serves interactive elaboration as a stateless service backed by a load-balanced pool of replicas (\cref{fig:infinilean}). Clients send Lean source over JSON-RPC instead of server-local file paths, so any ready replica running the pinned Lean and Mathlib versions can serve any request. Each replica combines a Lean language-server MCP interface, an ephemeral workspace, and a pool of REPL workers with Mathlib preloaded. Reusing workers amortizes import costs. In our setup, imports take tens of seconds, while elaboration on a warm worker takes about $0.2\,\mathrm{s}$. The replicas run on Kubernetes, and target-tracking autoscaling adjusts their number to demand.

\paragraph{Candidate promotion and termination.}
Candidate proofs are promoted from branch-local workspaces to the solution file under a lock. The harness rechecks the promoted file against the build and the acceptance criteria and reverts the change if the check fails. The first accepted candidate ends the search.

\subsection{\sysname{}: The Data Environment}
\label{sec:env}

\begin{figure}
    \centering
    \includegraphics[width=1\linewidth]{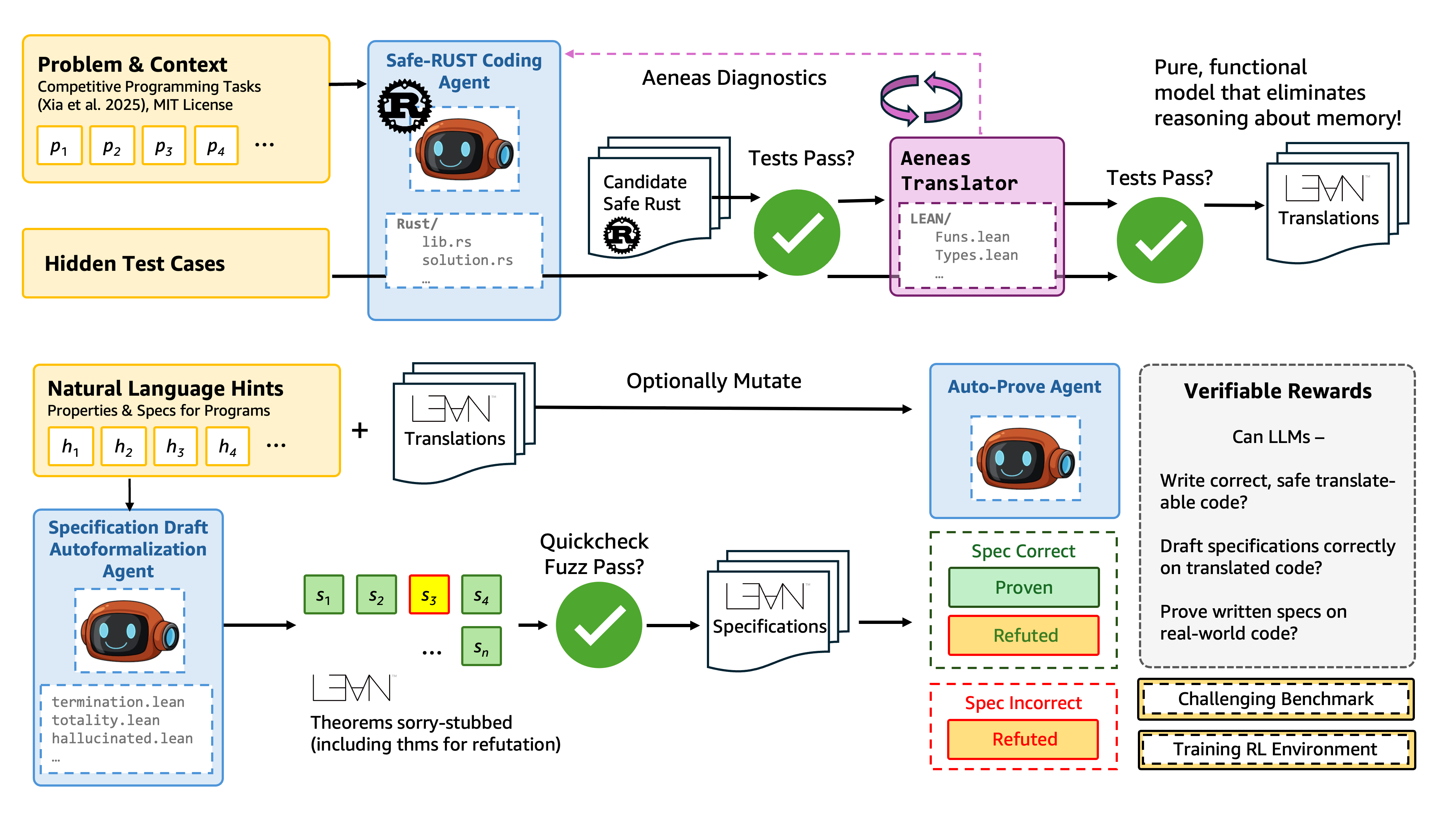}
    \caption{\textbf{The \sysname{} pipeline.}
    The coding agent writes safe Rust that passes the retained
    reference tests, translates through \aeneas{}, and produces
    matching Lean outputs. The specification agent writes a property
    requested to hold or to fail, and a separate proving agent
    proves or refutes the frozen statement. The resulting tasks serve
    as a benchmark; used as an RL environment, checked outcomes reward
    the prover for resolving the task and the specification agent
    for matching the requested polarity. An optional counterexample
    screen (Plausible) runs before proof search.}
    \label{fig:data-pipeline}
\end{figure}

\Cref{sec:results} shows that \harnessname{} nearly saturates several established benchmarks. To evaluate it on tasks it does not already solve, we introduce \sysname{}, which generates much harder Lean verification tasks from competitive programming problems (\cref{fig:data-pipeline}). In this paper, we use \sysname{} as a test set for \harnessname{}. Because every label is a kernel-checked proof or refutation, the same tasks can also benchmark other provers and serve as an RLVR environment, in which implementation checks reward the coding agent and kernel-checked outcomes reward the specification and proving agents. We describe these uses below but do not evaluate them here.

\vspace{-1em}

\paragraph{Building the implementation corpus.}
The coding agent writes safe Rust solutions to LeetCode problems~\citep{xia2025leetcodedataset}. We keep problems whose interface is a single pure function over first-order data, including integers, booleans, lists, nested lists, strings, trees, and linked lists. Each data representation must pass through the whole pipeline: Rust testing, extraction by \charon{}, translation by \aeneas{}~\citep{aeneas2022}, Lean evaluation, and consistent rendering of values in both languages. This excludes stateful or class-based interfaces and features the pipeline does not support, such as hash tables, closures, nested references, and unrestricted aliasing.

\textbf{Data structures.}
LeetCode inputs are represented in forms that both Rust and Lean support. Strings become ASCII byte arrays, matrices are vectors of owned rows, and graphs keep their edge-list inputs. Trees and linked lists use owned recursive enums whose \li{Box} wrappers disappear under translation, yielding Lean inductive types. Each structured input is parsed once and rendered as matching Rust and Lean literals. Integers, sequence lengths, and test-set sizes are bounded to keep Lean evaluation tractable. A small subset of floating-point problems uses tolerance-based output comparison.

\textbf{Differential validation.}
For each implementation, \li{cargo test} runs every retained reference case. \charon{} extracts the Rust program, \aeneas{} translates it into an executable Lean model, and Lean evaluates the model on the same inputs. An implementation is accepted only if all reference tests pass and every Rust output matches the corresponding Lean output. Of the $2{,}356$ problems attempted, $2{,}045$ ($86.8\%$) passed every stage: Rust testing, extraction, translation, Lean elaboration, and differential evaluation. These validated programs, optionally augmented with mutated variants, are the program instances for specification generation.

\paragraph{Proof obligations in translated programs.}
The proof targets are translated implementations, not hand-written mathematical functions. Operations that may overflow, index out of bounds, or divide by zero return \li{Result $\alpha$} and are sequenced monadically. Even when a failure is unreachable, proving totality requires showing that the failing branch cannot occur. Fixed-width values such as \li{Std.U32} and \li{Std.I64} expose a mathematical projection \li{.val}, and arithmetic and width casts introduce explicit range obligations. A proof of totality for the machine-word program can therefore require bounds that do not affect the mathematical answer.

\aeneas{} also makes control flow and mutation explicit. Each Rust loop becomes a body that returns \li{ControlFlow}, a driver built from the loop combinator, and a tuple that carries the mutable loop variables. Nested loops generate further components, which proofs can handle bottom-up with helper lemmas. Read-only inputs use \li{Slice}, and mutable working storage uses \li{alloc.vec.Vec}. An \li{index\_mut} operation returns the current value together with a backward function that rebuilds the updated vector, so proofs must reason about list updates, length preservation, and invariants over whole collections. Bitwise operations and machine-word constants add further obligations of the same kind.

\paragraph{Generating specifications.}
For each program instance, a specification agent receives the original problem, the complete extracted Lean program, its signature, and several differentially validated examples. Using Lean's language-server tools, it writes an elaborating theorem statement about the program with a \li{sorry} proof body, together with a natural-language description. We request properties intended to hold, including functional correctness, totality, output bounds and shape, edge cases, and relational behavior, as well as properties intended to fail, such as claims with missing preconditions or off-by-one bounds. Statements are filtered for duplicates, trivial or vacuous claims, and elaboration errors. Each surviving statement is frozen and passed to the proving agent.

\paragraph{Proof outcomes and rewards.}
\harnessname{} attempts to resolve the frozen proposition $P$ by proving either $P$ or $\neg P$ without changing the statement. The task counts as resolved if the proof or refutation passes the acceptance check of \cref{sec:prover} under the task's declared trust policy, regardless of whether the specification was requested to hold or to fail. The same outcome also checks the specification: a proof confirms a specification requested to hold, a refutation confirms one requested to fail, and the opposite outcome marks a mismatch with the requested polarity. An unresolved task gives no signal. When \sysname{} is used for RLVR, these outcomes serve directly as rewards for the prover and the specification agent.

\paragraph{Optional counterexample screening.}
Before proof search, specifications requested to hold can be screened with Plausible~\citep{plausible}, a property-based testing tool for Lean, to catch drafting errors and save prover compute. Plausible samples concrete inputs for decidable propositions, and we add \li{Decidable} instances for common forms that \aeneas{} produces. A counterexample enters the same proof-or-refutation pipeline and counts as a refutation only after Lean checks it. Specifications for which no counterexample is found, or which cannot be tested, go to proof search. The screen produces no positive labels.

\section{Results}
\label{sec:results}

\paragraph{Reporting protocol.}
We count a proof only if the target statement is unchanged, the file compiles without \li{sorry}, and the target's axiom closure is within the benchmark's allowlist. A kernel-checked proof of the negation counts as a refutation.

\paragraph{Formal mathematics.}
\Cref{tab:ep_reported_math} compares \harnessname{} with published results, including those collected by \citet{chung2026goedel}. \harnessname{} solves $669$ of $672$ PutnamBench problems ($99.6\%$), $72$ more than the strongest configuration reported by \citet{chung2026goedel}, which uses natural-language guidance ($597/672$). It also solves all six IMO~2026 problems and all twelve Putnam~2025 problems. These results use an off-the-shelf model with no fine-tuning and no separately trained policy or value network. We did not evaluate \harnessname{} on IMO~2025 or USAMO~2026; those columns report other systems' published results. The model's training cutoff is not public, so we cannot rule out that informal solutions to some of these competition problems appear in its training data. Such contamination would affect how the results are interpreted but not their correctness, since every accepted proof is kernel-checked. The \sysname{} statements, by contrast, were generated by our pipeline and did not exist before this work.

\begin{figure}[!htbp]
    \centering
    \includegraphics[width=1\linewidth]{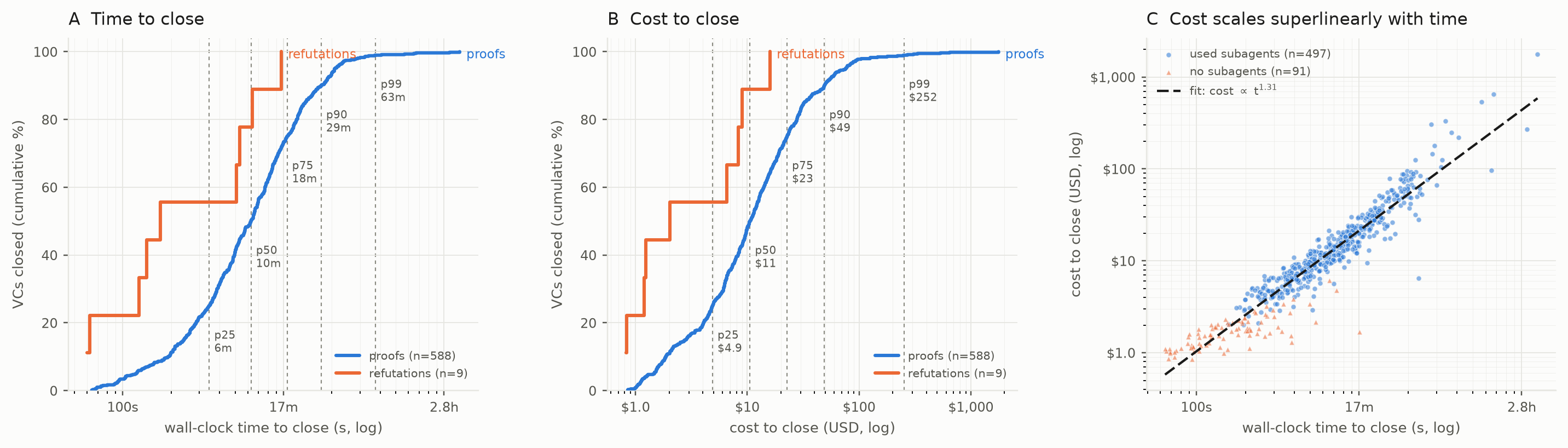}
    \caption{\textbf{NTP4VC time and cost.} Left/middle: cumulative time and cost distributions, normalized separately over 588 proofs and nine refutations. Right: proof cost versus time, grouped by subagent use. Unresolved tasks are excluded.}
    \label{fig:ntp4vc-cost}
    \vspace{-1em}
\end{figure}

\paragraph{Code verification.}
\Cref{tab:ep_reported_code} compares \harnessname{} with published code-verification results: the proof-generation baselines evaluated by \citet{goedelcodeprover2026}, the Lean results reported by \citet{ntp4vc2026}, and the system reports of \citet{harmonicverina2025} and \citet{alephverina2026}. \harnessname{} resolves all $189$ VERINA tasks, all $77$ AlgoVeri tasks, and $597$ of $600$ NTP4VC tasks ($99.5\%$), with $588$ proofs and $9$ refutations on NTP4VC. On CLEVER, it resolves $113$ of the $143$ tasks that translate to our Lean~4.32 toolchain ($79.0\%$). \Cref{fig:ntp4vc-cost} shows the time and cost of each resolved NTP4VC task.

\begin{table}[t]
\centering
\caption{Formal-mathematics results by reporting study; miniF2F and PutnamBench in \%. Small grey numbers give the attempt budget (@$k$) or, in brackets, the number of PutnamBench problems in that source's evaluation. Bold: best; underlined: second best.}
\label{tab:ep_reported_math}
\footnotesize
\setlength{\tabcolsep}{3.7pt}
\setlength{\aboverulesep}{0pt}\setlength{\belowrulesep}{0pt}\setlength{\extrarowheight}{0.6ex}
\begin{tabular}{lcccccc}
\toprule
Source & miniF2F\,$\uparrow$ & PutnamBench\,$\uparrow$ & IMO '25\,$\uparrow$ & Putnam '25\,$\uparrow$ & USAMO '26\,$\uparrow$ & IMO '26\,$\uparrow$ \\
\midrule
\citet{goedelproverv2}                  & 92.6\budget{@1024} & 13.0\budget{@184} & \na & \na & \na & \na \\
\citet{longcat2026}                     & 97.1\budget{@72}   & 41.5\budget{@118} & \na & \na & \na & \na \\
\citet{seedprover2025} (initial)        & \second{99.6}      & 50.4\budget{[657]} & \best{5/6} & \na & \na & \na \\
\citet{seedprover152025} (follow-up)    & \na                & 87.9\budget{[660]} & \best{5/6} & \second{11/12} & \na & \na \\
\citet{hilbert2025}                     & 99.2               & 70.0\budget{[660]\,@1840} & \na & \na & \na & \na \\
\citet{requena2026minimalagent}         & \na                & 54.7\budget{@1} & \na & \na & \na & \na \\
\citet{numinaleanagent2026}             & \na                & \na & \na & \best{12/12} & \na & \na \\
\citet{chung2026goedel} (+NL)           & \best{100.0}       & 88.8\budget{@4} & \second{4/6} & \second{11/12} & 3/6 & \na \\
\midrule
\citet{putnambenchleaderboard2026}$^{\ddagger}$ & \na & \best{100.0} & \na & \na & \na & \na \\
\citet{zhang2026humanizeolympiadagentsmultisession} & \na & \best{100.0} & \na & \na & \na & \best{6/6} \\
\citet{axiomimo2026}                    & \na & \na & \na & \na & \na & \best{6/6} \\
\midrule
\ours \harnessname{} (one run)          & \second{99.6} & \second{99.6} & \na & \best{12/12} & \na & \best{6/6} \\
\bottomrule
\end{tabular}
\par\vspace{1pt}
\parbox{\linewidth}{\scriptsize $^{\ddagger}$Live leaderboard; the same score is reported on both the with-answer and no-answer tracks.}
\end{table}

\begin{table}[t]
\centering
\caption{Code-verification results (\%) by reporting study. Rows report the share of tasks proved (kernel-checked proof of the original goal) unless the setting is marked resolved (kernel-checked proof of the goal or its negation). Ranges span the systems evaluated in each study. Bold: best; underlined: second best.}
\label{tab:ep_reported_code}
\footnotesize
\setlength{\tabcolsep}{4.5pt}
\setlength{\aboverulesep}{0pt}\setlength{\belowrulesep}{0pt}\setlength{\extrarowheight}{0.6ex}
\begin{tabular}{llccccc}
\toprule
Source & Setting & pass@$k$ & \begin{tabular}[b]{@{}c@{}}VERINA\,$\uparrow$\\[-1pt]\budget{(189)}\end{tabular} & \begin{tabular}[b]{@{}c@{}}CLEVER\,$\uparrow$\\[-1pt]\budget{(161)}\end{tabular} & \begin{tabular}[b]{@{}c@{}}AlgoVeri\,$\uparrow$\\[-1pt]\budget{(77)}\end{tabular} & \begin{tabular}[b]{@{}c@{}}NTP4VC\,$\uparrow$\\[-1pt]\budget{(600)}\end{tabular} \\
\midrule
\citet{verina2025}           & best reported               & 1   & 4.9 & \na & \na & \na \\
\citet{goedelcodeprover2026} & general LLMs                & 128 & 10.6--19.6 & 11.2--23.6 & 5.2--18.2 & \na \\
\citet{goedelcodeprover2026} & neural provers              & 128 & 15.9--24.3 & 23.0--25.5 & 19.5--23.4 & \na \\
\citet{goedelcodeprover2026} & hierarchical search         & 32  & 68.8 & \second{54.0} & \second{62.3} & \na \\
\citet{ntp4vc2026}           & general LLMs                & 8   & \na & \na & \na & 0.00--1.67 \\
\citet{ntp4vc2026}           & general LLM                 & 1   & \na & \na & \na & 0.50 \\
\citet{ntp4vc2026}           & specialized provers         & 8   & \na & \na & \na & \second{2.17--3.00} \\
\citet{harmonicverina2025}   & proved\budget{160/189}       & \na & 84.7 & \na & \na & \na \\
\citet{harmonicverina2025}   & resolved\budget{183/189}     & \na & \second{96.8} & \na & \na & \na \\
\citet{alephverina2026}      & reported score              & \na & \best{100.0} & \na & \na & \na \\
\midrule
\ours \harnessname{}         & resolved                    & 1   & \best{100.0} & \best{79.0}$^{\dagger}$ & \best{100.0} & \best{99.5} \\
\bottomrule
\end{tabular}
\par\vspace{1pt}
\parbox{\linewidth}{\scriptsize $^{\dagger}$113 of the 143 CLEVER tasks that translate to Lean~4.32.}
\end{table}

\paragraph{\sysname{}.}
From the $11{,}533$ candidate specifications, we drew a sample of $2{,}000$ statements over $1{,}257$ translated programs, stratified by LeetCode difficulty and specification category. The sample excludes statements already refuted by the Plausible screen and the category requested to fail. With a 30-minute budget per statement, \harnessname{} attempted $1{,}997$ statements. It proved $1{,}054$ statements ($52.7\%$ of the sample) and left $943$ unresolved. The proof rate ranged from $84.1\%$ for edge-case specifications to $31.0\%$ for explicit contracts, and from $66.7\%$ on Easy problems to $37.6\%$ on Hard ones. The same prover that nearly saturates the benchmarks above leaves almost half of these statements unresolved.

We do not ablate the individual components of the harness, because each configuration requires a full run over the benchmarks. In particular, we do not compare against the same model using Lean as a tool without the harness. Evaluating other provers on \sysname{}, and using it for reinforcement learning and multi-teacher on-policy distillation, is left to future work.

\vspace{-1em}
\section{Conclusion}
\label{sec:conclusion}
\vspace{-1em}
\harnessname{} shows that an agentic harness around an off-the-shelf frontier model, with no additional training, can resolve nearly all tasks in several established mathematics and verification benchmarks. \sysname{} provides much harder tasks that these benchmarks lack: specifications over mechanically translated safe Rust, with obligations from machine arithmetic, explicit failure, generated loops, casts, and mutation. The same prover proves $52.7\%$ of a sampled set of these statements, so the tasks remain challenging for a prover that nearly saturates current benchmarks. Their kernel-checked labels also make them suitable for benchmarking other provers and as rewards for reinforcement learning.

\bibliography{ref}

\begin{thebibliography}{60}
\providecommand{\natexlab}[1]{#1}
\providecommand{\url}[1]{\texttt{#1}}
\expandafter\ifx\csname urlstyle\endcsname\relax
  \providecommand{\doi}[1]{doi: #1}\else
  \providecommand{\doi}{doi: \begingroup \urlstyle{rm}\Url}\fi

\bibitem[Anthropic(2024)]{anthropic2024mcp}
Anthropic.
\newblock Introducing the model context protocol.
\newblock \url{https://www.anthropic.com/news/model-context-protocol}, 2024.

\bibitem[{Axiom Math}(2026)]{axiomimo2026}
{Axiom Math}.
\newblock {AxiomProver} at {IMO} 2026: Lean formalizations and proofs, 2026.
\newblock URL \url{https://github.com/AxiomMath/IMO2026}.

\bibitem[Baker et~al.(2025)Baker, Huizinga, Gao, Dou, Guan, M{\k{a}}dry, Zaremba, Pachocki, and Farhi]{baker2025monitoring}
Bowen Baker, Joost Huizinga, Leo Gao, Zehao Dou, Melody~Y. Guan, Aleksander M{\k{a}}dry, Wojciech Zaremba, J.~Pachocki, and David Farhi.
\newblock Monitoring reasoning models for misbehavior and the risks of promoting obfuscation.
\newblock \emph{arXiv.org}, 2025.
\newblock \doi{10.48550/arXiv.2503.11926}.

\bibitem[Carlini et~al.(2022)Carlini, Ippolito, Jagielski, Lee, Tramèr, and Zhang]{carlini2023quantifying}
Nicholas Carlini, Daphne Ippolito, Matthew Jagielski, Katherine Lee, Florian Tramèr, and Chiyuan Zhang.
\newblock Quantifying memorization across neural language models.
\newblock In \emph{International Conference on Learning Representations}, 2022.

\bibitem[Chen et~al.(2025{\natexlab{a}})Chen, Chen, Du, Hu, Jiang, et~al.]{seedprover152025}
Jiangjie Chen, Wenxiang Chen, Jiacheng Du, Jinyi Hu, Zhicheng Jiang, et~al.
\newblock {Seed-Prover 1.5}: Mastering undergraduate-level theorem proving via learning from experience.
\newblock \emph{arXiv preprint arXiv:2512.17260}, 2025{\natexlab{a}}.
\newblock URL \url{https://arxiv.org/abs/2512.17260}.

\bibitem[Chen et~al.(2025{\natexlab{b}})Chen, Gu, Huang, Huang, Jiang, Jie, Jin, Jin, Li, Ma, Ren, Shen, Shi, Sun, Sun, Wang, Wang, Wang, Wei, Wei, Wu, Wu, Xia, Xin, Yang, Ying, Yuan, Yuan, Zhan, Zhang, Zhang, Zhang, Zhao, Zhao, Zhou, and Zhu]{seedprover2025}
Luoxin Chen, Jinming Gu, Liankai Huang, Wenhao Huang, Zhicheng Jiang, Allan Jie, Xiaoran Jin, Xing Jin, Chenggang Li, Kaijing Ma, Cheng Ren, Jiawei Shen, Wenlei Shi, Tong Sun, He~Sun, Jiahui Wang, Siran Wang, Zhihong Wang, Chenrui Wei, Shufa Wei, Yong-Xu Wu, Yuchen Wu, Yihang Xia, Hua Xin, Fan Yang, Huaiyuan Ying, Hongyi Yuan, Zheng Yuan, Tianyang Zhan, Chi Zhang, Yue Zhang, Ge-Hang Zhang, Tian Zhao, Jianqiu Zhao, Yichi Zhou, and Thomas~(Hanwen) Zhu.
\newblock {Seed-Prover}: Deep and broad reasoning for automated theorem proving.
\newblock \emph{arXiv.org}, 2025{\natexlab{b}}.
\newblock \doi{10.48550/arXiv.2507.23726}.

\bibitem[Chen et~al.(2021)Chen, Tworek, Jun, Yuan, Pinto, Kaplan, Edwards, Burda, Joseph, Brockman, et~al.]{chen2021humaneval}
Mark Chen, Jerry Tworek, Heewoo Jun, Qiming Yuan, Henrique Ponde De~Oliveira Pinto, Jared Kaplan, Harri Edwards, Yuri Burda, Nicholas Joseph, Greg Brockman, et~al.
\newblock Evaluating large language models trained on code.
\newblock \emph{arXiv preprint arXiv:2107.03374}, 2021.

\bibitem[Chung et~al.(2026)Chung, Cai, Li, Yin, Agarwal, Park, Porto, Ri, Yang, Tang, et~al.]{chung2026goedel}
Jui-Hui Chung, Ziyang Cai, Zihao Li, Qishuo Yin, Rohit Agarwal, Simon Park, Rodrigo Porto, Narutatsu Ri, Ziran Yang, Shange Tang, et~al.
\newblock Goedel-architect: Streamlining formal theorem proving with blueprint generation and refinement.
\newblock \emph{arXiv preprint arXiv:2606.06468}, 2026.

\bibitem[Dong and Ma(2025)]{dong2025stp}
Kefan Dong and Tengyu Ma.
\newblock {STP}: Self-play {LLM} theorem provers with iterative conjecturing and proving.
\newblock \emph{International Conference on Machine Learning}, 2025.
\newblock \doi{10.48550/arXiv.2502.00212}.

\bibitem[Dong et~al.(2024)Dong, Jiang, Liu, Jin, Gu, Yang, and Li]{dong2024contamination}
Yihong Dong, Xue Jiang, Huanyu Liu, Zhi Jin, Bin Gu, Mengfei Yang, and Ge~Li.
\newblock Generalization or memorization: Data contamination and trustworthy evaluation for large language models.
\newblock In \emph{Findings of the Association for Computational Linguistics: ACL 2024}, pages 12039--12050, 2024.

\bibitem[Dressler(2025)]{lean-lsp-mcp}
Oliver Dressler.
\newblock {Lean LSP MCP: Tools for agentic interaction with the Lean theorem prover}.
\newblock \url{https://github.com/oOo0oOo/lean-lsp-mcp}, March 2025.

\bibitem[First et~al.(2023)First, Rabe, Ringer, and Brun]{first2023baldur}
Emily First, Markus~N Rabe, Talia Ringer, and Yuriy Brun.
\newblock Baldur: Whole-proof generation and repair with large language models.
\newblock \emph{ESEC/SIGSOFT FSE}, pages 1229--1241, 2023.
\newblock \doi{10.1145/3611643.3616243}.

\bibitem[Freer(2025)]{lean4-skills}
Cameron Freer.
\newblock Lean 4 {Skills}: Theorem proving skill and workflow pack for {AI} coding agents.
\newblock \url{https://github.com/cameronfreer/lean4-skills}, October 2025.

\bibitem[{Harmonic}(2025)]{harmonicverina2025}
{Harmonic}.
\newblock {Aristotle} learns to code, achieving new state-of-the-art of 96.8\% on code verification benchmark, 2025.
\newblock URL \url{https://www.harmonic.fun/news/verina-benchmark/}.

\bibitem[Helff et~al.(2026)Helff, Delfosse, Steinmann, Härle, Shindo, Schramowski, Stammer, Kersting, and Friedrich]{gamingverifiers2026}
Lukas Helff, Quentin Delfosse, David Steinmann, Ruben Härle, Hikaru Shindo, P.~Schramowski, Wolfgang Stammer, Kristian Kersting, and Felix Friedrich.
\newblock {LLMs} gaming verifiers: {RLVR} can lead to reward hacking.
\newblock \emph{arXiv.org}, 2026.
\newblock \doi{10.48550/arXiv.2604.15149}.

\bibitem[Ho and Protzenko(2022)]{aeneas2022}
Son Ho and Jonathan Protzenko.
\newblock {Aeneas}: Rust verification by functional translation.
\newblock \emph{Proc. ACM Program. Lang.}, 6\penalty0 (ICFP):\penalty0 711--741, 2022.
\newblock \doi{10.1145/3547647}.

\bibitem[Ho et~al.(2025)Ho, Boisseau, Franceschino, Prak, Fromherz, and Protzenko]{charon2024}
Son Ho, Guillaume Boisseau, Lucas Franceschino, Yoann Prak, Aymeric Fromherz, and Jonathan Protzenko.
\newblock {Charon}: An analysis framework for rust, 2025.

\bibitem[Jain et~al.(2025)Jain, Han, Gu, Li, Yan, Zhang, Wang, Solar-Lezama, Sen, and Stoica]{livecodebench2024}
Naman Jain, King Han, Alex Gu, Wen-Ding Li, Fanjia Yan, Tianjun Zhang, Sida Wang, Armando Solar-Lezama, Koushik Sen, and Ion Stoica.
\newblock {LiveCodeBench}: Holistic and contamination free evaluation of large language models for code.
\newblock In \emph{International Conference on Learning Representations}, 2025.
\newblock \doi{10.48550/arXiv.2403.07974}.

\bibitem[Jia and Harman(2010)]{jia2011mutation}
Yue Jia and Mark Harman.
\newblock An analysis and survey of the development of mutation testing.
\newblock \emph{IEEE transactions on software engineering}, 37\penalty0 (5):\penalty0 649--678, 2010.

\bibitem[Jiang et~al.(2022)Jiang, Welleck, Zhou, Li, Liu, Jamnik, Lacroix, Wu, and Lample]{jiang2022draft}
Albert~Q Jiang, Sean Welleck, Jin~Peng Zhou, Wenda Li, Jiacheng Liu, Mateja Jamnik, Timoth{\'e}e Lacroix, Yuhuai Wu, and Guillaume Lample.
\newblock Draft, sketch, and prove: Guiding formal theorem provers with informal proofs.
\newblock \emph{International Conference on Learning Representations}, 2022.
\newblock \doi{10.48550/arXiv.2210.12283}.

\bibitem[Jimenez et~al.(2024)Jimenez, Yang, Wettig, Yao, Pei, Press, and Narasimhan]{jimenez2024swebench}
Carlos~E. Jimenez, John Yang, Alexander Wettig, Shunyu Yao, Kexin Pei, Ofir Press, and Karthik Narasimhan.
\newblock {SWE}-bench: Can language models resolve real-world github issues?
\newblock In \emph{International Conference on Learning Representations}, 2024.
\newblock \doi{10.48550/arXiv.2310.06770}.

\bibitem[Klaus et~al.(2026)Klaus, Tolmach, and Conejero]{klaus2026rustlean}
Natalia Klaus, Palina Tolmach, and J.~Conejero.
\newblock A {Rust}-to-{Lean} verification pipeline with {AI} provers: An experience report.
\newblock \emph{arXiv preprint arXiv:2605.30106}, 2026.
\newblock \doi{10.48550/arXiv.2605.30106}.

\bibitem[Kocsis and Szepesvari(2006)]{kocsis2006bandit}
Levente Kocsis and Csaba Szepesvari.
\newblock Bandit based {Monte-Carlo} planning.
\newblock In \emph{European Conference on Machine Learning}, pages 282--293. Springer Berlin Heidelberg, 2006.
\newblock \doi{10.1007/11871842_29}.

\bibitem[Lambert et~al.(2024)Lambert, Morrison, Pyatkin, Huang, Ivison, Brahman, Miranda, Liu, Dziri, Lyu, Gu, Malik, Graf, Hwang, Yang, Bras, Tafjord, Wilhelm, Soldaini, Smith, Wang, Dasigi, and Hajishirzi]{lambert2024tulu3}
Nathan Lambert, Jacob~Daniel Morrison, Valentina Pyatkin, Shengyi Huang, Hamish Ivison, Faeze Brahman, Lester James~Validad Miranda, Alisa Liu, Nouha Dziri, Xinxi Lyu, Yuling Gu, Saumya Malik, Victoria Graf, Jena~D. Hwang, Jiangjiang Yang, Ronan~Le Bras, Oyvind Tafjord, Christopher Wilhelm, Luca Soldaini, Noah~A. Smith, Yizhong Wang, Pradeep Dasigi, and Hanna Hajishirzi.
\newblock {Tulu 3}: Pushing frontiers in open language model post-training.
\newblock \emph{arXiv.org}, 2024.
\newblock \doi{10.48550/arXiv.2411.15124}.

\bibitem[Lample et~al.(2022)Lample, Lachaux, Lavril, Martinet, Hayat, Ebner, Rodriguez, and Lacroix]{lample2022hypertree}
Guillaume Lample, Marie-Anne Lachaux, Thibaut Lavril, Xavier Martinet, Amaury Hayat, Gabriel Ebner, Aur{\'e}lien Rodriguez, and Timoth{\'e}e Lacroix.
\newblock {HyperTree} proof search for neural theorem proving.
\newblock In \emph{Neural Information Processing Systems}, pages 26337--26349. Neural Information Processing Systems Foundation, Inc. (NeurIPS), 2022.
\newblock \doi{10.52202/068431-1910}.

\bibitem[{Lean FRO}(2026)]{comparator}
{Lean FRO}.
\newblock {Comparator}.
\newblock \url{https://github.com/leanprover/comparator}, 2026.
\newblock [Accessed 13-09-2026].

\bibitem[{Lean Prover Community}(2026)]{plausible}
{Lean Prover Community}.
\newblock {Plausible}.
\newblock \url{https://github.com/leanprover-community/plausible}, 2026.
\newblock [Accessed 13-09-2026].

\bibitem[Li et~al.(2026)Li, Yang, He, Zhao, Zhao, Tang, Yang, Gupta, Su, and Jin]{goedelcodeprover2026}
Zenan Li, Ziran Yang, Deyuan He, Haoyu Zhao, Andrew Zhao, Shange Tang, Kaiyu Yang, Aarti Gupta, Zhendong Su, and Chi Jin.
\newblock {Goedel-Code-Prover}: Hierarchical proof search for open state-of-the-art code verification.
\newblock \emph{arXiv.org}, 2026.
\newblock \doi{10.48550/arXiv.2603.19329}.

\bibitem[Lin et~al.(2025)Lin, Tang, Lyu, Yang, Chung, Zhao, Jiang, Geng, Ge, Sun, Wu, Gesi, Lu, Acuna, Yang, Lin, Choi, Chen, Arora, and Jin]{goedelproverv2}
Yong Lin, Shange Tang, Bohan Lyu, Ziran Yang, Jui-Hui Chung, Haoyu Zhao, Lai Jiang, Yihan Geng, Jiawei Ge, Jing Sun, Jiayun Wu, Jiri Gesi, Ximing Lu, David Acuna, Kaiyu Yang, Hongzhou Lin, Yejin Choi, Danqi Chen, Sanjeev Arora, and Chi Jin.
\newblock Goedel-prover-v2: Scaling formal theorem proving with scaffolded data synthesis and self-correction.
\newblock \emph{arXiv.org}, 2025.
\newblock \doi{10.48550/arXiv.2508.03613}.

\bibitem[Liu et~al.(2023)Liu, Xia, Wang, and Zhang]{liu2023evalplus}
Jiawei Liu, Chun Xia, Yuyao Wang, and Lingming Zhang.
\newblock Is your code generated by {ChatGPT} really correct? rigorous evaluation of large language models for code generation.
\newblock In \emph{Neural Information Processing Systems}, pages 21558--21572. Neural Information Processing Systems Foundation, Inc. (NeurIPS), 2023.
\newblock \doi{10.52202/075280-0943}.

\bibitem[Liu et~al.(2026)Liu, Zhou, Zhu, Dos~Santos, He, et~al.]{numinaleanagent2026}
Junqi Liu, Zihao Zhou, Zekai Zhu, Marco Dos~Santos, Weikun He, et~al.
\newblock {Numina-Lean-Agent}: An open and general agentic reasoning system for formal mathematics.
\newblock In \emph{Proceedings of the 43rd International Conference on Machine Learning}, volume 306 of \emph{Proceedings of Machine Learning Research}, pages 80284--80297, 2026.
\newblock URL \url{https://proceedings.mlr.press/v306/liu26gp.html}.

\bibitem[{Logical Intelligence}(2026)]{alephverina2026}
{Logical Intelligence}.
\newblock {Aleph} reaches state-of-the-art across the leading formal reasoning benchmarks as verified code generation nears reality, 2026.
\newblock URL \url{https://logicalintelligence.com/blog/aleph-leading-benchmarks}.

\bibitem[Loughridge et~al.(2024)Loughridge, Sun, Ahrenbach, Cassano, Sun, Sheng, Mudide, Misu, Amin, and Tegmark]{loughridge2024dafnybench}
Chloe Loughridge, Qinyi Sun, Seth Ahrenbach, Federico Cassano, Chuyue Sun, Ying Sheng, Anish Mudide, Md~Rakib~Hossain Misu, Nada Amin, and Max Tegmark.
\newblock {DafnyBench}: A benchmark for formal software verification.
\newblock \emph{Trans. Mach. Learn. Res.}, 2024.
\newblock \doi{10.48550/arXiv.2406.08467}.

\bibitem[{Mistral AI}(2026)]{mistralLeanstral15}
{Mistral AI}.
\newblock {Leanstral 1.5}: Proof abundance for all.
\newblock \url{https://mistral.ai/news/leanstral-1-5/}, 2026.
\newblock [Accessed 13-09-2026].

\bibitem[Moura and Ullrich(2021)]{demoura2021lean4}
L.~D. Moura and Sebastian Ullrich.
\newblock The {Lean 4} theorem prover and programming language.
\newblock In \emph{CADE}, pages 625--635. Springer International Publishing, 2021.
\newblock \doi{10.1007/978-3-030-79876-5_37}.

\bibitem[Poesia et~al.(2024)Poesia, Broman, Haber, and Goodman]{poesia2024minimo}
Gabriel Poesia, David Broman, Nick Haber, and Noah~D. Goodman.
\newblock Learning formal mathematics from intrinsic motivation.
\newblock In \emph{Advances in Neural Information Processing Systems 37}, pages 43032--43057. Neural Information Processing Systems Foundation, Inc. (NeurIPS), 2024.
\newblock \doi{10.52202/079017-1362}.

\bibitem[Polu and Sutskever(2020)]{polu2020gpt}
Stanislas Polu and Ilya Sutskever.
\newblock Generative language modeling for automated theorem proving.
\newblock In \emph{arXiv.org}, 2020.

\bibitem[Polu et~al.(2022)Polu, Han, Zheng, Baksys, Babuschkin, and Sutskever]{polu2022curriculum}
Stanislas Polu, Jesse~Michael Han, Kunhao Zheng, Mantas Baksys, Igor Babuschkin, and Ilya Sutskever.
\newblock Formal mathematics statement curriculum learning.
\newblock \emph{International Conference on Learning Representations}, 2022.
\newblock \doi{10.48550/arXiv.2202.01344}.

\bibitem[{PutnamBench Project}(2026)]{putnambenchleaderboard2026}
{PutnamBench Project}.
\newblock {PutnamBench}: Live leaderboard (lean with-answer and no-answer tracks), 2026.
\newblock URL \url{https://trishullab.github.io/PutnamBench/leaderboard}.
\newblock Dynamic leaderboard; accessed October 2026.

\bibitem[Ren et~al.(2025)Ren, Shao, Song, Xin, Wang, Zhao, Zhang, Fu, Zhu, Yang, et~al.]{ren2025deepseek}
ZZ~Ren, Zhihong Shao, Junxiao Song, Huajian Xin, Haocheng Wang, Wanjia Zhao, Liyue Zhang, Zhe Fu, Qihao Zhu, Dejian Yang, et~al.
\newblock Deepseek-prover-v2: Advancing formal mathematical reasoning via reinforcement learning for subgoal decomposition.
\newblock \emph{arXiv preprint arXiv:2504.21801}, 2025.

\bibitem[Requena et~al.(2026)Requena, Letson, Nowakowski, Ferreiro, and Sarra]{requena2026minimalagent}
Borja Requena, Austin Letson, Krystian Nowakowski, Izan~Beltran Ferreiro, and Leopoldo Sarra.
\newblock A minimal agent for automated theorem proving.
\newblock In \emph{Proceedings of the 43rd International Conference on Machine Learning}, volume 306 of \emph{Proceedings of Machine Learning Research}, pages 104861--104905, 2026.
\newblock URL \url{https://proceedings.mlr.press/v306/requena26a.html}.

\bibitem[Silver et~al.(2016)Silver, Huang, Maddison, Guez, Sifre, Driessche, Schrittwieser, Antonoglou, Panneershelvam, Lanctot, Dieleman, Grewe, Nham, Kalchbrenner, Sutskever, Lillicrap, Leach, Kavukcuoglu, Graepel, and Hassabis]{silver2016alphago}
David Silver, Aja Huang, Chris~J. Maddison, A.~Guez, L.~Sifre, George van~den Driessche, Julian Schrittwieser, Ioannis Antonoglou, Vedavyas Panneershelvam, Marc Lanctot, S.~Dieleman, Dominik Grewe, John Nham, Nal Kalchbrenner, I.~Sutskever, T.~Lillicrap, Madeleine Leach, K.~Kavukcuoglu, T.~Graepel, and D.~Hassabis.
\newblock Mastering the game of {Go} with deep neural networks and tree search.
\newblock \emph{Nature}, 529\penalty0 (7587):\penalty0 484--489, 2016.
\newblock \doi{10.1038/nature16961}.

\bibitem[Thakur et~al.(2025)Thakur, Lee, Tsoukalas, Sistla, Zhao, Zetzsche, Durrett, Yue, and Chaudhuri]{clever2025}
A.~Thakur, Jasper Lee, G.~Tsoukalas, M.~Sistla, Matthew Zhao, Stefan Zetzsche, Greg Durrett, Yisong Yue, and Swarat Chaudhuri.
\newblock {CLEVER}: A curated benchmark for formally verified code generation.
\newblock \emph{Neural Information Processing Systems}, 2025.
\newblock \doi{10.48550/arXiv.2505.13938}.

\bibitem[Tsoukalas et~al.(2024)Tsoukalas, Lee, Jennings, Xin, Ding, Jennings, Thakur, and Chaudhuri]{putnambench2024}
George Tsoukalas, Jasper Lee, John Jennings, Jimmy Xin, Michelle Ding, Michael Jennings, Amitayush Thakur, and Swarat Chaudhuri.
\newblock {PutnamBench}: Evaluating neural theorem-provers on the {Putnam} mathematical competition.
\newblock \emph{Advances in Neural Information Processing Systems 37}, pages 11545--11569, 2024.
\newblock \doi{10.52202/079017-0368}.

\bibitem[Varambally et~al.(2025)Varambally, Voice, Sun, Chen, Yu, and Ye]{hilbert2025}
Sumanth Varambally, Thomas Voice, Yanchao Sun, Zhifeng Chen, Rose Yu, and Ke~Ye.
\newblock {Hilbert}: Recursively building formal proofs with informal reasoning.
\newblock \emph{arXiv preprint arXiv:2509.22819}, 2025.
\newblock URL \url{https://arxiv.org/abs/2509.22819}.

\bibitem[Wang et~al.(2023)Wang, Xie, Jiang, Mandlekar, Xiao, Zhu, Fan, and Anandkumar]{wang2023voyager}
Guanzhi Wang, Yuqi Xie, Yunfan Jiang, Ajay Mandlekar, Chaowei Xiao, Yuke Zhu, Linxi Fan, and Anima Anandkumar.
\newblock Voyager: An open-ended embodied agent with large language models.
\newblock \emph{Trans. Mach. Learn. Res.}, 2023.
\newblock \doi{10.48550/arXiv.2305.16291}.

\bibitem[Wang et~al.(2026)Wang, Zhang, Guo, et~al.]{longcat2026}
Jianing Wang, Jianfei Zhang, Qi~Guo, et~al.
\newblock {LongCat-Flash-Prover}: Advancing native formal reasoning via agentic tool-integrated reinforcement learning.
\newblock \emph{arXiv preprint arXiv:2603.21065}, 2026.
\newblock URL \url{https://arxiv.org/abs/2603.21065}.

\bibitem[Xia et~al.(2025)Xia, Shen, Wang, Liu, Sun, Wu, Hu, and Xu]{xia2025leetcodedataset}
Yunhui Xia, Wei Shen, Yan Wang, Jason~Klein Liu, Huifeng Sun, Siyue Wu, Jian Hu, and Xiaolong Xu.
\newblock Leetcodedataset: A temporal dataset for robust evaluation and efficient training of code llms.
\newblock \emph{arXiv preprint arXiv:2504.14655}, 2025.

\bibitem[Xin et~al.(2024)Xin, Guo, Shao, Ren, Zhu, Liu, Ruan, Li, and Liang]{deepseekproverv1}
Huajian Xin, Daya Guo, Zhihong Shao, Zhizhou Ren, Qihao Zhu, Bo~Liu, Chong Ruan, Wenda Li, and Xiaodan Liang.
\newblock {DeepSeek-Prover}: Advancing theorem proving in {LLMs} through large-scale synthetic data.
\newblock \emph{arXiv.org}, 2024.
\newblock \doi{10.48550/arXiv.2405.14333}.

\bibitem[Xu et~al.(2026)Xu, Luan, Wang, Leang, Wang, Li, Li, and Watt]{ntp4vc2026}
Qiyuan Xu, Xiaokun Luan, Renxi Wang, Joshua Ong~Jun Leang, Peixin Wang, Haonan Li, Wenda Li, and Conrad Watt.
\newblock Neural theorem proving for verification conditions: A real-world benchmark.
\newblock In \emph{International Conference on Learning Representations (ICLR)}, 2026.
\newblock \doi{10.48550/arXiv.2601.18944}.

\bibitem[Yang et~al.(2023)Yang, Swope, Gu, Chalamala, Song, Yu, Godil, Prenger, and Anandkumar]{leandojo2023}
Kaiyu Yang, Aidan Swope, Alex Gu, Rahul Chalamala, Peiyang Song, Shixing Yu, Saad Godil, Ryan Prenger, and Anima Anandkumar.
\newblock {LeanDojo}: Theorem proving with retrieval-augmented language models.
\newblock \emph{Advances in Neural Information Processing Systems 36}, pages 21573--21612, 2023.
\newblock \doi{10.52202/075280-0944}.

\bibitem[Yao et~al.(2022)Yao, Zhao, Yu, Du, Shafran, Narasimhan, and Cao]{yao2023react}
Shunyu Yao, Jeffrey Zhao, Dian Yu, Nan Du, Izhak Shafran, Karthik Narasimhan, and Yuan Cao.
\newblock {ReAct}: Synergizing reasoning and acting in language models.
\newblock \emph{International Conference on Learning Representations}, 2022.

\bibitem[Ye et~al.(2025)Ye, Yan, He, Kasriel, Yang, and Song]{verina2025}
Zhe Ye, Zhengxu Yan, Jingxuan He, Timothé Kasriel, Kaiyu Yang, and Dawn Song.
\newblock {Verina}: Benchmarking verifiable code generation.
\newblock \emph{arXiv.org}, 2025.
\newblock \doi{10.48550/arXiv.2505.23135}.

\bibitem[Zhang et~al.(2025)Zhang, Kraska, and Khattab]{zhang2025recursive}
Alex~L Zhang, Tim Kraska, and Omar Khattab.
\newblock Recursive language models.
\newblock \emph{arXiv preprint arXiv:2512.24601}, 2025.

\bibitem[Zhang et~al.(2026)Zhang, Xiong, Lin, Chung, Zhang, Han, Liu, and Zhu]{zhang2026humanizeolympiadagentsmultisession}
Zhengyang Zhang, Jing Xiong, Hongzhou Lin, Juihui Chung, Zijian Zhang, Song Han, Sihao Liu, and Ligeng Zhu.
\newblock Humanize olympiad agents: A multi-session workflow for olympiad problem solving with solve, review, wiki, verify, 2026.
\newblock URL \url{https://arxiv.org/abs/2610.13632}.

\bibitem[Zhao et~al.(2025)Zhao, Wu, Yue, Wu, Xu, Lin, Wang, Wu, Zheng, and Huang]{zhao2025absolutezero}
Andrew Zhao, Yiran Wu, Yang Yue, Tong Wu, Quentin Xu, Matthieu Lin, Shenzhi Wang, Qingyun Wu, Zilong Zheng, and Gao Huang.
\newblock Absolute zero: Reinforced self-play reasoning with zero data.
\newblock \emph{Neural Information Processing Systems}, 2025.
\newblock \doi{10.48550/arXiv.2505.03335}.

\bibitem[Zhao et~al.(2026)Zhao, Yang, Li, He, Li, Jin, Veeravalli, Gupta, and Arora]{zhao2026algoveri}
Haoyu Zhao, Ziran Yang, Jiawei Li, Deyuan He, Zenan Li, Chi Jin, Venugopal~V Veeravalli, Aarti Gupta, and Sanjeev Arora.
\newblock Algoveri: An aligned benchmark for verified code generation on classical algorithms.
\newblock \emph{arXiv preprint arXiv:2602.09464}, 2026.

\bibitem[Zheng et~al.(2021)Zheng, Han, and Polu]{zheng2021minif2f}
Kunhao Zheng, Jesse~Michael Han, and Stanislas Polu.
\newblock Minif2f: a cross-system benchmark for formal olympiad-level mathematics.
\newblock \emph{arXiv preprint arXiv:2109.00110}, 2021.

\bibitem[Zhou et~al.(2024)Zhou, Staats, Li, Szegedy, Weinberger, and Wu]{zhou2024lego}
Jin~Peng Zhou, Charles Staats, Wenda Li, Christian Szegedy, Kilian~Q Weinberger, and Yuhuai Wu.
\newblock Don't trust: Verify -- grounding {LLM} quantitative reasoning with autoformalization.
\newblock \emph{International Conference on Learning Representations}, 2024.

\bibitem[Zhu et~al.(2025)Zhu, Clune, Avigad, Jiang, and Welleck]{zhu2025leanhammer}
Thomas Zhu, Joshua Clune, Jeremy Avigad, Albert~Q. Jiang, and Sean Welleck.
\newblock Premise selection for a lean hammer.
\newblock \emph{arXiv preprint arXiv:2506.07477}, 2025.

\end{thebibliography}
\bibliographystyle{plainnat}

\appendix

\providecommand{\filename}[1]{\texttt{\detokenize{#1}}}

\newpage

\section{Case Studies and Measurement Conventions}
\label{app:translated-case-studies}

This appendix examines seven safe Rust programs translated to Lean by
\aeneas{}.  The source programs use ordinary loops, arrays, and fixed-width
integers.  Their Lean translations make checked arithmetic, bounds-checked
indexing, failure, termination, loop-carried state, and mutation explicit, and
the proofs must handle each of these.

For each program, we state the problem and the frozen theorem and explain the
proof obligations that the translation introduces.  We then list the complete
Rust source, the executable Lean definitions generated by \aeneas{}, and the
accepted proof.  For the last program, which was not proved, we list the
unresolved theorem instead.

None of the selected programs introduces a user-defined datatype.  Their
\filename{Types.lean} files consequently contain only imports, namespace
declarations, and configuration options.  We omit that repeated boilerplate and
reproduce the complete executable definitions from \filename{Funs.lean}.

Two notational conventions recur below.  For a fixed-width scalar such as
\lstinline{x : Std.I64} or \lstinline{i : Std.Usize}, the projection
\lstinline{.val} exposes its mathematical integer or natural-number value.  The
weakest-precondition judgment \lstinline{f x ⦃ v => Q v ⦄} asserts that the
computation terminates with \lstinline{ok v} and that $Q(v)$ holds.  A
postcondition of \lstinline{True} therefore still proves totality: neither
failure nor divergence satisfies the judgment.

The generated listings split each Rust loop into two definitions.  A loop body
returns \lstinline{Result (ControlFlow σ α)}: \lstinline{cont} carries the next
loop state of type $\sigma$, while \lstinline{done} carries the final result of
type $\alpha$.  A separate driver invokes the body through the \aeneas{} loop
combinator.  We refer to these two definitions as a loop-body/driver pair.
Mutation is translated in the same functional style.  The operation
\lstinline{index_mut} returns the current element together with a backward
function that builds the updated vector.

\subsection{Measuring proof size}

The 1,054 accepted proof artifacts occupy 145,646 physical lines.
The median artifact has 104 lines, and the largest has 1,167.  Each count
includes comments, mathematical models, helper lemmas, loop invariants, lemmas
relating the generated code to those models, and the target proof.  We count the complete accepted artifact because its supporting
definitions and lemmas are part of the submitted verification development.

In \cref{tab:appendix-case-summary}, ``Added'' is the number of lines
introduced when the original one-line \lstinline{sorry} is replaced by the
accepted development.  ``Rust'' and ``Aeneas'' report physical line counts for
\filename{solution.rs} and \filename{Funs.lean}.

\begin{table}[ht]
\centering
\small
\begin{tabularx}{\textwidth}{@{}Xrrrr>{\raggedright\arraybackslash}p{0.30\textwidth}@{}}
\toprule
Case & Rust & Aeneas & Proof & Added & Principal obligation \\
\midrule
Diagonal distinct-value shift invariance & 55 & 253 & 1167 & 1108 & Relating two executions through six loop components \\
Triangular-sum reversal invariance & 22 & 124 & 431 & 395 & List reduction, modular arithmetic, and relational symmetry \\
Subset-XOR exact correctness & 19 & 83 & 412 & 391 & Bitvectors, sublists, and overflow-safe multiplication \\
Matrix reshape exact correctness & 50 & 387 & 366 & 318 & Nested vectors, casts, products, division, and modulo \\
Even/odd independent sort totality and output length & 67 & 281 & 336 & 307 & Mutation, parity accounting, and six loops \\
Relative sort totality & 59 & 293 & 289 & 244 & Insertion-sort writes and backward closures \\
Bottom-up merge-sort totality & 66 & 328 & --- & --- & Unresolved: eight loops and coupled cursor bounds \\
\bottomrule
\end{tabularx}
\caption{Physical line counts for the selected programs, translations, and
proof artifacts, ordered by accepted proof size.  The merge-sort theorem is
unresolved.}
\label{tab:appendix-case-summary}
\end{table}

\clearpage
\section{Independent Sorting at Even and Odd Indices}
\label{app:sort-even-odd}

\subsection{Problem and source-level approach}

The input is an array of signed integers.  Values at even indices are sorted in
nondecreasing order, values at odd indices are sorted in nonincreasing order,
and the two sequences are returned to their original index parities.  The
implementation scans the input once to split it into two vectors, applies a
quadratic selection sort to each vector, and interleaves the results.  In the
Rust source, a swap requires two indexed assignments, and the final loop pushes
one element for each input position.

\subsection{What the proof must establish}

The frozen theorem
\filename{sort_even_odd_output_bound} does not claim that the output is sorted.
It proves that the translated call succeeds and that the returned vector has
exactly the same length as the input.  It has no explicit precondition, so its
domain is broader than the original problem constraints of at most 100 elements
with values in $[1,100]$.  The program performs no arithmetic on element values,
and its vectors never grow beyond the input length.  The result is nevertheless
available only after all preceding monadic computations succeed, so the proof
must verify the two sorting phases even though their element order is irrelevant
to the postcondition.

\aeneas{} produces six loop-body/driver pairs.  Each selection-sort inner loop reads
two indexed elements and implements a swap through two successive
\lstinline{alloc.vec.Vec.index_mut} calls.  The proof must show that each
reconstructed vector has the same length, so later indices remain valid.  It
also tracks exact parity counts: after processing $i$ positions, the two
temporary lengths are $(i+1)/2$ and $i/2$.  These equalities justify the reads
performed by the final interleaving loop.

The proof is organized bottom-up:

\begin{enumerate}
  \item prove the splitting loop's exact length invariants for both vectors;
  \item prove that each inner selection-sort loop preserves vector length;
  \item lift those facts to the two outer sorting loops;
  \item prove a four-component invariant for the emission loop, connecting its
        cursor to the result length and to the two parity cursors;
  \item compose the loop specifications and conclude totality and exact output
        length for the top-level function.
\end{enumerate}

\subsection{Complete Rust implementation}
\begin{lstlisting}[language=Rust,style=code]
pub fn sort_even_odd(nums: &[i64]) -> Vec<i64> {
    let n = nums.len();

    // Collect even-indexed values
    let mut evens: Vec<i64> = Vec::new();
    let mut odds: Vec<i64> = Vec::new();

    let mut i: usize = 0;
    while i < n {
        if i % 2 == 0 {
            evens.push(nums[i]);
        } else {
            odds.push(nums[i]);
        }
        i += 1;
    }

    // Sort evens in non-decreasing order (ascending)
    let en = evens.len();
    let mut a: usize = 0;
    while a < en {
        let mut b: usize = a + 1;
        while b < en {
            if evens[b] < evens[a] {
                let tmp = evens[a];
                evens[a] = evens[b];
                evens[b] = tmp;
            }
            b += 1;
        }
        a += 1;
    }

    // Sort odds in non-increasing order (descending)
    let on = odds.len();
    let mut c: usize = 0;
    while c < on {
        let mut d: usize = c + 1;
        while d < on {
            if odds[d] > odds[c] {
                let tmp = odds[c];
                odds[c] = odds[d];
                odds[d] = tmp;
            }
            d += 1;
        }
        c += 1;
    }

    // Merge back
    let mut result: Vec<i64> = Vec::new();
    let mut ei: usize = 0;
    let mut oi: usize = 0;
    let mut k: usize = 0;
    while k < n {
        if k % 2 == 0 {
            result.push(evens[ei]);
            ei += 1;
        } else {
            result.push(odds[oi]);
            oi += 1;
        }
        k += 1;
    }

    result
}
\end{lstlisting}

\subsection{Complete \aeneas{}-generated Lean implementation}
\begin{lstlisting}[style=code]
-- THIS FILE WAS AUTOMATICALLY GENERATED BY AENEAS
-- [sort_even_odd]: function definitions
import Aeneas
import Aegis.SortEvenOdd.Code.Types
open Aeneas Aeneas.Std Result ControlFlow Error
set_option linter.dupNamespace false
set_option linter.hashCommand false
set_option linter.unusedVariables false

/- You can set the `maxHeartbeats` value with the `-max-heartbeats` CLI option -/
set_option maxHeartbeats 1000000

/- You can set the `maxRecDepth` value with the `-max-recdepth` CLI option -/
set_option maxRecDepth 2048

namespace Aegis.SortEvenOdd

/-- [sort_even_odd::sort_even_odd]: loop body 0:
    Source: 'sort_even_odd/src/lib.rs', lines 9:4-16:5
    Visibility: public -/
@[rust_loop_body]
def sort_even_odd_loop0.body
  (nums : Slice Std.I64) (n : Std.Usize) (evens : alloc.vec.Vec Std.I64)
  (odds : alloc.vec.Vec Std.I64) (i : Std.Usize) :
  Result (ControlFlow ((alloc.vec.Vec Std.I64) × (alloc.vec.Vec Std.I64) ×
    Std.Usize) ((alloc.vec.Vec Std.I64) × (alloc.vec.Vec Std.I64)))
  := do
  if i < n
  then
    let i1 ← i % 2#usize
    let (evens1, odds1) ←
      if i1 = 0#usize
      then
        do
        let i2 ← Slice.index_usize nums i
        let evens2 ← alloc.vec.Vec.push evens i2
        ok (evens2, odds)
      else
        do
        let i2 ← Slice.index_usize nums i
        let odds2 ← alloc.vec.Vec.push odds i2
        ok (evens, odds2)
    let i2 ← i + 1#usize
    ok (cont (evens1, odds1, i2))
  else ok (done (evens, odds))

/-- [sort_even_odd::sort_even_odd]: loop 0:
    Source: 'sort_even_odd/src/lib.rs', lines 9:4-16:5
    Visibility: public -/
@[rust_loop]
def sort_even_odd_loop0
  (nums : Slice Std.I64) (n : Std.Usize) (evens : alloc.vec.Vec Std.I64)
  (odds : alloc.vec.Vec Std.I64) (i : Std.Usize) :
  Result ((alloc.vec.Vec Std.I64) × (alloc.vec.Vec Std.I64))
  := do
  loop
    (fun (evens1, odds1, i1) => sort_even_odd_loop0.body nums n evens1 odds1
      i1)
    (evens, odds, i)

/-- [sort_even_odd::sort_even_odd]: loop body 2:
    Source: 'sort_even_odd/src/lib.rs', lines 23:8-30:9
    Visibility: public -/
@[rust_loop_body]
def sort_even_odd_loop1_loop0.body
  (en : Std.Usize) (a : Std.Usize) (evens : alloc.vec.Vec Std.I64)
  (b : Std.Usize) :
  Result (ControlFlow ((alloc.vec.Vec Std.I64) × Std.Usize) (alloc.vec.Vec
    Std.I64))
  := do
  if b < en
  then
    let i ←
      alloc.vec.Vec.index (core.slice.index.SliceIndexUsizeSlice Std.I64) evens
        b
    let i1 ←
      alloc.vec.Vec.index (core.slice.index.SliceIndexUsizeSlice Std.I64) evens
        a
    let evens1 ←
      if i < i1
      then
        do
        let (_, index_mut_back) ←
          alloc.vec.Vec.index_mut (core.slice.index.SliceIndexUsizeSlice
            Std.I64) evens a
        let evens2 := index_mut_back i
        let (_, index_mut_back1) ←
          alloc.vec.Vec.index_mut (core.slice.index.SliceIndexUsizeSlice
            Std.I64) evens2 b
        ok (index_mut_back1 i1)
      else ok evens
    let b1 ← b + 1#usize
    ok (cont (evens1, b1))
  else ok (done evens)

/-- [sort_even_odd::sort_even_odd]: loop 2:
    Source: 'sort_even_odd/src/lib.rs', lines 23:8-30:9
    Visibility: public -/
@[rust_loop]
def sort_even_odd_loop1_loop0
  (evens : alloc.vec.Vec Std.I64) (en : Std.Usize) (a : Std.Usize)
  (b : Std.Usize) :
  Result (alloc.vec.Vec Std.I64)
  := do
  loop
    (fun (evens1, b1) => sort_even_odd_loop1_loop0.body en a evens1 b1)
    (evens, b)

/-- [sort_even_odd::sort_even_odd]: loop body 1:
    Source: 'sort_even_odd/src/lib.rs', lines 21:4-32:5
    Visibility: public -/
@[rust_loop_body]
def sort_even_odd_loop1.body
  (en : Std.Usize) (evens : alloc.vec.Vec Std.I64) (a : Std.Usize) :
  Result (ControlFlow ((alloc.vec.Vec Std.I64) × Std.Usize) (alloc.vec.Vec
    Std.I64))
  := do
  if a < en
  then
    let b ← a + 1#usize
    let evens1 ← sort_even_odd_loop1_loop0 evens en a b
    let a1 ← a + 1#usize
    ok (cont (evens1, a1))
  else ok (done evens)

/-- [sort_even_odd::sort_even_odd]: loop 1:
    Source: 'sort_even_odd/src/lib.rs', lines 21:4-32:5
    Visibility: public -/
@[rust_loop]
def sort_even_odd_loop1
  (evens : alloc.vec.Vec Std.I64) (en : Std.Usize) (a : Std.Usize) :
  Result (alloc.vec.Vec Std.I64)
  := do
  loop
    (fun (evens1, a1) => sort_even_odd_loop1.body en evens1 a1)
    (evens, a)

/-- [sort_even_odd::sort_even_odd]: loop body 4:
    Source: 'sort_even_odd/src/lib.rs', lines 39:8-46:9
    Visibility: public -/
@[rust_loop_body]
def sort_even_odd_loop2_loop0.body
  (on : Std.Usize) (c : Std.Usize) (odds : alloc.vec.Vec Std.I64)
  (d : Std.Usize) :
  Result (ControlFlow ((alloc.vec.Vec Std.I64) × Std.Usize) (alloc.vec.Vec
    Std.I64))
  := do
  if d < on
  then
    let i ←
      alloc.vec.Vec.index (core.slice.index.SliceIndexUsizeSlice Std.I64) odds
        d
    let i1 ←
      alloc.vec.Vec.index (core.slice.index.SliceIndexUsizeSlice Std.I64) odds
        c
    let odds1 ←
      if i > i1
      then
        do
        let (_, index_mut_back) ←
          alloc.vec.Vec.index_mut (core.slice.index.SliceIndexUsizeSlice
            Std.I64) odds c
        let odds2 := index_mut_back i
        let (_, index_mut_back1) ←
          alloc.vec.Vec.index_mut (core.slice.index.SliceIndexUsizeSlice
            Std.I64) odds2 d
        ok (index_mut_back1 i1)
      else ok odds
    let d1 ← d + 1#usize
    ok (cont (odds1, d1))
  else ok (done odds)

/-- [sort_even_odd::sort_even_odd]: loop 4:
    Source: 'sort_even_odd/src/lib.rs', lines 39:8-46:9
    Visibility: public -/
@[rust_loop]
def sort_even_odd_loop2_loop0
  (odds : alloc.vec.Vec Std.I64) (on : Std.Usize) (c : Std.Usize)
  (d : Std.Usize) :
  Result (alloc.vec.Vec Std.I64)
  := do
  loop
    (fun (odds1, d1) => sort_even_odd_loop2_loop0.body on c odds1 d1)
    (odds, d)

/-- [sort_even_odd::sort_even_odd]: loop body 3:
    Source: 'sort_even_odd/src/lib.rs', lines 37:4-48:5
    Visibility: public -/
@[rust_loop_body]
def sort_even_odd_loop2.body
  (on : Std.Usize) (odds : alloc.vec.Vec Std.I64) (c : Std.Usize) :
  Result (ControlFlow ((alloc.vec.Vec Std.I64) × Std.Usize) (alloc.vec.Vec
    Std.I64))
  := do
  if c < on
  then
    let d ← c + 1#usize
    let odds1 ← sort_even_odd_loop2_loop0 odds on c d
    let c1 ← c + 1#usize
    ok (cont (odds1, c1))
  else ok (done odds)

/-- [sort_even_odd::sort_even_odd]: loop 3:
    Source: 'sort_even_odd/src/lib.rs', lines 37:4-48:5
    Visibility: public -/
@[rust_loop]
def sort_even_odd_loop2
  (odds : alloc.vec.Vec Std.I64) (on : Std.Usize) (c : Std.Usize) :
  Result (alloc.vec.Vec Std.I64)
  := do
  loop
    (fun (odds1, c1) => sort_even_odd_loop2.body on odds1 c1)
    (odds, c)

/-- [sort_even_odd::sort_even_odd]: loop body 5:
    Source: 'sort_even_odd/src/lib.rs', lines 55:4-64:5
    Visibility: public -/
@[rust_loop_body]
def sort_even_odd_loop3.body
  (n : Std.Usize) (evens : alloc.vec.Vec Std.I64)
  (odds : alloc.vec.Vec Std.I64) (result : alloc.vec.Vec Std.I64)
  (ei : Std.Usize) (oi : Std.Usize) (k : Std.Usize) :
  Result (ControlFlow ((alloc.vec.Vec Std.I64) × Std.Usize × Std.Usize ×
    Std.Usize) (alloc.vec.Vec Std.I64))
  := do
  if k < n
  then
    let i ← k % 2#usize
    let (result1, ei1, oi1) ←
      if i = 0#usize
      then
        do
        let i1 ←
          alloc.vec.Vec.index (core.slice.index.SliceIndexUsizeSlice Std.I64)
            evens ei
        let result2 ← alloc.vec.Vec.push result i1
        let ei2 ← ei + 1#usize
        ok (result2, ei2, oi)
      else
        do
        let i1 ←
          alloc.vec.Vec.index (core.slice.index.SliceIndexUsizeSlice Std.I64)
            odds oi
        let result2 ← alloc.vec.Vec.push result i1
        let oi2 ← oi + 1#usize
        ok (result2, ei, oi2)
    let k1 ← k + 1#usize
    ok (cont (result1, ei1, oi1, k1))
  else ok (done result)

/-- [sort_even_odd::sort_even_odd]: loop 5:
    Source: 'sort_even_odd/src/lib.rs', lines 55:4-64:5
    Visibility: public -/
@[rust_loop]
def sort_even_odd_loop3
  (n : Std.Usize) (evens : alloc.vec.Vec Std.I64)
  (odds : alloc.vec.Vec Std.I64) (result : alloc.vec.Vec Std.I64)
  (ei : Std.Usize) (oi : Std.Usize) (k : Std.Usize) :
  Result (alloc.vec.Vec Std.I64)
  := do
  loop
    (fun (result1, ei1, oi1, k1) => sort_even_odd_loop3.body n evens odds
      result1 ei1 oi1 k1)
    (result, ei, oi, k)

/-- [sort_even_odd::sort_even_odd]:
    Source: 'sort_even_odd/src/lib.rs', lines 1:0-67:1
    Visibility: public -/
def sort_even_odd (nums : Slice Std.I64) : Result (alloc.vec.Vec Std.I64) := do
  let n := Slice.len nums
  let (evens, odds) ←
    sort_even_odd_loop0 nums n (alloc.vec.Vec.new Std.I64) (alloc.vec.Vec.new
      Std.I64) 0#usize
  let en := alloc.vec.Vec.len evens
  let evens1 ← sort_even_odd_loop1 evens en 0#usize
  let on := alloc.vec.Vec.len odds
  let odds1 ← sort_even_odd_loop2 odds on 0#usize
  sort_even_odd_loop3 n evens1 odds1 (alloc.vec.Vec.new Std.I64) 0#usize
    0#usize 0#usize

end Aegis.SortEvenOdd
\end{lstlisting}

\subsection{Accepted proof artifact}
\begin{lstlisting}[style=code]
import Aegis.SortEvenOdd.Code.Funs
open Aeneas Aeneas.Std Result

namespace Aegis.SortEvenOdd

/-! ## Proof scaffolding: length specifications for the four loops.

Counting facts: the number of even indices in `[0, i)` is `(i + 1) / 2` and the number of odd
indices is `i / 2`. Loop 0 splits the input by index parity, loops 1 and 2 sort `evens`/`odds`
in place (preserving length), and loop 3 emits exactly one element per index `k < n`. -/

/-- Splitting loop: from a state where `evens`/`odds` hold the parity-split of the first `i`
indices, the loop succeeds and returns vectors of lengths `(n+1)/2` and `n/2`. -/
private theorem loop0_len_spec (nums : Slice Std.I64) (n : Std.Usize)
    (evens odds : alloc.vec.Vec Std.I64) (i : Std.Usize)
    (hn : n.val = nums.val.length) (hi : i.val ≤ n.val)
    (he : evens.val.length = (i.val + 1) / 2) (ho : odds.val.length = i.val / 2) :
    sort_even_odd_loop0 nums n evens odds i ⦃ p =>
      p.1.val.length = (n.val + 1) / 2 ∧ p.2.val.length = n.val / 2 ⦄ := by
  have hslen := nums.length_ineq
  unfold sort_even_odd_loop0
  apply Std.loop.spec_decr_nat
    (measure := fun p => n.val - p.2.2.val)
    (inv := fun p => p.2.2.val ≤ n.val ∧ p.1.val.length = (p.2.2.val + 1) / 2 ∧
      p.2.1.val.length = p.2.2.val / 2)
  · rintro ⟨evens1, odds1, i1⟩ ⟨hle, he1, ho1⟩
    simp only at hle he1 ho1
    unfold sort_even_odd_loop0.body
    by_cases hlt : i1 < n
    · simp only [hlt, if_true]
      apply WP.spec_bind (Usize.rem_spec i1 (by simp))
      intro r hr
      have hbound : i1.val < nums.val.length := by scalar_tac
      have hlt' : i1.val < n.val := by scalar_tac
      by_cases hz : r = 0#usize
      · simp only [hz, if_true, Std.bind_assoc_eq]
        have hpar : i1.val % 2 = 0 := by scalar_tac
        apply WP.spec_bind (Slice.index_usize_spec nums i1 hbound)
        intro a ha
        have hpushlen : (evens1.val).length < Usize.max := by scalar_tac
        apply WP.spec_bind (alloc.vec.Vec.push_spec evens1 a hpushlen)
        intro evens2 hpush
        have hlen2 : evens2.val.length = evens1.val.length + 1 := by
          simp [hpush]
        simp only [Std.bind_tc_ok]
        apply WP.spec_bind (Usize.add_spec (by scalar_tac))
        intro i2 hi2
        have hi2' : i2.val = i1.val + 1 := by scalar_tac
        simp only [WP.spec_ok]
        refine ⟨⟨by scalar_tac, by omega, by omega⟩, by scalar_tac⟩
      · simp only [hz, if_false, Std.bind_assoc_eq]
        have hpar : i1.val % 2 = 1 := by scalar_tac
        apply WP.spec_bind (Slice.index_usize_spec nums i1 hbound)
        intro a ha
        have hpushlen : (odds1.val).length < Usize.max := by scalar_tac
        apply WP.spec_bind (alloc.vec.Vec.push_spec odds1 a hpushlen)
        intro odds2 hpush
        have hlen2 : odds2.val.length = odds1.val.length + 1 := by
          simp [hpush]
        simp only [Std.bind_tc_ok]
        apply WP.spec_bind (Usize.add_spec (by scalar_tac))
        intro i2 hi2
        have hi2' : i2.val = i1.val + 1 := by scalar_tac
        simp only [WP.spec_ok]
        refine ⟨⟨by scalar_tac, by omega, by omega⟩, by scalar_tac⟩
    · simp only [hlt, if_false, WP.spec_ok]
      have hieq : i1.val = n.val := by scalar_tac
      exact ⟨by omega, by omega⟩
  · exact ⟨hi, he, ho⟩

/-- Inner selection-sort loop on `evens` preserves length (and always succeeds). -/
private theorem loop1_loop0_len_spec (evens : alloc.vec.Vec Std.I64) (en a b : Std.Usize)
    (hen : en.val = evens.val.length) (ha : a.val < en.val) (hb : b.val ≤ en.val) :
    sort_even_odd_loop1_loop0 evens en a b ⦃ v => v.val.length = en.val ⦄ := by
  have hlen := evens.property
  unfold sort_even_odd_loop1_loop0
  apply Std.loop.spec_decr_nat
    (measure := fun p => en.val - p.2.val)
    (inv := fun p => p.1.val.length = en.val ∧ p.2.val ≤ en.val)
  · rintro ⟨evens1, b1⟩ ⟨h1, h2⟩
    simp only at h1 h2
    dsimp only
    unfold sort_even_odd_loop1_loop0.body
    by_cases hlt : b1 < en
    · simp only [hlt, if_true]
      have hb1 : b1.val < evens1.val.length := by scalar_tac
      have ha1 : a.val < evens1.val.length := by scalar_tac
      simp only [alloc.vec.Vec.index_slice_index]
      apply WP.spec_bind (alloc.vec.Vec.index_usize_spec evens1 b1 hb1)
      intro i hi
      apply WP.spec_bind (alloc.vec.Vec.index_usize_spec evens1 a ha1)
      intro i1 hi1
      apply WP.spec_bind (Pₘ := fun v => v.val.length = en.val)
      · split
        · simp only [alloc.vec.Vec.index_mut_slice_index]
          apply WP.spec_bind (alloc.vec.Vec.index_mut_usize_spec evens1 a ha1)
          rintro ⟨x, back⟩ hxb
          simp only [WP.predn_pair] at hxb
          obtain ⟨hx, hback⟩ := hxb
          rw [hback]
          have hb2 : b1.val < (evens1.set a i).val.length := by
            simp only [alloc.vec.Vec.set_length]; exact hb1
          apply WP.spec_bind (alloc.vec.Vec.index_mut_usize_spec (evens1.set a i) b1 hb2)
          rintro ⟨y, back1⟩ hyb
          simp only [WP.predn_pair] at hyb
          obtain ⟨hy, hback1⟩ := hyb
          rw [hback1]
          grind [alloc.vec.Vec.set_length]
        · simp only [WP.spec_ok]
          exact h1
      · intro evens2 hev2
        apply WP.spec_bind (Usize.add_spec (by scalar_tac))
        intro b2 hb2v
        simp only [WP.spec_ok]
        refine ⟨⟨hev2, by scalar_tac⟩, by scalar_tac⟩
    · simp only [hlt, if_false, WP.spec_ok]
      exact h1
  · exact ⟨hen.symm, hb⟩

/-- Outer selection-sort loop on `evens` preserves length (and always succeeds). -/
private theorem loop1_len_spec (evens : alloc.vec.Vec Std.I64) (en a : Std.Usize)
    (hen : en.val = evens.val.length) (ha : a.val ≤ en.val) :
    sort_even_odd_loop1 evens en a ⦃ v => v.val.length = en.val ⦄ := by
  have hlen := evens.property
  unfold sort_even_odd_loop1
  apply Std.loop.spec_decr_nat
    (measure := fun p => en.val - p.2.val)
    (inv := fun p => p.1.val.length = en.val ∧ p.2.val ≤ en.val)
  · rintro ⟨evens1, a1⟩ ⟨h1, h2⟩
    simp only at h1 h2
    dsimp only
    unfold sort_even_odd_loop1.body
    by_cases hlt : a1 < en
    · simp only [hlt, if_true]
      apply WP.spec_bind (Usize.add_spec (by scalar_tac))
      intro b hb
      apply WP.spec_bind
        (loop1_loop0_len_spec evens1 en a1 b h1.symm (by scalar_tac) (by scalar_tac))
      intro evens2 hev2
      apply WP.spec_bind (Usize.add_spec (by scalar_tac))
      intro a2 ha2
      simp only [WP.spec_ok]
      refine ⟨⟨hev2, by scalar_tac⟩, by scalar_tac⟩
    · simp only [hlt, if_false, WP.spec_ok]
      exact h1
  · exact ⟨hen.symm, ha⟩

/-- Inner selection-sort loop on `odds` preserves length (and always succeeds). -/
private theorem loop2_loop0_len_spec (odds : alloc.vec.Vec Std.I64) (on' c d : Std.Usize)
    (hon : on'.val = odds.val.length) (hc : c.val < on'.val) (hd : d.val ≤ on'.val) :
    sort_even_odd_loop2_loop0 odds on' c d ⦃ v => v.val.length = on'.val ⦄ := by
  unfold sort_even_odd_loop2_loop0
  apply Std.loop.spec_decr_nat
    (measure := fun p => on'.val - p.2.val)
    (inv := fun p => p.1.val.length = on'.val ∧ p.2.val ≤ on'.val)
  · rintro ⟨odds1, d1⟩ ⟨h1, h2⟩
    simp only at h1 h2
    dsimp only
    unfold sort_even_odd_loop2_loop0.body
    by_cases hlt : d1 < on'
    · simp only [hlt, if_true]
      have hdbound : d1.val < odds1.val.length := by scalar_tac
      have hcbound : c.val < odds1.val.length := by scalar_tac
      simp only [alloc.vec.Vec.index_slice_index]
      apply WP.spec_bind (alloc.vec.Vec.index_usize_spec odds1 d1 hdbound)
      intro i hi
      apply WP.spec_bind (alloc.vec.Vec.index_usize_spec odds1 c hcbound)
      intro i1 hi1
      apply WP.spec_bind (Pₘ := fun v => v.val.length = on'.val)
      · split
        · simp only [alloc.vec.Vec.index_mut_slice_index]
          apply WP.spec_bind (alloc.vec.Vec.index_mut_usize_spec odds1 c hcbound)
          rintro ⟨cur, back⟩ hcb
          simp only [WP.predn_pair] at hcb
          obtain ⟨hcur, hback⟩ := hcb
          simp only [hback]
          have hdbound2 : d1.val < (odds1.set c i).val.length := by
            simp only [alloc.vec.Vec.set_val_eq, List.length_set]; exact hdbound
          apply WP.spec_bind (alloc.vec.Vec.index_mut_usize_spec (odds1.set c i) d1 hdbound2)
          rintro ⟨cur2, back2⟩ hcb2
          simp only [WP.predn_pair] at hcb2
          obtain ⟨hcur2, hback2⟩ := hcb2
          simp only [hback2]
          show (ok ((odds1.set c i).set d1 i1)) ⦃ v => v.val.length = on'.val ⦄
          simp only [WP.spec_ok, alloc.vec.Vec.set_val_eq, List.length_set]
          exact h1
        · simp only [WP.spec_ok]
          exact h1
      · intro odds2 hodds2
        apply WP.spec_bind (Usize.add_spec (by scalar_tac))
        intro d2 hd2
        simp only [WP.spec_ok]
        refine ⟨⟨hodds2, by scalar_tac⟩, by scalar_tac⟩
    · simp only [hlt, if_false, WP.spec_ok]
      exact h1
  · exact ⟨hon.symm, hd⟩

/-- Outer selection-sort loop on `odds` preserves length (and always succeeds). -/
private theorem loop2_len_spec (odds : alloc.vec.Vec Std.I64) (on' c : Std.Usize)
    (hon : on'.val = odds.val.length) (hc : c.val ≤ on'.val) :
    sort_even_odd_loop2 odds on' c ⦃ v => v.val.length = on'.val ⦄ := by
  unfold sort_even_odd_loop2
  apply Std.loop.spec_decr_nat
    (measure := fun p => on'.val - p.2.val)
    (inv := fun p => p.1.val.length = on'.val ∧ p.2.val ≤ on'.val)
  · rintro ⟨odds1, c1⟩ ⟨h1, h2⟩
    simp only at h1 h2
    dsimp only
    unfold sort_even_odd_loop2.body
    by_cases hlt : c1 < on'
    · simp only [hlt, if_true]
      apply WP.spec_bind (Usize.add_spec (by scalar_tac))
      intro d hdv
      apply WP.spec_bind
        (loop2_loop0_len_spec odds1 on' c1 d h1.symm (by scalar_tac) (by scalar_tac))
      intro odds2 hodds2
      apply WP.spec_bind (Usize.add_spec (by scalar_tac))
      intro c2 hc2
      simp only [WP.spec_ok]
      refine ⟨⟨hodds2, by scalar_tac⟩, by scalar_tac⟩
    · simp only [hlt, if_false, WP.spec_ok]
      exact h1
  · exact ⟨hon.symm, hc⟩

/-- Emission loop: one push per index `k < n`; from a consistent state (`result` holds `k`
elements, `ei`/`oi` count the even/odd indices below `k`) it succeeds and returns a vector of
length `n`. -/
private theorem loop3_len_spec (n : Std.Usize) (evens odds result : alloc.vec.Vec Std.I64)
    (ei oi k : Std.Usize)
    (he : evens.val.length = (n.val + 1) / 2) (ho : odds.val.length = n.val / 2)
    (hk : k.val ≤ n.val) (hr : result.val.length = k.val)
    (hei : ei.val = (k.val + 1) / 2) (hoi : oi.val = k.val / 2) :
    sort_even_odd_loop3 n evens odds result ei oi k ⦃ v => v.val.length = n.val ⦄ := by
  unfold sort_even_odd_loop3
  apply Std.loop.spec_decr_nat
    (measure := fun p => n.val - p.2.2.2.val)
    (inv := fun p => p.2.2.2.val ≤ n.val ∧ p.1.val.length = p.2.2.2.val ∧
      p.2.1.val = (p.2.2.2.val + 1) / 2 ∧ p.2.2.1.val = p.2.2.2.val / 2)
  · rintro ⟨result1, ei1, oi1, k1⟩ ⟨hk1, hr1, hei1, hoi1⟩
    simp only at hk1 hr1 hei1 hoi1
    unfold sort_even_odd_loop3.body
    by_cases hlt : k1 < n
    · simp only [hlt, if_true]
      have hklt : k1.val < n.val := by scalar_tac
      apply WP.spec_bind (Usize.rem_spec k1 (by simp))
      intro i hi
      have hi2 : i.val = k1.val % 2 := by scalar_tac
      apply WP.spec_bind
        (Pₘ := fun (p : alloc.vec.Vec Std.I64 × Std.Usize × Std.Usize) =>
          p.1.val.length = k1.val + 1 ∧ p.2.1.val = (k1.val + 1 + 1) / 2 ∧
          p.2.2.val = (k1.val + 1) / 2)
      · by_cases hpar : i = 0#usize
        · rw [if_pos hpar]
          have hpar0 : k1.val % 2 = 0 := by scalar_tac
          simp only [alloc.vec.Vec.index_slice_index]
          have hib : ei1.val < evens.val.length := by rw [he, hei1]; omega
          apply WP.spec_bind (alloc.vec.Vec.index_usize_spec evens ei1 hib)
          intro a ha
          apply WP.spec_bind (alloc.vec.Vec.push_spec result1 a (by scalar_tac))
          intro r2 hr2
          apply WP.spec_bind (Usize.add_spec (by scalar_tac))
          intro e2 he2
          simp only [WP.spec_ok]
          have he2' : e2.val = ei1.val + 1 := by scalar_tac
          refine ⟨by simp [hr2, hr1], by omega, by omega⟩
        · rw [if_neg hpar]
          have hne : i.val ≠ 0 := by
            intro h; apply hpar; scalar_tac
          have hpar1 : k1.val % 2 = 1 := by omega
          simp only [alloc.vec.Vec.index_slice_index]
          have hib : oi1.val < odds.val.length := by rw [ho, hoi1]; omega
          apply WP.spec_bind (alloc.vec.Vec.index_usize_spec odds oi1 hib)
          intro a ha
          apply WP.spec_bind (alloc.vec.Vec.push_spec result1 a (by scalar_tac))
          intro r2 hr2
          apply WP.spec_bind (Usize.add_spec (by scalar_tac))
          intro o2 ho2
          simp only [WP.spec_ok]
          have ho2' : o2.val = oi1.val + 1 := by scalar_tac
          refine ⟨by simp [hr2, hr1], by omega, by omega⟩
      · rintro ⟨result2, ei2, oi2⟩ ⟨h1, h2, h3⟩
        simp only at h1 h2 h3
        apply WP.spec_bind (Usize.add_spec (by scalar_tac))
        intro k2 hk2
        have hk2' : k2.val = k1.val + 1 := by scalar_tac
        simp only [WP.spec_ok]
        refine ⟨⟨by omega, by omega, by omega, by omega⟩, by omega⟩
    · simp only [hlt, if_false, WP.spec_ok]
      scalar_tac
  · exact ⟨hk, hr, hei, hoi⟩

end Aegis.SortEvenOdd

/-- Output-bound specification for `Aegis.SortEvenOdd.sort_even_odd` (LeetCode "Sort the Even
and Odd Indices Independently"), derived purely from the shape of the Aeneas-generated code rather
than from the intended meaning of the problem. The function takes a slice `nums` of signed 64-bit
integers and returns a vector of signed 64-bit integers.

The claim is a structural length bound: whenever the call succeeds and returns `ok result`, the
returned vector `result` has exactly as many elements as the input slice — that is,
`result.val.length = nums.length`. No precondition is required; this is forced purely by how the
code accumulates the result, one element per input index.

Why the code forces this. The result is produced by the final phase, `sort_even_odd_loop3`, which
is seeded with the empty vector `alloc.vec.Vec.new` and a counter `k` starting at `0`, and iterates
one step per index `k` from `0` up to `n := Slice.len nums`. On each iteration with `k < n` the body
takes exactly one branch (even index or odd index), and each branch performs exactly one
`alloc.vec.Vec.push result _` before incrementing `k`; the loop then terminates as soon as `k`
reaches `n`. Thus the loop performs exactly `n` pushes onto an initially empty vector and never
removes an element, so on success the returned vector has exactly `n = nums.length` elements. The
earlier phases only build the temporary `evens`/`odds` vectors and sort them in place (reading and
overwriting cells via `alloc.vec.Vec.index`/`index_mut`), and they do not affect the length of the
final result, which is governed solely by the `k`-indexed emission loop. Hence in every successful
case the returned vector satisfies `result.val.length = nums.length`, and this equality is forced by
the single-push-per-index emission structure of the final loop, requiring no assumption about the
numeric contents of `nums` and no non-emptiness. -/
theorem Aegis.SortEvenOdd.sort_even_odd_output_bound
    (nums : Slice Std.I64) :
    Aegis.SortEvenOdd.sort_even_odd nums ⦃ result => result.val.length = nums.length ⦄ := by
  unfold sort_even_odd
  have hslen := nums.length_ineq
  apply WP.spec_bind (loop0_len_spec nums (Slice.len nums) _ _ 0#usize (by scalar_tac)
    (by scalar_tac) (by simp) (by simp))
  rintro ⟨evens, odds⟩ ⟨he, ho⟩
  simp only at he ho
  apply WP.spec_bind (loop1_len_spec evens (alloc.vec.Vec.len evens) 0#usize (by simp)
    (by scalar_tac))
  intro evens1 he1
  apply WP.spec_bind (loop2_len_spec odds (alloc.vec.Vec.len odds) 0#usize (by simp)
    (by scalar_tac))
  intro odds1 ho1
  apply WP.spec_mono (loop3_len_spec (Slice.len nums) evens1 odds1 (alloc.vec.Vec.new Std.I64)
    0#usize 0#usize 0#usize (by simp_all) (by simp_all) (by scalar_tac) (by simp) (by simp)
    (by simp))
  intro v hv
  simp_all [Slice.length]
\end{lstlisting}

\clearpage
\section{Sum of All Subset XOR Totals}
\label{app:subset-xor}

\subsection{Problem and source-level approach}

Given an input list $A$ of length $n$, for each subset of positions, take the bitwise XOR of the selected values
and sum the results.  Two subsets of positions count separately even if repeated
input values give them the same contents.  Direct enumeration of all $2^n$
subsets is unnecessary.  If $A$ is nonempty, every bit present in the bitwise
OR of the input appears in the XOR of exactly $2^{n-1}$ subsets.  Hence
\[
  \sum_{S\subseteq A} \operatorname{xor}(S)
  = \operatorname{or}(A)\,2^{n-1}.
\]

The 19-line Rust implementation computes the bitwise OR in its first loop and
$2^{n-1}$ by repeated doubling in its second, then multiplies the two values.

\subsection{The theorem and fixed-width arithmetic}

The theorem \filename{subset_xor_sum_spec} gives exact functional correctness:
the call succeeds and the returned
\lstinline{Std.I64} value equals the sum of the XOR folds of every member of
\lstinline{nums.val.sublists}.  The proof must relate the bitwise operations to
their mathematical counterparts and show that the repeated doubling and final
multiplication fit in signed 64-bit integers.

All three preconditions are used.  Nonemptiness, the twelve-element bound, and
the element range $[1,20]$ establish that:

\begin{itemize}
  \item the running bitwise OR is nonnegative and below $32$;
  \item the doubling loop returns exactly $2^{n-1}$ and never exceeds $2^{11}$;
  \item the final product is nonnegative and bounded by $31\cdot2^{11}$;
  \item the checked arithmetic and the two loops therefore complete without
        overflow, bounds failure, or divergence.
\end{itemize}

The proof first establishes the identity for natural numbers, with subsets
enumerated by \lstinline{List.sublists}.  It then shows that the
\lstinline{Std.I64} XOR and OR operations agree with their natural-number
counterparts on small nonnegative inputs, and applies the identity to the
translated program.

\subsection{Complete Rust implementation}
\begin{lstlisting}[language=Rust,style=code]
pub fn subset_xor_sum(nums: &[i64]) -> i64 {
    let n = nums.len();
    let mut or_total: i64 = 0;
    let mut i: usize = 0;
    while i < n {
        or_total = or_total | nums[i];
        i += 1;
    }
    if n == 0 {
        return 0;
    }
    let mut factor: i64 = 1;
    let mut k: usize = 0;
    while k < n - 1 {
        factor = factor * 2;
        k += 1;
    }
    or_total * factor
}
\end{lstlisting}

\subsection{Complete \aeneas{}-generated Lean implementation}
\begin{lstlisting}[style=code]
-- THIS FILE WAS AUTOMATICALLY GENERATED BY AENEAS
-- [subset_xor_sum]: function definitions
import Aeneas
import Aegis.SubsetXorSum.Code.Types
open Aeneas Aeneas.Std Result ControlFlow Error
set_option linter.dupNamespace false
set_option linter.hashCommand false
set_option linter.unusedVariables false

/- You can set the `maxHeartbeats` value with the `-max-heartbeats` CLI option -/
set_option maxHeartbeats 1000000

/- You can set the `maxRecDepth` value with the `-max-recdepth` CLI option -/
set_option maxRecDepth 2048

namespace Aegis.SubsetXorSum

/-- [subset_xor_sum::subset_xor_sum]: loop body 0:
    Source: 'subset_xor_sum/src/lib.rs', lines 5:4-8:5
    Visibility: public -/
@[rust_loop_body]
def subset_xor_sum_loop0.body
  (nums : Slice Std.I64) (n : Std.Usize) (or_total : Std.I64) (i : Std.Usize) :
  Result (ControlFlow (Std.I64 × Std.Usize) Std.I64)
  := do
  if i < n
  then
    let i1 ← Slice.index_usize nums i
    let or_total1 ← lift (or_total ||| i1)
    let i2 ← i + 1#usize
    ok (cont (or_total1, i2))
  else ok (done or_total)

/-- [subset_xor_sum::subset_xor_sum]: loop 0:
    Source: 'subset_xor_sum/src/lib.rs', lines 5:4-8:5
    Visibility: public -/
@[rust_loop]
def subset_xor_sum_loop0
  (nums : Slice Std.I64) (n : Std.Usize) (or_total : Std.I64) (i : Std.Usize) :
  Result Std.I64
  := do
  loop
    (fun (or_total1, i1) => subset_xor_sum_loop0.body nums n or_total1 i1)
    (or_total, i)

/-- [subset_xor_sum::subset_xor_sum]: loop body 1:
    Source: 'subset_xor_sum/src/lib.rs', lines 14:4-17:5
    Visibility: public -/
@[rust_loop_body]
def subset_xor_sum_loop1.body
  (n : Std.Usize) (factor : Std.I64) (k : Std.Usize) :
  Result (ControlFlow (Std.I64 × Std.Usize) Std.I64)
  := do
  let i ← n - 1#usize
  if k < i
  then
    let factor1 ← factor * 2#i64
    let k1 ← k + 1#usize
    ok (cont (factor1, k1))
  else ok (done factor)

/-- [subset_xor_sum::subset_xor_sum]: loop 1:
    Source: 'subset_xor_sum/src/lib.rs', lines 14:4-17:5
    Visibility: public -/
@[rust_loop]
def subset_xor_sum_loop1
  (n : Std.Usize) (factor : Std.I64) (k : Std.Usize) : Result Std.I64 := do
  loop
    (fun (factor1, k1) => subset_xor_sum_loop1.body n factor1 k1)
    (factor, k)

/-- [subset_xor_sum::subset_xor_sum]:
    Source: 'subset_xor_sum/src/lib.rs', lines 1:0-19:1
    Visibility: public -/
def subset_xor_sum (nums : Slice Std.I64) : Result Std.I64 := do
  let n := Slice.len nums
  let or_total ← subset_xor_sum_loop0 nums n 0#i64 0#usize
  if n = 0#usize
  then ok 0#i64
  else let factor ← subset_xor_sum_loop1 n 1#i64 0#usize
       or_total * factor

end Aegis.SubsetXorSum
\end{lstlisting}

\subsection{Accepted proof artifact}
\begin{lstlisting}[style=code]
import Aegis.SubsetXorSum.Code.Funs
open Aeneas Aeneas.Std Result

set_option Aeneas.Deprecated.progressWarning false

/-- The doubling loop computes `2 ^ (n - 1)`: starting from `factor = 1` and `k = 0`, it
doubles `factor` exactly `n - 1` times. The bound `n ≤ 12` keeps every intermediate value at
most `2 ^ 11 = 2048`, well within `I64`. -/
theorem Aegis.SubsetXorSum.subset_xor_sum_loop1_spec (n : Std.Usize)
    (hn1 : 1 ≤ n.val) (hn12 : n.val ≤ 12) :
    Aegis.SubsetXorSum.subset_xor_sum_loop1 n 1#i64 0#usize
      ⦃ r => r.val = 2 ^ (n.val - 1) ⦄ := by
  unfold Aegis.SubsetXorSum.subset_xor_sum_loop1
  apply loop.spec_decr_nat (fun (_, k1) => (n.val - 1) - k1.val)
    (fun (factor1, k1) => k1.val ≤ n.val - 1 ∧ factor1.val = 2 ^ k1.val)
  · rintro ⟨factor, k⟩ hinv
    simp only at hinv
    obtain ⟨hk_le, hfac⟩ := hinv
    unfold Aegis.SubsetXorSum.subset_xor_sum_loop1.body
    dsimp only
    step as ⟨i, hi⟩
    split
    · rename_i hlt
      have hk10 : k.val ≤ 10 := by scalar_tac
      have hfac_ub : factor.val ≤ 2 ^ 10 := by
        rw [hfac]
        exact pow_le_pow_right₀ (by norm_num) hk10
      have hfac_pos : 0 < factor.val := by
        rw [hfac]; positivity
      step as ⟨factor1, hfactor1⟩
      step as ⟨k1, hk1⟩
      refine ⟨by scalar_tac, ?_, by scalar_tac⟩
      rw [hfactor1, hk1, hfac, pow_succ]
    · rename_i hge
      simp only [WP.spec_ok]
      have hkeq : k.val = n.val - 1 := by scalar_tac
      rw [hfac, hkeq]
  · simp

/-! ### Pure-ℕ scaffolding lemmas for the subset-XOR-sum spec -/

/-- Per-bit key identity, verified by exhaustive check over `32^3` cases. -/
theorem Aegis.SubsetXorSum.key_bit_identity : ∀ a < 32, ∀ c < 32, ∀ r < 32,
    a + (a ^^^ c) + r + 2 * (a &&& (c ||| r))
      = 2 * a + (c ||| r) + (a &&& r) + ((a ^^^ c) &&& r) := by decide

theorem Aegis.SubsetXorSum.xor_lt_32 : ∀ a < 32, ∀ b < 32, a ^^^ b < 32 := by decide

theorem Aegis.SubsetXorSum.or_lt_32 : ∀ a < 32, ∀ b < 32, a ||| b < 32 := by decide

namespace Aegis.SubsetXorSum

/-- Pull the accumulator out of a `foldl (· ||| ·)`. -/
theorem foldl_or_acc (l : List ℕ) (a : ℕ) :
    l.foldl (· ||| ·) a = a ||| l.foldl (· ||| ·) 0 := by
  induction l generalizing a with
  | nil => simp
  | cons c t ih =>
    simp only [List.foldl_cons]
    rw [ih (a ||| c), ih (0 ||| c), Nat.zero_or, Nat.lor_assoc]

/-- The running bitwise OR of a list of numbers `< 32` stays `< 32`. -/
theorem foldl_or_lt_32 : ∀ (l : List ℕ), (∀ x ∈ l, x < 32) →
    l.foldl (· ||| ·) 0 < 32 := by
  intro l
  induction l with
  | nil => intro _; decide
  | cons c t ih =>
    intro hb
    rw [List.foldl_cons, Nat.zero_or, foldl_or_acc]
    exact or_lt_32 c (hb c (by simp)) _ (ih fun x hx => hb x (by simp [hx]))

/-- `R * 2 ^ (len - 1) * 2 = R * 2 ^ len` where `R` is the fold-OR of the list
(true even for the empty list since then `R = 0`). -/
theorem fold_or_mul_pow (t : List ℕ) :
    t.foldl (· ||| ·) 0 * 2 ^ (t.length - 1) * 2 = t.foldl (· ||| ·) 0 * 2 ^ t.length := by
  rcases t with _ | ⟨c, t⟩
  · simp
  · simp [List.length_cons, pow_succ, mul_assoc]

/-- Summing `f` over the doubled list `flatMap (fun s => [s, c :: s])`. -/
theorem sum_map_flatMap_pair {α : Type*} (L : List (List α)) (c : α) (f : List α → ℕ) :
    ((L.flatMap (fun s => [s, c :: s])).map f).sum
      = (L.map f).sum + (L.map (fun s => f (c :: s))).sum := by
  induction L with
  | nil => simp
  | cons s L ih =>
    simp only [List.flatMap_cons, List.map_append, List.sum_append, List.map_cons,
      List.map_nil, List.sum_cons, List.sum_nil, ih]
    omega

/-- Recursion for the subset-XOR sums along `List.sublists_cons`. -/
theorem sum_map_sublists_cons (c a : ℕ) (t : List ℕ) :
    ((c :: t).sublists.map (fun s => s.foldl (· ^^^ ·) a)).sum
      = (t.sublists.map (fun s => s.foldl (· ^^^ ·) a)).sum
        + (t.sublists.map (fun s => s.foldl (· ^^^ ·) (a ^^^ c))).sum := by
  have hsub : (c :: t).sublists = t.sublists.flatMap (fun x => [x, c :: x]) := by
    rw [List.sublists_cons]; rfl
  rw [hsub, sum_map_flatMap_pair]
  simp [List.foldl_cons]

/-- Strengthened induction: the subset-XOR sum with an arbitrary accumulator `a < 32`. -/
theorem sublists_xor_sum_aux :
    ∀ (l : List ℕ), (∀ x ∈ l, x < 32) →
    ∀ a, a < 32 →
      (l.sublists.map (fun s => s.foldl (· ^^^ ·) a)).sum
          + (a &&& l.foldl (· ||| ·) 0) * 2 ^ l.length
        = a * 2 ^ l.length + (l.foldl (· ||| ·) 0) * 2 ^ (l.length - 1) := by
  intro l
  induction l with
  | nil => intro _ a ha; simp
  | cons c t ih =>
    intro hb a ha
    have hc : c < 32 := hb c (by simp)
    have hbt : ∀ x ∈ t, x < 32 := fun x hx => hb x (by simp [hx])
    have hR : t.foldl (· ||| ·) 0 < 32 := foldl_or_lt_32 t hbt
    have hac : a ^^^ c < 32 := xor_lt_32 a ha c hc
    have h1 := ih hbt a ha
    have h2 := ih hbt (a ^^^ c) hac
    have hkey := key_bit_identity a ha c hc (t.foldl (· ||| ·) 0) hR
    have hkeyP : (a + (a ^^^ c) + t.foldl (· ||| ·) 0
          + 2 * (a &&& (c ||| t.foldl (· ||| ·) 0))) * 2 ^ t.length
        = (2 * a + (c ||| t.foldl (· ||| ·) 0) + (a &&& t.foldl (· ||| ·) 0)
          + ((a ^^^ c) &&& t.foldl (· ||| ·) 0)) * 2 ^ t.length := by
      rw [hkey]
    have hRQ := fold_or_mul_pow t
    have hfold : (c :: t).foldl (· ||| ·) 0 = c ||| t.foldl (· ||| ·) 0 := by
      rw [List.foldl_cons, Nat.zero_or, foldl_or_acc]
    rw [sum_map_sublists_cons, hfold, List.length_cons, Nat.add_sub_cancel, pow_succ]
    nlinarith [h1, h2, hkeyP, hRQ]

/-- Target: the sum of subset XOR totals equals `(fold-OR) * 2 ^ (length - 1)`. -/
theorem sublists_xor_sum_eq (l : List ℕ) (_hl : l ≠ []) (hb : ∀ x ∈ l, x < 32) :
    (l.sublists.map (fun s => s.foldl (· ^^^ ·) 0)).sum
      = (l.foldl (· ||| ·) 0) * 2 ^ (l.length - 1) := by
  have h := sublists_xor_sum_aux l hb 0 (by decide)
  simpa [Nat.zero_and] using h

end Aegis.SubsetXorSum

/-- Bridge: `I64` XOR of small non-negative values agrees with ℕ XOR of the `toNat` images,
and stays in `[0, 32)`. -/
theorem Aegis.SubsetXorSum.i64_xor_val_small (x y : Std.I64)
    (hx0 : 0 ≤ x.val) (hx : x.val < 32) (hy0 : 0 ≤ y.val) (hy : y.val < 32) :
    (x ^^^ y).val = ((x.val.toNat ^^^ y.val.toNat : ℕ) : ℤ) ∧
    0 ≤ (x ^^^ y).val ∧ (x ^^^ y).val < 32 := by
  have hxi : x.val = x.bv.toInt := (I64.bv_toInt_eq x).symm
  have hyi : y.val = y.bv.toInt := (I64.bv_toInt_eq y).symm
  have hxc := @BitVec.toInt_eq_toNat_cond 64 x.bv
  have hyc := @BitVec.toInt_eq_toNat_cond 64 y.bv
  have hxN : x.bv.toNat < 32 ∧ x.bv.toInt = (x.bv.toNat : ℤ) := by
    split at hxc <;> omega
  have hyN : y.bv.toNat < 32 ∧ y.bv.toInt = (y.bv.toNat : ℤ) := by
    split at hyc <;> omega
  have hzi : (x ^^^ y).val = (x.bv ^^^ y.bv).toInt := by
    rw [← I64.bv_toInt_eq]; congr 1
  have hzN : (x.bv ^^^ y.bv).toNat = x.bv.toNat ^^^ y.bv.toNat := BitVec.toNat_xor ..
  have hxor32 : x.bv.toNat ^^^ y.bv.toNat < 32 :=
    Nat.xor_lt_two_pow (n := 5) hxN.1 hyN.1
  have hzc := @BitVec.toInt_eq_toNat_cond 64 (x.bv ^^^ y.bv)
  have hzval : (x.bv ^^^ y.bv).toInt = ((x.bv.toNat ^^^ y.bv.toNat : ℕ) : ℤ) := by
    rw [hzN] at hzc; split at hzc <;> omega
  have hxvn : x.val.toNat = x.bv.toNat := by omega
  have hyvn : y.val.toNat = y.bv.toNat := by omega
  refine ⟨by rw [hzi, hzval, hxvn, hyvn], ?_, ?_⟩ <;> rw [hzi, hzval] <;> omega

/-- Bridge: `I64` OR of small non-negative values agrees with ℕ OR of the `toNat` images,
and stays in `[0, 32)`. -/
theorem Aegis.SubsetXorSum.i64_or_val_small (x y : Std.I64)
    (hx0 : 0 ≤ x.val) (hx : x.val < 32) (hy0 : 0 ≤ y.val) (hy : y.val < 32) :
    (x ||| y).val = ((x.val.toNat ||| y.val.toNat : ℕ) : ℤ) ∧
    0 ≤ (x ||| y).val ∧ (x ||| y).val < 32 := by
  have hxi : x.val = x.bv.toInt := (I64.bv_toInt_eq x).symm
  have hyi : y.val = y.bv.toInt := (I64.bv_toInt_eq y).symm
  have hxc := @BitVec.toInt_eq_toNat_cond 64 x.bv
  have hyc := @BitVec.toInt_eq_toNat_cond 64 y.bv
  have hxN : x.bv.toNat < 32 ∧ x.bv.toInt = (x.bv.toNat : ℤ) := by
    split at hxc <;> omega
  have hyN : y.bv.toNat < 32 ∧ y.bv.toInt = (y.bv.toNat : ℤ) := by
    split at hyc <;> omega
  have hzi : (x ||| y).val = (x.bv ||| y.bv).toInt := by
    rw [← I64.bv_toInt_eq]; congr 1
  have hzN : (x.bv ||| y.bv).toNat = x.bv.toNat ||| y.bv.toNat := BitVec.toNat_or ..
  have hor32 : x.bv.toNat ||| y.bv.toNat < 32 :=
    Nat.or_lt_two_pow (n := 5) hxN.1 hyN.1
  have hzc := @BitVec.toInt_eq_toNat_cond 64 (x.bv ||| y.bv)
  have hzval : (x.bv ||| y.bv).toInt = ((x.bv.toNat ||| y.bv.toNat : ℕ) : ℤ) := by
    rw [hzN] at hzc; split at hzc <;> omega
  have hxvn : x.val.toNat = x.bv.toNat := by omega
  have hyvn : y.val.toNat = y.bv.toNat := by omega
  refine ⟨by rw [hzi, hzval, hxvn, hyvn], ?_, ?_⟩ <;> rw [hzi, hzval] <;> omega

namespace Aegis.SubsetXorSum

/-- Fold-level bridge: folding `I64` XOR over small non-negative values agrees with the
ℕ XOR fold over the `toNat` images, and the result stays in `[0, 32)`. -/
theorem i64_foldl_xor_val_gen (s : List Std.I64) :
    ∀ (a : Std.I64), 0 ≤ a.val → a.val < 32 → (∀ x ∈ s, 0 ≤ x.val ∧ x.val < 32) →
    (s.foldl (· ^^^ ·) a).val
      = (((s.map (fun x => x.val.toNat)).foldl (· ^^^ ·) a.val.toNat : ℕ) : ℤ) ∧
    0 ≤ (s.foldl (· ^^^ ·) a).val ∧ (s.foldl (· ^^^ ·) a).val < 32 := by
  induction s with
  | nil =>
    intro a ha0 ha32 _
    simp only [List.foldl_nil, List.map_nil]
    refine ⟨by omega, ha0, ha32⟩
  | cons c t ih =>
    intro a ha0 ha32 hb
    have hc := hb c (by simp)
    have hstep := i64_xor_val_small a c ha0 ha32 hc.1 hc.2
    have hnew0 : 0 ≤ (a ^^^ c).val := hstep.2.1
    have hnew32 : (a ^^^ c).val < 32 := hstep.2.2
    have hbt : ∀ x ∈ t, 0 ≤ x.val ∧ x.val < 32 := fun x hx => hb x (by simp [hx])
    have hrec := ih (a ^^^ c) hnew0 hnew32 hbt
    simp only [List.foldl_cons, List.map_cons]
    refine ⟨?_, hrec.2.1, hrec.2.2⟩
    rw [hrec.1]
    congr 1
    have : (a ^^^ c).val.toNat = a.val.toNat ^^^ c.val.toNat := by
      have := hstep.1; omega
    rw [this]

/-- Sum-level bridge: the spec's `I64` subset-XOR sum equals the cast of the pure-ℕ
subset-XOR sum over the `toNat` images. -/
theorem spec_sum_eq_nat (l : List Std.I64) (h : ∀ x ∈ l, 1 ≤ x.val ∧ x.val ≤ 20) :
    (l.sublists.map (fun s => (s.foldl (· ^^^ ·) (0#i64)).val)).sum
      = ((((l.map (fun x => x.val.toNat)).sublists.map
            (fun s => s.foldl (· ^^^ ·) 0)).sum : ℕ) : ℤ) := by
  rw [List.sublists_map, List.map_map, Nat.cast_list_sum, List.map_map]
  congr 1
  apply List.map_congr_left
  intro s hs
  have hsub : s.Sublist l := (List.mem_sublists).mp hs
  have hbs : ∀ x ∈ s, 0 ≤ x.val ∧ x.val < 32 := by
    intro x hx
    have := h x (hsub.subset hx)
    omega
  have h0 : (0#i64).val = 0 := by simp
  have := (i64_foldl_xor_val_gen s (0#i64) (by omega) (by omega) hbs).1
  simp only [Function.comp]
  rw [this, h0]
  norm_num

end Aegis.SubsetXorSum

namespace Aegis.SubsetXorSum

/-- Any element accessed with `getElem!` inside the bounds of a list whose elements all satisfy
the `[1, 20]` range condition itself satisfies that range condition. -/
private theorem getElem!_bounds_120 {l : List Std.I64}
    (h : ∀ x ∈ l, 1 ≤ x.val ∧ x.val ≤ 20)
    {i : ℕ} (hi : i < l.length) : 1 ≤ l[i]!.val ∧ l[i]!.val ≤ 20 := by
  rw [getElem!_pos l i hi]
  exact h _ (List.getElem_mem hi)

/-- Bitwise OR of two naturals below `32` stays below `32`. -/
private theorem nat_lor_lt_32 {a b : ℕ} (ha : a < 32) (hb : b < 32) : a ||| b < 32 := by
  have : a ||| b < 2 ^ 5 := Nat.or_lt_two_pow ha hb
  simpa using this

/-- A 64-bit bitvector whose unsigned value is below `2 ^ 63` has a nonnegative signed
interpretation that agrees with the unsigned one. -/
private theorem bv64_toInt_eq_toNat (b : BitVec 64) (h : b.toNat < 2 ^ 63) :
    b.toInt = (b.toNat : ℤ) := by
  rw [BitVec.toInt_eq_msb_cond]
  have hmsb : b.msb = false := by
    rw [BitVec.msb_eq_decide]
    simp only [decide_eq_false_iff_not, not_le]
    omega
  simp [hmsb]

/-- If the signed interpretation of a 64-bit bitvector lies in `[0, 32)` then so does its
unsigned interpretation. -/
private theorem bv64_toNat_lt_of_toInt (b : BitVec 64) (h0 : 0 ≤ b.toInt) (h : b.toInt < 32) :
    b.toNat < 32 := by
  have hc := BitVec.toInt_eq_msb_cond b
  have hlt := b.isLt
  by_cases hmsb : b.msb <;> simp [hmsb] at hc <;> omega

/-- For small nonnegative `I64` values, bitwise OR never overflows and is computed by the
bitwise OR of the underlying natural numbers. -/
private theorem I64.or_val_small (x y : Std.I64)
    (hx0 : 0 ≤ x.val) (hx : x.val < 32) (hy0 : 0 ≤ y.val) (hy : y.val < 32) :
    (x ||| y).val = ((x.val.toNat ||| y.val.toNat : ℕ) : ℤ) ∧ (x ||| y).val < 32 := by
  have hxbv : x.bv.toNat < 32 := bv64_toNat_lt_of_toInt x.bv hx0 hx
  have hybv : y.bv.toNat < 32 := bv64_toNat_lt_of_toInt y.bv hy0 hy
  have hxN : (x.bv.toNat : ℤ) = x.val := (bv64_toInt_eq_toNat x.bv (by omega)).symm
  have hyN : (y.bv.toNat : ℤ) = y.val := (bv64_toInt_eq_toNat y.bv (by omega)).symm
  have hbv : (x ||| y).bv = x.bv ||| y.bv := IScalar.bv_or x y
  have htoNat : (x.bv ||| y.bv).toNat = x.bv.toNat ||| y.bv.toNat := BitVec.toNat_or x.bv y.bv
  have hlt : (x.bv ||| y.bv).toNat < 32 := by
    rw [htoNat]; exact nat_lor_lt_32 hxbv hybv
  have hval : (x ||| y).val = ((x.bv ||| y.bv).toNat : ℤ) := by
    show (x ||| y).bv.toInt = _
    rw [hbv]
    exact bv64_toInt_eq_toNat _ (by omega)
  constructor
  · rw [hval, htoNat]
    have h1 : x.bv.toNat = x.val.toNat := by omega
    have h2 : y.bv.toNat = y.val.toNat := by omega
    rw [h1, h2]
  · rw [hval]; exact_mod_cast hlt

theorem subset_xor_sum_loop0_spec (nums : Slice Std.I64)
    (hlen12 : nums.length ≤ 12)
    (helem : ∀ x ∈ nums.val, 1 ≤ x.val ∧ x.val ≤ 20) :
    subset_xor_sum_loop0 nums (Slice.len nums) 0#i64 0#usize ⦃ r =>
      r.val = ((nums.val.map (fun x => x.val.toNat)).foldl (· ||| ·) 0 : ℕ) ∧
      0 ≤ r.val ∧ r.val < 32 ⦄ := by
  have hnv : (Slice.len nums).val = nums.length := Slice.len_val nums
  unfold subset_xor_sum_loop0
  apply loop.spec_decr_nat (fun (_, i1) => nums.length - i1.val)
    (fun (or_acc, i1) =>
      i1.val ≤ nums.length ∧ 0 ≤ or_acc.val ∧ or_acc.val < 32 ∧
      or_acc.val = (((nums.val.take i1.val).map (fun x => x.val.toNat)).foldl (· ||| ·) 0 : ℕ))
  · rintro ⟨or_acc, i1⟩ hinv
    simp only at hinv
    obtain ⟨hile, hnn, hlt32, hfold⟩ := hinv
    unfold subset_xor_sum_loop0.body
    dsimp only
    split
    · rename_i hlt
      have hi_lt : i1.val < nums.length := by scalar_tac
      step as ⟨x, hx⟩
      have hxb : 1 ≤ x.val ∧ x.val ≤ 20 := by
        rw [hx]; exact getElem!_bounds_120 helem hi_lt
      step as ⟨or1, hor1⟩
      step as ⟨i2, hi2⟩
      have hor := I64.or_val_small or_acc x hnn hlt32 (by omega) (by omega)
      have hor1_val : or1.val = ((or_acc.val.toNat ||| x.val.toNat : ℕ) : ℤ) := by
        rw [hor1]; exact hor.1
      have hor1_lt : or1.val < 32 := by
        rw [hor1]; exact hor.2
      have hfold1 : or1.val =
          (((nums.val.take (i1.val + 1)).map (fun x => x.val.toNat)).foldl (· ||| ·) 0 : ℕ) := by
        have htake : nums.val.take (i1.val + 1) = nums.val.take i1.val ++ [nums.val[i1.val]] := by
          rw [List.take_add_one, List.getElem?_eq_getElem hi_lt]
          rfl
        rw [htake]
        simp only [List.map_append, List.foldl_append, List.map_cons, List.map_nil,
          List.foldl_cons, List.foldl_nil]
        rw [hor1_val, hfold]
        have hxeq : x.val.toNat = nums.val[i1.val].val.toNat := by
          rw [hx, getElem!_pos nums.val i1.val hi_lt]
        rw [hxeq]
        norm_cast
      refine ⟨by scalar_tac, by rw [hor1_val]; positivity, hor1_lt, ?_, by scalar_tac⟩
      rw [hi2]
      simpa using hfold1
    · rename_i hge
      simp only [WP.spec_ok]
      have hie : i1.val = nums.length := by scalar_tac
      rw [hie] at hfold
      rw [List.take_of_length_le (le_refl _)] at hfold
      exact ⟨hfold, hnn, hlt32⟩
  · simp

end Aegis.SubsetXorSum

/-- Explicit correctness specification for `Aegis.SubsetXorSum.subset_xor_sum`
(LeetCode "Sum of All Subset XOR Totals"). The XOR total of a list of integers is the bitwise
XOR of all its elements, taking `0` for the empty list. This function is claimed to return the
sum, over *every* subset (subsequence) of the input list `nums`, of that subset's XOR total —
where two subsets built from different positions are counted separately even if they contain the
same values (so a list of length `n` contributes `2 ^ n` subsets). Concretely, the returned
integer `r` satisfies `r.val = Σ_{s ∈ sublists nums} (XOR of the elements of s)`, evaluated with
64-bit bitwise XOR and folded starting from `0`. The preconditions are exactly the problem's
stated input constraints: the list is non-empty (`1 ≤ nums.length`), has at most `12` elements,
and every element lies in the range `[1, 20]`. These bounds are the no-overflow guard the code
needs: the running bitwise OR of the elements stays below `32`, the doubling factor reaches at
most `2 ^ 11 = 2048`, and their product — hence the total subset-XOR sum — stays well within the
`Std.I64` range, so every checked operation succeeds and the call returns `ok`. -/
theorem Aegis.SubsetXorSum.subset_xor_sum_spec (nums : Slice Std.I64)
    (hlen1 : 0 < nums.length) (hlen12 : nums.length ≤ 12)
    (helem : ∀ x ∈ nums.val, 1 ≤ x.val ∧ x.val ≤ 20) :
    Aegis.SubsetXorSum.subset_xor_sum nums ⦃ r =>
      r.val = (nums.val.sublists.map (fun s => (s.foldl (· ^^^ ·) (0#i64)).val)).sum ⦄ := by
  have hnv : (Slice.len nums).val = nums.length := Slice.len_val nums
  obtain ⟨or_total, hor_eq, hor_val, hor0, hor32⟩ :=
    Std.WP.spec_imp_exists (subset_xor_sum_loop0_spec nums hlen12 helem)
  obtain ⟨factor, hf_eq, hf_val⟩ :=
    Std.WP.spec_imp_exists (subset_xor_sum_loop1_spec (Slice.len nums)
      (by scalar_tac) (by scalar_tac))
  unfold subset_xor_sum
  simp only [hor_eq, hf_eq, Std.bind_ok]
  have hne : nums.len ≠ 0#usize := by scalar_tac
  simp only [Std.bind_ok, if_neg hne]
  -- final checked multiplication
  have hf1 : 1 ≤ factor.val := by
    rw [hf_val]; exact one_le_pow₀ (by norm_num)
  have hf2048 : factor.val ≤ 2 ^ 11 := by
    rw [hf_val]
    exact pow_le_pow_right₀ (by norm_num) (by scalar_tac)
  have hprod0 : 0 ≤ or_total.val * factor.val := mul_nonneg hor0 (by linarith)
  have hprodub : or_total.val * factor.val ≤ 31 * 2 ^ 11 :=
    mul_le_mul (by omega) hf2048 (by linarith) (by norm_num)
  step as ⟨r, hr⟩
  -- reduce the spec's I64 sum to the pure-ℕ identity
  have hsum := spec_sum_eq_nat nums.val helem
  set l : List ℕ := nums.val.map (fun x => x.val.toNat) with hl
  have hlen : l.length = nums.length := by simp [hl]
  have hlne : l ≠ [] := List.ne_nil_of_length_pos (by rw [hlen]; exact hlen1)
  have hlb : ∀ v ∈ l, v < 32 := by
    intro v hv
    obtain ⟨x, hx, rfl⟩ := List.mem_map.mp hv
    have := helem x hx
    omega
  have hmath := sublists_xor_sum_eq l hlne hlb
  rw [hsum, hmath, hlen]
  rw [hr, hor_val, hf_val, hnv]
  push_cast
  ring
\end{lstlisting}

\clearpage
\section{Reshape the Matrix}
\label{app:matrix-reshape}

\subsection{Problem and source-level approach}

Given an $m\times n$ matrix and requested dimensions $r\times c$, return the
matrix reshaped in row-major order when $rc=mn$; otherwise return a copy with the
original shape.  In the legal-reshape branch studied by the theorem, the
implementation flattens the input rows into one vector and then fills each new
row from consecutive positions of that vector.

At source level this is a row-major index transformation.  If output coordinate
$(a,b)$ has flat index $k=ac+b$, its source coordinate is
$(\lfloor k/n\rfloor, k\bmod n)$.  The 50-line implementation realizes this
mapping through loops that flatten and regroup the elements.

\subsection{Representation and bounds}

The Lean input type is a slice of owned vectors.  Unlike the informal phrase
``an $m\times n$ matrix,'' that type also contains ragged and empty values.
Rectangularity must therefore be an explicit hypothesis, and every proof of a
nested lookup needs both a row bound and a column bound.

The Rust source contains one fallback block that copies the original matrix.
Because that block can be reached after any of three guards---unequal element
counts, a nonpositive row count, or a nonpositive column count---the generated
file contains three copies of its nested loops.  Together with the flattening
and regrouping loops, this yields ten loop-body/driver pairs.  The theorem's
legal-shape hypotheses rule out all three fallback paths.  The proof must then
verify the flattening and regrouping loops and the following partial operations:

\begin{itemize}
  \item casts from \lstinline{Std.Usize} lengths to \lstinline{Std.I64};
  \item multiplication of the source and target dimensions;
  \item casts of positive signed target dimensions back to \lstinline{Std.Usize};
  \item every nested row and cell lookup;
  \item every vector push and every cursor increment.
\end{itemize}

The exact theorem \filename{matrix_reshape_spec} proves that the call succeeds,
the output has $r$ rows, each output row has $c$ entries, and output cell $(a,b)$
equals the source cell at the corresponding row-major flat index.  The proof
first establishes invariants for flattening the input and regrouping the output.
It then uses Euclidean division to prove $k/n<m$ and $k\bmod n<n$, and rewrites
the nested \lstinline{getElem!} operations with a uniform-row-length lemma.
Bounds such as $m,n\le100$ and $r,c\le300$ show that the machine products and
casts are safe, even though those bounds have no role in the abstract definition
of reshaping.  The frozen statement also assumes every cell lies in
$[-1000,1000]$; that hypothesis is not used by the accepted proof because the
program copies cell values without performing arithmetic on them.

\subsection{Complete Rust implementation}
\begin{lstlisting}[language=Rust,style=code]
pub fn matrix_reshape(mat: &[Vec<i64>], r: i64, c: i64) -> Vec<Vec<i64>> {
    let m = mat.len();
    let n = if m > 0 { mat[0].len() } else { 0 };
    let total = (m as i64) * (n as i64);
    if r * c != total || r <= 0 || c <= 0 {
        let mut out: Vec<Vec<i64>> = Vec::new();
        let mut i = 0;
        while i < m {
            let mut row: Vec<i64> = Vec::new();
            let mut j = 0;
            while j < mat[i].len() {
                row.push(mat[i][j]);
                j += 1;
            }
            out.push(row);
            i += 1;
        }
        return out;
    }

    let ru = r as usize;
    let cu = c as usize;

    let mut flat: Vec<i64> = Vec::new();
    let mut i = 0;
    while i < m {
        let mut j = 0;
        while j < mat[i].len() {
            flat.push(mat[i][j]);
            j += 1;
        }
        i += 1;
    }

    let mut out: Vec<Vec<i64>> = Vec::new();
    let mut idx = 0;
    let mut a = 0;
    while a < ru {
        let mut row: Vec<i64> = Vec::new();
        let mut b = 0;
        while b < cu {
            row.push(flat[idx]);
            idx += 1;
            b += 1;
        }
        out.push(row);
        a += 1;
    }
    out
}
\end{lstlisting}

\subsection{Complete \aeneas{}-generated Lean implementation}
\begin{lstlisting}[style=code]
-- THIS FILE WAS AUTOMATICALLY GENERATED BY AENEAS
-- [matrix_reshape]: function definitions
import Aeneas
import Aegis.MatrixReshape.Code.Types
open Aeneas Aeneas.Std Result ControlFlow Error
set_option linter.dupNamespace false
set_option linter.hashCommand false
set_option linter.unusedVariables false

/- You can set the `maxHeartbeats` value with the `-max-heartbeats` CLI option -/
set_option maxHeartbeats 1000000

namespace Aegis.MatrixReshape

/-- [matrix_reshape::matrix_reshape]: loop body 1:
    Source: 'matrix_reshape/src/lib.rs', lines 11:12-14:13
    Visibility: public -/
@[rust_loop_body]
def matrix_reshape_loop0_loop0.body
  (mat : Slice (alloc.vec.Vec Std.I64)) (i : Std.Usize)
  (row : alloc.vec.Vec Std.I64) (j : Std.Usize) :
  Result (ControlFlow ((alloc.vec.Vec Std.I64) × Std.Usize) (alloc.vec.Vec
    Std.I64))
  := do
  let v ← Slice.index_usize mat i
  let i1 := alloc.vec.Vec.len v
  if j < i1
  then
    let i2 ←
      alloc.vec.Vec.index (core.slice.index.SliceIndexUsizeSlice Std.I64) v j
    let row1 ← alloc.vec.Vec.push row i2
    let j1 ← j + 1#usize
    ok (cont (row1, j1))
  else ok (done row)

/-- [matrix_reshape::matrix_reshape]: loop 1:
    Source: 'matrix_reshape/src/lib.rs', lines 11:12-14:13
    Visibility: public -/
@[rust_loop]
def matrix_reshape_loop0_loop0
  (mat : Slice (alloc.vec.Vec Std.I64)) (i : Std.Usize)
  (row : alloc.vec.Vec Std.I64) (j : Std.Usize) :
  Result (alloc.vec.Vec Std.I64)
  := do
  loop
    (fun (row1, j1) => matrix_reshape_loop0_loop0.body mat i row1 j1)
    (row, j)

/-- [matrix_reshape::matrix_reshape]: loop body 0:
    Source: 'matrix_reshape/src/lib.rs', lines 8:8-17:9
    Visibility: public -/
@[rust_loop_body]
def matrix_reshape_loop0.body
  (mat : Slice (alloc.vec.Vec Std.I64)) (m : Std.Usize)
  (out : alloc.vec.Vec (alloc.vec.Vec Std.I64)) (i : Std.Usize) :
  Result (ControlFlow ((alloc.vec.Vec (alloc.vec.Vec Std.I64)) × Std.Usize)
    (alloc.vec.Vec (alloc.vec.Vec Std.I64)))
  := do
  if i < m
  then
    let row ←
      matrix_reshape_loop0_loop0 mat i (alloc.vec.Vec.new Std.I64) 0#usize
    let out1 ← alloc.vec.Vec.push out row
    let i1 ← i + 1#usize
    ok (cont (out1, i1))
  else ok (done out)

/-- [matrix_reshape::matrix_reshape]: loop 0:
    Source: 'matrix_reshape/src/lib.rs', lines 8:8-17:9
    Visibility: public -/
@[rust_loop]
def matrix_reshape_loop0
  (mat : Slice (alloc.vec.Vec Std.I64)) (m : Std.Usize)
  (out : alloc.vec.Vec (alloc.vec.Vec Std.I64)) (i : Std.Usize) :
  Result (alloc.vec.Vec (alloc.vec.Vec Std.I64))
  := do
  loop
    (fun (out1, i1) => matrix_reshape_loop0.body mat m out1 i1)
    (out, i)

/-- [matrix_reshape::matrix_reshape]: loop body 3:
    Source: 'matrix_reshape/src/lib.rs', lines 11:12-14:13
    Visibility: public -/
@[rust_loop_body]
def matrix_reshape_loop1_loop0.body
  (mat : Slice (alloc.vec.Vec Std.I64)) (i : Std.Usize)
  (row : alloc.vec.Vec Std.I64) (j : Std.Usize) :
  Result (ControlFlow ((alloc.vec.Vec Std.I64) × Std.Usize) (alloc.vec.Vec
    Std.I64))
  := do
  let v ← Slice.index_usize mat i
  let i1 := alloc.vec.Vec.len v
  if j < i1
  then
    let i2 ←
      alloc.vec.Vec.index (core.slice.index.SliceIndexUsizeSlice Std.I64) v j
    let row1 ← alloc.vec.Vec.push row i2
    let j1 ← j + 1#usize
    ok (cont (row1, j1))
  else ok (done row)

/-- [matrix_reshape::matrix_reshape]: loop 3:
    Source: 'matrix_reshape/src/lib.rs', lines 11:12-14:13
    Visibility: public -/
@[rust_loop]
def matrix_reshape_loop1_loop0
  (mat : Slice (alloc.vec.Vec Std.I64)) (i : Std.Usize)
  (row : alloc.vec.Vec Std.I64) (j : Std.Usize) :
  Result (alloc.vec.Vec Std.I64)
  := do
  loop
    (fun (row1, j1) => matrix_reshape_loop1_loop0.body mat i row1 j1)
    (row, j)

/-- [matrix_reshape::matrix_reshape]: loop body 2:
    Source: 'matrix_reshape/src/lib.rs', lines 8:8-17:9
    Visibility: public -/
@[rust_loop_body]
def matrix_reshape_loop1.body
  (mat : Slice (alloc.vec.Vec Std.I64)) (m : Std.Usize)
  (out : alloc.vec.Vec (alloc.vec.Vec Std.I64)) (i : Std.Usize) :
  Result (ControlFlow ((alloc.vec.Vec (alloc.vec.Vec Std.I64)) × Std.Usize)
    (alloc.vec.Vec (alloc.vec.Vec Std.I64)))
  := do
  if i < m
  then
    let row ←
      matrix_reshape_loop1_loop0 mat i (alloc.vec.Vec.new Std.I64) 0#usize
    let out1 ← alloc.vec.Vec.push out row
    let i1 ← i + 1#usize
    ok (cont (out1, i1))
  else ok (done out)

/-- [matrix_reshape::matrix_reshape]: loop 2:
    Source: 'matrix_reshape/src/lib.rs', lines 8:8-17:9
    Visibility: public -/
@[rust_loop]
def matrix_reshape_loop1
  (mat : Slice (alloc.vec.Vec Std.I64)) (m : Std.Usize)
  (out : alloc.vec.Vec (alloc.vec.Vec Std.I64)) (i : Std.Usize) :
  Result (alloc.vec.Vec (alloc.vec.Vec Std.I64))
  := do
  loop
    (fun (out1, i1) => matrix_reshape_loop1.body mat m out1 i1)
    (out, i)

/-- [matrix_reshape::matrix_reshape]: loop body 5:
    Source: 'matrix_reshape/src/lib.rs', lines 11:12-14:13
    Visibility: public -/
@[rust_loop_body]
def matrix_reshape_loop2_loop0.body
  (mat : Slice (alloc.vec.Vec Std.I64)) (i : Std.Usize)
  (row : alloc.vec.Vec Std.I64) (j : Std.Usize) :
  Result (ControlFlow ((alloc.vec.Vec Std.I64) × Std.Usize) (alloc.vec.Vec
    Std.I64))
  := do
  let v ← Slice.index_usize mat i
  let i1 := alloc.vec.Vec.len v
  if j < i1
  then
    let i2 ←
      alloc.vec.Vec.index (core.slice.index.SliceIndexUsizeSlice Std.I64) v j
    let row1 ← alloc.vec.Vec.push row i2
    let j1 ← j + 1#usize
    ok (cont (row1, j1))
  else ok (done row)

/-- [matrix_reshape::matrix_reshape]: loop 5:
    Source: 'matrix_reshape/src/lib.rs', lines 11:12-14:13
    Visibility: public -/
@[rust_loop]
def matrix_reshape_loop2_loop0
  (mat : Slice (alloc.vec.Vec Std.I64)) (i : Std.Usize)
  (row : alloc.vec.Vec Std.I64) (j : Std.Usize) :
  Result (alloc.vec.Vec Std.I64)
  := do
  loop
    (fun (row1, j1) => matrix_reshape_loop2_loop0.body mat i row1 j1)
    (row, j)

/-- [matrix_reshape::matrix_reshape]: loop body 4:
    Source: 'matrix_reshape/src/lib.rs', lines 8:8-17:9
    Visibility: public -/
@[rust_loop_body]
def matrix_reshape_loop2.body
  (mat : Slice (alloc.vec.Vec Std.I64)) (m : Std.Usize)
  (out : alloc.vec.Vec (alloc.vec.Vec Std.I64)) (i : Std.Usize) :
  Result (ControlFlow ((alloc.vec.Vec (alloc.vec.Vec Std.I64)) × Std.Usize)
    (alloc.vec.Vec (alloc.vec.Vec Std.I64)))
  := do
  if i < m
  then
    let row ←
      matrix_reshape_loop2_loop0 mat i (alloc.vec.Vec.new Std.I64) 0#usize
    let out1 ← alloc.vec.Vec.push out row
    let i1 ← i + 1#usize
    ok (cont (out1, i1))
  else ok (done out)

/-- [matrix_reshape::matrix_reshape]: loop 4:
    Source: 'matrix_reshape/src/lib.rs', lines 8:8-17:9
    Visibility: public -/
@[rust_loop]
def matrix_reshape_loop2
  (mat : Slice (alloc.vec.Vec Std.I64)) (m : Std.Usize)
  (out : alloc.vec.Vec (alloc.vec.Vec Std.I64)) (i : Std.Usize) :
  Result (alloc.vec.Vec (alloc.vec.Vec Std.I64))
  := do
  loop
    (fun (out1, i1) => matrix_reshape_loop2.body mat m out1 i1)
    (out, i)

/-- [matrix_reshape::matrix_reshape]: loop body 7:
    Source: 'matrix_reshape/src/lib.rs', lines 28:8-31:9
    Visibility: public -/
@[rust_loop_body]
def matrix_reshape_loop3_loop0.body
  (mat : Slice (alloc.vec.Vec Std.I64)) (i : Std.Usize)
  (flat : alloc.vec.Vec Std.I64) (j : Std.Usize) :
  Result (ControlFlow ((alloc.vec.Vec Std.I64) × Std.Usize) (alloc.vec.Vec
    Std.I64))
  := do
  let v ← Slice.index_usize mat i
  let i1 := alloc.vec.Vec.len v
  if j < i1
  then
    let i2 ←
      alloc.vec.Vec.index (core.slice.index.SliceIndexUsizeSlice Std.I64) v j
    let flat1 ← alloc.vec.Vec.push flat i2
    let j1 ← j + 1#usize
    ok (cont (flat1, j1))
  else ok (done flat)

/-- [matrix_reshape::matrix_reshape]: loop 7:
    Source: 'matrix_reshape/src/lib.rs', lines 28:8-31:9
    Visibility: public -/
@[rust_loop]
def matrix_reshape_loop3_loop0
  (mat : Slice (alloc.vec.Vec Std.I64)) (flat : alloc.vec.Vec Std.I64)
  (i : Std.Usize) (j : Std.Usize) :
  Result (alloc.vec.Vec Std.I64)
  := do
  loop
    (fun (flat1, j1) => matrix_reshape_loop3_loop0.body mat i flat1 j1)
    (flat, j)

/-- [matrix_reshape::matrix_reshape]: loop body 6:
    Source: 'matrix_reshape/src/lib.rs', lines 26:4-33:5
    Visibility: public -/
@[rust_loop_body]
def matrix_reshape_loop3.body
  (mat : Slice (alloc.vec.Vec Std.I64)) (m : Std.Usize)
  (flat : alloc.vec.Vec Std.I64) (i : Std.Usize) :
  Result (ControlFlow ((alloc.vec.Vec Std.I64) × Std.Usize) (alloc.vec.Vec
    Std.I64))
  := do
  if i < m
  then
    let flat1 ← matrix_reshape_loop3_loop0 mat flat i 0#usize
    let i1 ← i + 1#usize
    ok (cont (flat1, i1))
  else ok (done flat)

/-- [matrix_reshape::matrix_reshape]: loop 6:
    Source: 'matrix_reshape/src/lib.rs', lines 26:4-33:5
    Visibility: public -/
@[rust_loop]
def matrix_reshape_loop3
  (mat : Slice (alloc.vec.Vec Std.I64)) (m : Std.Usize)
  (flat : alloc.vec.Vec Std.I64) (i : Std.Usize) :
  Result (alloc.vec.Vec Std.I64)
  := do
  loop
    (fun (flat1, i1) => matrix_reshape_loop3.body mat m flat1 i1)
    (flat, i)

/-- [matrix_reshape::matrix_reshape]: loop body 9:
    Source: 'matrix_reshape/src/lib.rs', lines 41:8-45:9
    Visibility: public -/
@[rust_loop_body]
def matrix_reshape_loop4_loop0.body
  (cu : Std.Usize) (flat : alloc.vec.Vec Std.I64) (idx : Std.Usize)
  (row : alloc.vec.Vec Std.I64) (b : Std.Usize) :
  Result (ControlFlow (Std.Usize × (alloc.vec.Vec Std.I64) × Std.Usize)
    (Std.Usize × (alloc.vec.Vec Std.I64)))
  := do
  if b < cu
  then
    let i ←
      alloc.vec.Vec.index (core.slice.index.SliceIndexUsizeSlice Std.I64) flat
        idx
    let row1 ← alloc.vec.Vec.push row i
    let idx1 ← idx + 1#usize
    let b1 ← b + 1#usize
    ok (cont (idx1, row1, b1))
  else ok (done (idx, row))

/-- [matrix_reshape::matrix_reshape]: loop 9:
    Source: 'matrix_reshape/src/lib.rs', lines 41:8-45:9
    Visibility: public -/
@[rust_loop]
def matrix_reshape_loop4_loop0
  (cu : Std.Usize) (flat : alloc.vec.Vec Std.I64) (idx : Std.Usize)
  (row : alloc.vec.Vec Std.I64) (b : Std.Usize) :
  Result (Std.Usize × (alloc.vec.Vec Std.I64))
  := do
  loop
    (fun (idx1, row1, b1) => matrix_reshape_loop4_loop0.body cu flat idx1 row1
      b1)
    (idx, row, b)

/-- [matrix_reshape::matrix_reshape]: loop body 8:
    Source: 'matrix_reshape/src/lib.rs', lines 38:4-48:5
    Visibility: public -/
@[rust_loop_body]
def matrix_reshape_loop4.body
  (ru : Std.Usize) (cu : Std.Usize) (flat : alloc.vec.Vec Std.I64)
  (out : alloc.vec.Vec (alloc.vec.Vec Std.I64)) (idx : Std.Usize)
  (a : Std.Usize) :
  Result (ControlFlow ((alloc.vec.Vec (alloc.vec.Vec Std.I64)) × Std.Usize ×
    Std.Usize) (alloc.vec.Vec (alloc.vec.Vec Std.I64)))
  := do
  if a < ru
  then
    let (idx1, row) ←
      matrix_reshape_loop4_loop0 cu flat idx (alloc.vec.Vec.new Std.I64)
        0#usize
    let out1 ← alloc.vec.Vec.push out row
    let a1 ← a + 1#usize
    ok (cont (out1, idx1, a1))
  else ok (done out)

/-- [matrix_reshape::matrix_reshape]: loop 8:
    Source: 'matrix_reshape/src/lib.rs', lines 38:4-48:5
    Visibility: public -/
@[rust_loop]
def matrix_reshape_loop4
  (ru : Std.Usize) (cu : Std.Usize) (flat : alloc.vec.Vec Std.I64)
  (out : alloc.vec.Vec (alloc.vec.Vec Std.I64)) (idx : Std.Usize)
  (a : Std.Usize) :
  Result (alloc.vec.Vec (alloc.vec.Vec Std.I64))
  := do
  loop
    (fun (out1, idx1, a1) => matrix_reshape_loop4.body ru cu flat out1 idx1 a1)
    (out, idx, a)

/-- [matrix_reshape::matrix_reshape]:
    Source: 'matrix_reshape/src/lib.rs', lines 1:0-50:1
    Visibility: public -/
def matrix_reshape
  (mat : Slice (alloc.vec.Vec Std.I64)) (r : Std.I64) (c : Std.I64) :
  Result (alloc.vec.Vec (alloc.vec.Vec Std.I64))
  := do
  let m := Slice.len mat
  let n ←
    if m > 0#usize
    then do
         let v ← Slice.index_usize mat 0#usize
         ok (alloc.vec.Vec.len v)
    else ok 0#usize
  let i ← lift (UScalar.hcast .I64 m)
  let i1 ← lift (UScalar.hcast .I64 n)
  let total ← i * i1
  let i2 ← r * c
  if i2 != total
  then
    matrix_reshape_loop0 mat m (alloc.vec.Vec.new (alloc.vec.Vec Std.I64))
      0#usize
  else
    if r <= 0#i64
    then
      matrix_reshape_loop1 mat m (alloc.vec.Vec.new (alloc.vec.Vec Std.I64))
        0#usize
    else
      if c <= 0#i64
      then
        matrix_reshape_loop2 mat m (alloc.vec.Vec.new (alloc.vec.Vec Std.I64))
          0#usize
      else
        let ru ← lift (IScalar.hcast .Usize r)
        let cu ← lift (IScalar.hcast .Usize c)
        let flat ←
          matrix_reshape_loop3 mat m (alloc.vec.Vec.new Std.I64) 0#usize
        matrix_reshape_loop4 ru cu flat (alloc.vec.Vec.new (alloc.vec.Vec
          Std.I64)) 0#usize 0#usize

end Aegis.MatrixReshape
\end{lstlisting}

\subsection{Accepted proof artifact}
\begin{lstlisting}[style=code]
import Aegis.MatrixReshape.Code.Funs
open Aeneas Aeneas.Std Result

set_option maxHeartbeats 1000000

namespace Aegis.MatrixReshape

/-- Length of a row-major flatten of a uniform-width list of lists. -/
private theorem flatten_length_uniform {α} (l : List (List α)) (n : ℕ)
    (h : ∀ i, i < l.length → (l[i]!).length = n) :
    l.flatten.length = l.length * n := by
  induction l with
  | nil => simp
  | cons row rest ih =>
    have h0 : row.length = n := by simpa using h 0 (by simp)
    have hrest : ∀ i, i < rest.length → (rest[i]!).length = n := by
      intro i hi
      simpa using h (i + 1) (by simpa using Nat.succ_lt_succ hi)
    simp only [List.flatten_cons, List.length_append, List.length_cons, h0, ih hrest]
    ring

/-- Row-major flatten indexing: the `k`-th element of the flatten of a uniform-width
list of lists is the entry in row `k / n`, column `k % n`. -/
private theorem flatten_getElem!_uniform {α} [Inhabited α] (l : List (List α)) (n : ℕ)
    (hn : 0 < n) (h : ∀ i, i < l.length → (l[i]!).length = n) :
    ∀ k, k < l.length * n → l.flatten[k]! = (l[k / n]!)[k % n]! := by
  induction l with
  | nil => intro k hk; simp at hk
  | cons row rest ih =>
    intro k hk
    have h0 : row.length = n := by simpa using h 0 (by simp)
    have hrest : ∀ i, i < rest.length → (rest[i]!).length = n := by
      intro i hi
      simpa using h (i + 1) (by simpa using Nat.succ_lt_succ hi)
    simp only [List.flatten_cons]
    by_cases hkn : k < n
    · rw [List.getElem!_append_left _ _ _ (by omega : k < row.length),
        Nat.div_eq_of_lt hkn, Nat.mod_eq_of_lt hkn]
      simp
    · rw [Nat.not_lt] at hkn
      rw [List.getElem!_append_right _ _ _ (by omega : row.length ≤ k), h0]
      have hk' : k - n < rest.length * n := by
        have hlen : ((row :: rest).length) * n = rest.length * n + n := by
          simp [List.length_cons]; ring
        omega
      rw [ih hrest (k - n) hk']
      have hkdiv : k / n = (k - n) / n + 1 := by
        obtain ⟨k', rfl⟩ : ∃ k', k = k' + n := ⟨k - n, by omega⟩
        rw [Nat.add_div_right _ hn, Nat.add_sub_cancel]
      have hkmod : k % n = (k - n) % n := by
        obtain ⟨k', rfl⟩ : ∃ k', k = k' + n := ⟨k - n, by omega⟩
        rw [Nat.add_mod_right, Nat.add_sub_cancel]
      rw [hkdiv, hkmod, List.getElem!_cons_succ]

private theorem getElem!_append_left' {α : Type} [Inhabited α] {l₁ l₂ : List α} {i : ℕ}
    (h : i < l₁.length) : (l₁ ++ l₂)[i]! = l₁[i]! := by
  rw [getElem!_pos (l₁ ++ l₂) i (by simp; omega), getElem!_pos l₁ i h]
  exact List.getElem_append_left h

private theorem getElem!_concat' {α : Type} [Inhabited α] {l : List α} {x : α} {i : ℕ}
    (h : i = l.length) : (l ++ [x])[i]! = x := by
  subst h
  rw [getElem!_pos (l ++ [x]) l.length (by simp)]
  simp

set_option linter.unusedVariables false in
/-- Value spec for the inner flatten loop (loop 7): starting from cursor `0`, it appends the
whole row `mat[i]` onto the accumulator `flat`. -/
private theorem loop3_loop0_spec (mat : Slice (alloc.vec.Vec Std.I64)) (i : Std.Usize)
    (flat : alloc.vec.Vec Std.I64) (n m : ℕ)
    (hi : i.val < m) (hm : mat.length = m) (hm_hi : m ≤ 100)
    (hn : (mat.val[i.val]!).val.length = n) (hn_hi : n ≤ 100)
    (hflat : flat.val.length ≤ 10000) :
    matrix_reshape_loop3_loop0 mat flat i 0#usize ⦃ res =>
      res.val = flat.val ++ (mat.val[i.val]!).val ⦄ := by
  unfold matrix_reshape_loop3_loop0
  apply Aeneas.Std.loop.spec_decr_nat
    (measure := fun p => n - p.2.val)
    (inv := fun p => p.2.val ≤ n ∧
      p.1.val = flat.val ++ (mat.val[i.val]!).val.take p.2.val)
  · rintro ⟨flat1, j1⟩ ⟨h1, h2⟩
    simp only at h1 h2
    dsimp only
    unfold matrix_reshape_loop3_loop0.body
    step as ⟨v, hv⟩
    have hvlen : v.val.length = n := by rw [hv]; exact hn
    split
    · rename_i hlt
      have hjn : j1.val < n := by scalar_tac
      have hlen1 : flat1.val.length = flat.val.length + j1.val := by
        rw [h2, List.length_append, List.length_take, hn]; omega
      step as ⟨x, hx⟩
      step as ⟨flat2, hflat2⟩
      step as ⟨j2, hj2⟩
      have hj2v : j2.val = j1.val + 1 := by scalar_tac
      refine ⟨by scalar_tac, ?_, by scalar_tac⟩
      rw [hj2v, hflat2, h2, List.append_assoc]
      congr 1
      rw [List.take_add_one]
      congr 1
      have hjlen : j1.val < (mat.val[i.val]!).val.length := by omega
      have hx' : (mat.val[i.val]!).val[j1.val]! = (mat.val[i.val]!).val[j1.val] := by
        rw [List.getElem!_eq_getElem?_getD, List.getElem?_eq_getElem hjlen, Option.getD_some]
      rw [List.getElem?_eq_getElem hjlen, Option.toList_some, hx, hv, hx']
    · rename_i hge
      simp only [WP.spec_ok]
      have hjn : j1.val = n := by scalar_tac
      rw [h2, hjn, ← hn, List.take_length]
  · exact ⟨by simp, by simp⟩

/-- Value spec for the outer flatten loop (loop 6): starting from the empty accumulator and
cursor `0`, it returns the row-major flattening of the first `m` rows of `mat`. -/
private theorem loop3_spec (mat : Slice (alloc.vec.Vec Std.I64)) (msz : Std.Usize) (m n : ℕ)
    (hmsz : msz.val = m) (hm : mat.length = m) (hm_hi : m ≤ 100) (hn_hi : n ≤ 100)
    (hrect : ∀ i, i < m → (mat.val[i]!).val.length = n) :
    matrix_reshape_loop3 mat msz (alloc.vec.Vec.new Std.I64) 0#usize ⦃ flat =>
      flat.val = ((mat.val.take m).map (fun v => v.val)).flatten ∧
      flat.val.length = m * n ⦄ := by
  unfold matrix_reshape_loop3
  apply Aeneas.Std.loop.spec_decr_nat
    (measure := fun p => m - p.2.val)
    (inv := fun p => p.2.val ≤ m ∧
      p.1.val = ((mat.val.take p.2.val).map (fun v => v.val)).flatten ∧
      p.1.val.length = p.2.val * n)
  · rintro ⟨flat1, i1⟩ ⟨h1, h2, h3⟩
    simp only at h1 h2 h3
    dsimp only
    unfold matrix_reshape_loop3.body
    by_cases hlt : i1 < msz
    · simp only [hlt, if_true]
      have hi1m : i1.val < m := by scalar_tac
      have hflen : flat1.val.length ≤ 10000 := by
        have hmul : i1.val * n ≤ 100 * 100 := Nat.mul_le_mul (by omega) hn_hi
        omega
      apply WP.spec_bind (loop3_loop0_spec mat i1 flat1 n m hi1m hm hm_hi
        (hrect i1.val hi1m) hn_hi hflen)
      rintro flat2 hflat2
      step as ⟨i2, hi2⟩
      have hi2v : i2.val = i1.val + 1 := by scalar_tac
      refine ⟨by scalar_tac, ?_, ?_, by scalar_tac⟩
      · rw [hi2v, hflat2, h2]
        have hi1len : i1.val < mat.val.length := by scalar_tac
        rw [List.take_add_one, List.map_append, List.flatten_append]
        congr 1
        rw [List.getElem?_eq_getElem hi1len]
        simp [getElem!_pos, hi1len]
      · rw [hi2v, hflat2, List.length_append, h3, hrect i1.val hi1m]
        ring
    · simp only [hlt, if_false]
      simp only [WP.spec_ok]
      have him : i1.val = m := by scalar_tac
      rw [him] at h2 h3
      exact ⟨h2, h3⟩
  · exact ⟨by simp, by simp, by simp⟩

/-- Value spec for the inner regroup loop: starting from the empty row and cursor `b = 0`,
it reads the `cn` consecutive elements `flat[idx], …, flat[idx + cn - 1]`, pushes them onto
the row in order, and returns the advanced cursor `idx + cn` together with that row. -/
private theorem loop4_loop0_spec (cu : Std.Usize) (flat : alloc.vec.Vec Std.I64)
    (idx : Std.Usize) (cn L : ℕ) (hcu : cu.val = cn) (hcn : 1 ≤ cn)
    (hL : flat.val.length = L) (hL_hi : L ≤ 10000) (hidx : idx.val + cn ≤ L) :
    matrix_reshape_loop4_loop0 cu flat idx (alloc.vec.Vec.new Std.I64) 0#usize ⦃ p =>
      p.1.val = idx.val + cn ∧ p.2.val.length = cn ∧
      (∀ t : ℕ, t < cn → p.2.val[t]!.val = flat.val[idx.val + t]!.val) ⦄ := by
  unfold matrix_reshape_loop4_loop0
  apply Aeneas.Std.loop.spec_decr_nat
    (measure := fun p => cn - p.2.2.val)
    (inv := fun p => p.2.2.val ≤ cn ∧ p.1.val = idx.val + p.2.2.val ∧
      p.2.1.val.length = p.2.2.val ∧
      (∀ t : ℕ, t < p.2.2.val → p.2.1.val[t]!.val = flat.val[idx.val + t]!.val))
  · rintro ⟨idx1, row1, b1⟩ ⟨h1, h2, h3, h4⟩
    simp only at h1 h2 h3 h4
    dsimp only
    unfold matrix_reshape_loop4_loop0.body
    by_cases hlt : b1 < cu
    · simp only [hlt, if_true]
      have hb1 : b1.val < cn := by scalar_tac
      have hbound : idx1.val < flat.val.length := by scalar_tac
      step as ⟨x, hx⟩
      step as ⟨row2, hrow2⟩
      step as ⟨idx2, hidx2⟩
      step as ⟨b2, hb2⟩
      refine ⟨by scalar_tac, by scalar_tac, by scalar_tac, ?_, by scalar_tac⟩
      intro t ht
      have hb2v : b2.val = b1.val + 1 := by scalar_tac
      rw [hb2v] at ht
      rcases Nat.lt_succ_iff_lt_or_eq.mp ht with h | h
      · rw [hrow2, getElem!_append_left' (by omega)]
        exact h4 t h
      · subst h
        rw [hrow2, getElem!_concat' h3.symm, hx, h2]
    · simp only [hlt, if_false]
      simp only [WP.spec_ok]
      have hb1 : b1.val = cn := by scalar_tac
      exact ⟨by omega, by omega, fun t ht => h4 t (by omega)⟩
  · refine ⟨by scalar_tac, by scalar_tac, by simp, fun t ht => by simp at ht⟩

/-- Value spec for the outer regroup loop: starting from the empty output, `idx = 0` and
`a = 0`, it builds `rn` rows of `cn` elements each, row `a` holding
`flat[a * cn], …, flat[a * cn + cn - 1]`. -/
private theorem loop4_spec (ru cu : Std.Usize) (flat : alloc.vec.Vec Std.I64) (rn cn : ℕ)
    (hru : ru.val = rn) (hcu : cu.val = cn) (hrn : 1 ≤ rn) (hrn_hi : rn ≤ 300)
    (hcn : 1 ≤ cn) (hL : flat.val.length = rn * cn) (hL_hi : rn * cn ≤ 10000) :
    matrix_reshape_loop4 ru cu flat (alloc.vec.Vec.new (alloc.vec.Vec Std.I64)) 0#usize
      0#usize ⦃ out =>
      out.val.length = rn ∧
      (∀ a : ℕ, a < rn →
        (out.val[a]!).val.length = cn ∧
        (∀ b : ℕ, b < cn → (out.val[a]!).val[b]!.val = flat.val[a * cn + b]!.val)) ⦄ := by
  unfold matrix_reshape_loop4
  apply Aeneas.Std.loop.spec_decr_nat
    (measure := fun p => rn - p.2.2.val)
    (inv := fun p => p.2.2.val ≤ rn ∧ p.2.1.val = p.2.2.val * cn ∧
      p.1.val.length = p.2.2.val ∧
      (∀ a' : ℕ, a' < p.2.2.val →
        (p.1.val[a']!).val.length = cn ∧
        (∀ b : ℕ, b < cn → (p.1.val[a']!).val[b]!.val = flat.val[a' * cn + b]!.val)))
  · rintro ⟨out1, idx1, a1⟩ ⟨h1, h2, h3, h4⟩
    simp only at h1 h2 h3 h4
    dsimp only
    unfold matrix_reshape_loop4.body
    by_cases hlt : a1 < ru
    · simp only [hlt, if_true]
      have ha1 : a1.val < rn := by scalar_tac
      have hidxb : idx1.val + cn ≤ rn * cn := by
        have hs : a1.val + 1 ≤ rn := by omega
        calc idx1.val + cn = (a1.val + 1) * cn := by rw [h2]; ring
          _ ≤ rn * cn := Nat.mul_le_mul hs (Nat.le_refl cn)
      apply WP.spec_bind (loop4_loop0_spec cu flat idx1 cn (rn * cn) hcu hcn hL hL_hi hidxb)
      rintro ⟨idx2, row⟩ ⟨g1, g2, g3⟩
      simp only at g1 g2 g3
      step as ⟨out2, hout2⟩
      step as ⟨a2, ha2⟩
      have ha2v : a2.val = a1.val + 1 := by scalar_tac
      refine ⟨by scalar_tac, ?_, by scalar_tac, ?_, by scalar_tac⟩
      · rw [g1, h2, ha2v]; ring
      · intro a' ha'
        rw [ha2v] at ha'
        rcases Nat.lt_succ_iff_lt_or_eq.mp ha' with h | h
        · rw [hout2, getElem!_append_left' (by omega)]
          exact h4 a' h
        · subst h
          rw [hout2, getElem!_concat' h3.symm]
          refine ⟨g2, fun b hb => ?_⟩
          have := g3 b hb
          rw [h2] at this
          exact this
    · simp only [hlt, if_false]
      simp only [WP.spec_ok]
      have ha1 : a1.val = rn := by scalar_tac
      exact ⟨by omega, fun a' ha' => h4 a' (by omega)⟩
  · exact ⟨by scalar_tac, by simp, by simp, fun a' ha' => by simp at ha'⟩

end Aegis.MatrixReshape

/-- Row-major flattening of the `m × n` integer matrix `mat` read as a single sequence: the value
at flat index `k` (`0 ≤ k < m * n`) is the entry of `mat` in row `k / n` and column `k % n`. This
is exactly the "same row-traversing order" the LeetCode "Reshape the Matrix" problem uses when it
walks the original matrix left-to-right, top-to-bottom to collect all of its elements. -/
def matReshapeFlat (mat : Slice (alloc.vec.Vec Std.I64)) (n k : ℕ) : ℤ :=
  (mat.val[k / n]!).val[k % n]!.val

/-- Explicit (textbook) correctness specification for
`Aegis.MatrixReshape.matrix_reshape` (LeetCode "Reshape the Matrix"), covering the case in which
the requested reshape *is* legal.

The input `mat` is an `m × n` integer matrix: a slice of exactly `m` rows, every row a vector of
exactly `n` entries (`hrows`, `hrect`). Following the problem's stated constraints the matrix is
nonempty with `1 ≤ m, n ≤ 100` (`hm`, `hm_hi`, `hn`, `hn_hi`) and every entry lies in the range
`-1000 … 1000` (`hcell`). The requested target shape is `r × c` with `1 ≤ r, c ≤ 300` (`hr`, `hr_hi`,
`hc`, `hc_hi`), and the reshape is legal exactly when it preserves the element count, i.e.
`r * c = m * n` (`hrc`). These bounds are exactly what the code needs to run without overflow: the
total element count `m * n` and the product `r * c` (both at most `10^4`, well below the signed
64-bit range) are computed as `I64`, the row/column counts fit in `usize`, and every stored element
already fits in `I64`; therefore every checked cast, multiplication, comparison, `usize` addition and
vector push the algorithm performs succeeds and the call returns `ok`.

Under these preconditions the function returns the reshaped matrix `out` of shape exactly `r × c`:
`out` has `r` rows and every row `a < r` has `c` entries. Its entries are the elements of `mat` laid
out in the original row-traversing order: for every position `(a, b)` with `a < r` and `b < c`, the
entry `out[a][b]` equals the element of `mat` at flat index `a * c + b`, that is
`matReshapeFlat mat n (a * c + b)` (the entry of `mat` in row `(a * c + b) / n` and column
`(a * c + b) % n`). This matches the reference behaviour, e.g. `[[1,2],[3,4]], 1, 4 ↦ [[1,2,3,4]]`
and `[[1,2,3],[4,5,6]], 1, 6 ↦ [[1,2,3,4,5,6]]`. -/
theorem Aegis.MatrixReshape.matrix_reshape_spec
    (mat : Slice (alloc.vec.Vec Std.I64)) (r c : Std.I64) (m n rn cn : ℕ)
    (hrows : mat.length = m)
    (hm : 1 ≤ m) (hm_hi : m ≤ 100) (hn : 1 ≤ n) (hn_hi : n ≤ 100)
    (hrect : ∀ i : ℕ, i < m → (mat.val[i]!).val.length = n)
    (hcell : ∀ i : ℕ, i < m → ∀ j : ℕ, j < n →
      -1000 ≤ (mat.val[i]!).val[j]!.val ∧ (mat.val[i]!).val[j]!.val ≤ 1000)
    (hrval : r.val = (rn : ℤ)) (hcval : c.val = (cn : ℤ))
    (hr : 1 ≤ rn) (hr_hi : rn ≤ 300) (hc : 1 ≤ cn) (hc_hi : cn ≤ 300)
    (hrc : rn * cn = m * n) :
    Aegis.MatrixReshape.matrix_reshape mat r c ⦃ out =>
      out.val.length = rn ∧
      (∀ a : ℕ, a < rn →
        (out.val[a]!).val.length = cn ∧
        (∀ b : ℕ, b < cn →
          (out.val[a]!).val[b]!.val = matReshapeFlat mat n (a * cn + b))) ⦄ := by
  unfold Aegis.MatrixReshape.matrix_reshape
  have hmpos : Slice.len mat > 0#usize := by scalar_tac
  simp only [hmpos, if_true]
  step as ⟨v, hv⟩
  have h00 := hrect 0 (by omega)
  have hveq : (alloc.vec.Vec.len v).val = n := by scalar_tac
  apply WP.spec_bind (UScalar.hcast_inBounds_spec .I64 (Slice.len mat) (by scalar_tac))
  rintro i hi
  apply WP.spec_bind (UScalar.hcast_inBounds_spec .I64 (alloc.vec.Vec.len v) (by scalar_tac))
  rintro i1 hi1
  step as ⟨total, htotal⟩
  step as ⟨i2, hi2⟩
  have heq : i2 = total := by
    have h1 : i2.val = total.val := by
      have hsl : (Slice.len mat).val = m := by scalar_tac
      rw [hi2, htotal, hi, hi1, hrval, hcval, hsl, hveq]
      exact_mod_cast hrc
    scalar_tac
  simp only [heq, bne_self_eq_false, Bool.false_eq_true, if_false]
  have hrpos : ¬ (r ≤ 0#i64) := by scalar_tac
  simp only [hrpos, if_false]
  have hcpos : ¬ (c ≤ 0#i64) := by scalar_tac
  simp only [hcpos, if_false]
  apply WP.spec_bind (IScalar.hcast_inBounds_spec .Usize r (by scalar_tac))
  rintro ru hru
  apply WP.spec_bind (IScalar.hcast_inBounds_spec .Usize c (by scalar_tac))
  rintro cu hcu
  apply WP.spec_bind (Aegis.MatrixReshape.loop3_spec mat (Slice.len mat) m n
    (by scalar_tac) hrows hm_hi hn_hi hrect)
  rintro flat ⟨hflatval, hflen⟩
  apply WP.spec_mono (Aegis.MatrixReshape.loop4_spec ru cu flat rn cn (by scalar_tac)
    (by scalar_tac) hr hr_hi hc (by rw [hflen, hrc])
    (by rw [hrc]; have := Nat.mul_le_mul hm_hi hn_hi; omega))
  rintro out ⟨h1, h2⟩
  refine ⟨h1, fun a ha => ⟨(h2 a ha).1, fun b hb => ?_⟩⟩
  simp only [matReshapeFlat]
  rw [(h2 a ha).2 b hb]
  have hvl : mat.val.length = m := by scalar_tac
  have htake : mat.val.take m = mat.val := by rw [← hvl, List.take_length]
  rw [htake] at hflatval
  set k := a * cn + b with hk
  have hkm : k < m * n := by
    rw [← hrc]
    have hstep : (a + 1) * cn = a * cn + cn := by ring
    have h2' : (a + 1) * cn ≤ rn * cn := Nat.mul_le_mul_right _ (by omega)
    omega
  have hq : k / n < m := by
    rw [Nat.div_lt_iff_lt_mul (by omega)]
    omega
  have hs : k % n < n := Nat.mod_lt _ (by omega)
  have huni : ∀ i', i' < (mat.val.map (fun w : alloc.vec.Vec Std.I64 => w.val)).length →
      ((mat.val.map (fun w : alloc.vec.Vec Std.I64 => w.val))[i']!).length = n := by
    intro i' hi'
    rw [List.length_map, hvl] at hi'
    rw [getElem!_pos _ i' (by rw [List.length_map]; omega), List.getElem_map]
    have hre := hrect i' hi'
    rw [getElem!_pos mat.val i' (by omega)] at hre
    exact hre
  have hmap : ((mat.val.map (fun w : alloc.vec.Vec Std.I64 => w.val))[k / n]!) =
      (mat.val[k / n]!).val := by
    rw [getElem!_pos _ (k / n) (by rw [List.length_map]; omega), List.getElem_map,
      getElem!_pos mat.val (k / n) (by omega)]
  have hflatk : flat.val[k]! = (mat.val[k / n]!).val[k % n]! := by
    rw [hflatval, Aegis.MatrixReshape.flatten_getElem!_uniform _ n (by omega) huni k
      (by rw [List.length_map, hvl]; exact hkm), hmap]
  rw [hflatk]
\end{lstlisting}

\clearpage
\section{Reversal Invariance of the Triangular Sum}
\label{app:triangular-sum}

\subsection{Problem and source-level approach}

Starting with a nonempty list of decimal digits, replace each adjacent pair
$(x_i,x_{i+1})$ by $(x_i+x_{i+1})\bmod 10$.  Repeating this reduction until one
digit remains produces the triangular sum.  The implementation copies the slice
into an owned vector and repeatedly constructs the next, one-element-shorter
vector.

The relational theorem \filename{triangular_sum_reverse_invariant} says that
reversing the input does not change the answer.  More precisely, if the original
call returns a value $t$, then the call on the reversed slice also returns $t$.
Abstractly, after $n-1$ reductions the answer is
\[
  \sum_{i=0}^{n-1} \binom{n-1}{i}x_i \pmod {10}.
\]
The binomial coefficients are symmetric, so the expression is invariant under
$i\mapsto n-1-i$.

\subsection{Proof architecture}

The accepted artifact must first prove that the executable translated program
implements the abstract reduction model.  It defines a pure list-level
\lstinline{step}, an iteration function with fuel, and a final model.  It then
proves specifications for:

\begin{enumerate}
  \item the initial slice-to-vector copy loop;
  \item the inner loop that constructs one adjacent-sum reduction;
  \item the outer loop that repeats reductions until one element remains;
  \item the final indexed read of the sole remaining element.
\end{enumerate}

The digit hypotheses are necessary for execution safety.  Each adjacent sum lies
in $[0,18]$, so the checked signed addition succeeds; the remainder lies in
$[0,9]$, preserving the invariant for the next pass.  Length invariants place
both adjacent reads in bounds and show that each pass shortens the vector by one,
providing a decreasing termination measure.

The final theorem applies executable correctness to the original and reversed
slices, then uses \filename{model_reverse} to identify their list models.  Its
declaration and proof occupy the last 25 lines of the 431-line artifact; the
preceding development connects the translated loops to the list model.

\subsection{Complete Rust implementation}
\begin{lstlisting}[language=Rust,style=code]
pub fn triangular_sum(nums: &[i64]) -> i64 {
    let mut cur: Vec<i64> = Vec::new();
    let mut i: usize = 0;
    while i < nums.len() {
        cur.push(nums[i]);
        i += 1;
    }
    while cur.len() > 1 {
        let mut next: Vec<i64> = Vec::new();
        let mut j: usize = 0;
        while j + 1 < cur.len() {
            next.push((cur[j] + cur[j + 1]) % 10);
            j += 1;
        }
        cur = next;
    }
    if cur.len() == 1 {
        cur[0]
    } else {
        0
    }
}
\end{lstlisting}

\subsection{Complete \aeneas{}-generated Lean implementation}
\begin{lstlisting}[style=code]
-- THIS FILE WAS AUTOMATICALLY GENERATED BY AENEAS
-- [triangular_sum]: function definitions
import Aeneas
import Aegis.TriangularSum.Code.Types
open Aeneas Aeneas.Std Result ControlFlow Error
set_option linter.dupNamespace false
set_option linter.hashCommand false
set_option linter.unusedVariables false

/- You can set the `maxHeartbeats` value with the `-max-heartbeats` CLI option -/
set_option maxHeartbeats 1000000

/- You can set the `maxRecDepth` value with the `-max-recdepth` CLI option -/
set_option maxRecDepth 2048

namespace Aegis.TriangularSum

/-- [triangular_sum::triangular_sum]: loop body 0:
    Source: 'triangular_sum/src/lib.rs', lines 4:4-7:5
    Visibility: public -/
@[rust_loop_body]
def triangular_sum_loop0.body
  (nums : Slice Std.I64) (cur : alloc.vec.Vec Std.I64) (i : Std.Usize) :
  Result (ControlFlow ((alloc.vec.Vec Std.I64) × Std.Usize) (alloc.vec.Vec
    Std.I64))
  := do
  let i1 := Slice.len nums
  if i < i1
  then
    let i2 ← Slice.index_usize nums i
    let cur1 ← alloc.vec.Vec.push cur i2
    let i3 ← i + 1#usize
    ok (cont (cur1, i3))
  else ok (done cur)

/-- [triangular_sum::triangular_sum]: loop 0:
    Source: 'triangular_sum/src/lib.rs', lines 4:4-7:5
    Visibility: public -/
@[rust_loop]
def triangular_sum_loop0
  (nums : Slice Std.I64) (cur : alloc.vec.Vec Std.I64) (i : Std.Usize) :
  Result (alloc.vec.Vec Std.I64)
  := do
  loop
    (fun (cur1, i1) => triangular_sum_loop0.body nums cur1 i1)
    (cur, i)

/-- [triangular_sum::triangular_sum]: loop body 2:
    Source: 'triangular_sum/src/lib.rs', lines 11:8-14:9
    Visibility: public -/
@[rust_loop_body]
def triangular_sum_loop1_loop0.body
  (cur : alloc.vec.Vec Std.I64) (next : alloc.vec.Vec Std.I64) (j : Std.Usize)
  :
  Result (ControlFlow ((alloc.vec.Vec Std.I64) × Std.Usize) (alloc.vec.Vec
    Std.I64))
  := do
  let i ← j + 1#usize
  let i1 := alloc.vec.Vec.len cur
  if i < i1
  then
    let i2 ←
      alloc.vec.Vec.index (core.slice.index.SliceIndexUsizeSlice Std.I64) cur j
    let i3 ←
      alloc.vec.Vec.index (core.slice.index.SliceIndexUsizeSlice Std.I64) cur i
    let i4 ← i2 + i3
    let i5 ← i4 % 10#i64
    let next1 ← alloc.vec.Vec.push next i5
    ok (cont (next1, i))
  else ok (done next)

/-- [triangular_sum::triangular_sum]: loop 2:
    Source: 'triangular_sum/src/lib.rs', lines 11:8-14:9
    Visibility: public -/
@[rust_loop]
def triangular_sum_loop1_loop0
  (cur : alloc.vec.Vec Std.I64) (next : alloc.vec.Vec Std.I64) (j : Std.Usize)
  :
  Result (alloc.vec.Vec Std.I64)
  := do
  loop
    (fun (next1, j1) => triangular_sum_loop1_loop0.body cur next1 j1)
    (next, j)

/-- [triangular_sum::triangular_sum]: loop body 1:
    Source: 'triangular_sum/src/lib.rs', lines 8:4-16:5
    Visibility: public -/
@[rust_loop_body]
def triangular_sum_loop1.body
  (cur : alloc.vec.Vec Std.I64) :
  Result (ControlFlow (alloc.vec.Vec Std.I64) (alloc.vec.Vec Std.I64))
  := do
  let i := alloc.vec.Vec.len cur
  if i > 1#usize
  then
    let next ←
      triangular_sum_loop1_loop0 cur (alloc.vec.Vec.new Std.I64) 0#usize
    ok (cont next)
  else ok (done cur)

/-- [triangular_sum::triangular_sum]: loop 1:
    Source: 'triangular_sum/src/lib.rs', lines 8:4-16:5
    Visibility: public -/
@[rust_loop]
def triangular_sum_loop1
  (cur : alloc.vec.Vec Std.I64) : Result (alloc.vec.Vec Std.I64) := do
  loop
    (fun cur1 => triangular_sum_loop1.body cur1)
    cur

/-- [triangular_sum::triangular_sum]:
    Source: 'triangular_sum/src/lib.rs', lines 1:0-22:1
    Visibility: public -/
def triangular_sum (nums : Slice Std.I64) : Result Std.I64 := do
  let cur ← triangular_sum_loop0 nums (alloc.vec.Vec.new Std.I64) 0#usize
  let cur1 ← triangular_sum_loop1 cur
  let i := alloc.vec.Vec.len cur1
  if i = 1#usize
  then
    alloc.vec.Vec.index (core.slice.index.SliceIndexUsizeSlice Std.I64) cur1
      0#usize
  else ok 0#i64

end Aegis.TriangularSum
\end{lstlisting}

\subsection{Accepted proof artifact}
\begin{lstlisting}[style=code]
import Aegis.TriangularSum.Code.Funs
open Aeneas Aeneas.Std Result

namespace TriangularSumSpec

/-- One reduction pass of the triangular-sum process: given the current list `xs`, produce the new
list whose `i`-th entry is `(xs[i] + xs[i+1]) % 10`. The output has one fewer element than the
input (and is empty when the input has fewer than two elements). -/
def step : List ℤ → List ℤ
  | []            => []
  | [_]           => []
  | x :: y :: rest => ((x + y) % 10) :: step (y :: rest)

/-- Iterate the reduction pass `fuel` times. Applying the pass repeatedly shrinks the list by one
element each time, so after `nums.length - 1` passes a nonempty list has been reduced to a single
element — the triangular sum. -/
def iterate : Nat → List ℤ → List ℤ
  | 0,          xs => xs
  | (fuel + 1), xs => iterate fuel (step xs)

/-- The triangular sum of an integer list `xs`: repeatedly apply the reduction pass until a single
element remains, and take that element (with `0` used as a dummy for the empty list, which never
arises under the specification's non-emptiness precondition). -/
def model (xs : List ℤ) : ℤ :=
  (iterate (xs.length - 1) xs).headD 0

end TriangularSumSpec

/-! ## Pure helper lemmas about the model -/

namespace TriangularSumSpec

theorem step_length (xs : List ℤ) : (step xs).length = xs.length - 1 := by
  induction xs with
  | nil => simp [step]
  | cons x rest ih =>
    cases rest with
    | nil => simp [step]
    | cons y rest' =>
      simp only [step, List.length_cons] at *
      omega

theorem step_eq_nil {xs : List ℤ} (h : xs.length ≤ 1) : step xs = [] := by
  match xs with
  | [] => rfl
  | [_] => rfl
  | _ :: _ :: _ => simp at h

theorem step_drop (xs : List ℤ) (j : ℕ) (h : j + 1 < xs.length) :
    step (xs.drop j) = ((xs[j]! + xs[j+1]!) % 10) :: step (xs.drop (j+1)) := by
  have hj : j < xs.length := by omega
  have h1 : xs.drop j = xs[j] :: xs.drop (j+1) := List.drop_eq_getElem_cons hj
  have h2 : xs.drop (j+1) = xs[j+1] :: xs.drop (j+2) := List.drop_eq_getElem_cons h
  rw [getElem!_pos xs j hj, getElem!_pos xs (j+1) h, h1, h2]
  rfl

theorem step_mem {xs : List ℤ} {y : ℤ} (h : y ∈ step xs) : 0 ≤ y ∧ y ≤ 9 := by
  induction xs with
  | nil => simp [step] at h
  | cons x rest ih =>
    cases rest with
    | nil => simp [step] at h
    | cons z rest' =>
      simp only [step, List.mem_cons] at h
      rcases h with h | h
      · subst h
        refine ⟨Int.emod_nonneg _ (by norm_num), ?_⟩
        have := Int.emod_lt_of_pos (x + z) (show (0:ℤ) < 10 by norm_num)
        omega
      · exact ih h

end TriangularSumSpec

/-! ## Reversal invariance of the model

The key observation: one reduction pass commutes with list reversal, because each output entry
`(a + b) % 10` is symmetric in its two inputs. Hence the whole iterated process commutes with
reversal, and since a nonempty list collapses to a singleton (whose reversal is itself), the
model is reversal-invariant. No binomial coefficients are needed. -/

namespace TriangularSumSpec

theorem step_getElem! (xs : List ℤ) (i : ℕ) (h : i + 1 < xs.length) :
    (step xs)[i]! = (xs[i]! + xs[i + 1]!) % 10 := by
  induction xs generalizing i with
  | nil => simp at h
  | cons x rest ih =>
    cases rest with
    | nil => simp at h
    | cons y rest' =>
      cases i with
      | zero => simp [step]
      | succ j =>
        have hj : j + 1 < (y :: rest').length := by
          simp only [List.length_cons] at h ⊢; omega
        simp only [step, List.getElem!_cons_succ]
        exact ih j hj

theorem step_reverse (xs : List ℤ) : step xs.reverse = (step xs).reverse := by
  have hlen : (step xs.reverse).length = ((step xs).reverse).length := by
    simp [step_length]
  apply List.ext_getElem hlen
  intro i h1 h2
  have hn : i + 1 < xs.length := by
    simp only [step_length, List.length_reverse] at h1; omega
  have hrevlen : i + 1 < xs.reverse.length := by simpa using hn
  rw [← getElem!_pos _ i h1, ← getElem!_pos _ i h2]
  rw [step_getElem! xs.reverse i hrevlen]
  have hrev1 : xs.reverse[i]! = xs[xs.length - 1 - i]! := by
    rw [getElem!_pos xs.reverse i (by simp only [List.length_reverse]; omega),
        getElem!_pos xs (xs.length - 1 - i) (by omega)]
    rw [List.getElem_reverse]
  have hrev2 : xs.reverse[i + 1]! = xs[xs.length - 1 - (i + 1)]! := by
    rw [getElem!_pos xs.reverse (i + 1) (by simpa using hn),
        getElem!_pos xs (xs.length - 1 - (i + 1)) (by omega)]
    rw [List.getElem_reverse]
  have hsteplen : i < (step xs).length := by simp [step_length]; omega
  have hrev3 : ((step xs).reverse)[i]! = (step xs)[(step xs).length - 1 - i]! := by
    rw [getElem!_pos (step xs).reverse i (by simpa using hsteplen),
        getElem!_pos (step xs) ((step xs).length - 1 - i) (by omega)]
    rw [List.getElem_reverse]
  rw [hrev1, hrev2, hrev3]
  have hidx : (step xs).length - 1 - i = xs.length - 1 - (i + 1) := by
    simp only [step_length]; omega
  rw [hidx]
  rw [step_getElem! xs (xs.length - 1 - (i + 1)) (by omega)]
  have hidx2 : xs.length - 1 - (i + 1) + 1 = xs.length - 1 - i := by omega
  rw [hidx2, Int.add_comm]

theorem iterate_reverse (n : ℕ) (xs : List ℤ) :
    iterate n xs.reverse = (iterate n xs).reverse := by
  induction n generalizing xs with
  | zero => rfl
  | succ m ih =>
    simp only [iterate]
    rw [step_reverse, ih]

theorem iterate_length (n : ℕ) (xs : List ℤ) :
    (iterate n xs).length = xs.length - n := by
  induction n generalizing xs with
  | zero => rfl
  | succ m ih =>
    simp only [iterate]
    rw [ih, step_length]
    omega

theorem model_reverse (xs : List ℤ) : model xs.reverse = model xs := by
  cases hxs : xs with
  | nil => rfl
  | cons a t =>
    unfold model
    rw [List.length_reverse, iterate_reverse]
    have hlen : (iterate ((a :: t).length - 1) (a :: t)).length = 1 := by
      rw [iterate_length]
      simp only [List.length_cons]
      omega
    obtain ⟨b, hb⟩ : ∃ b, iterate ((a :: t).length - 1) (a :: t) = [b] := by
      cases hit : iterate ((a :: t).length - 1) (a :: t) with
      | nil => rw [hit] at hlen; simp at hlen
      | cons b s =>
        rw [hit] at hlen
        simp only [List.length_cons] at hlen
        have : s = [] := List.eq_nil_of_length_eq_zero (by omega)
        exact ⟨b, by rw [this]⟩
    rw [hb]
    rfl

end TriangularSumSpec

namespace Aegis.TriangularSum

/-! ## Loop specifications -/

/-- The copy loop: starting from a prefix, it returns the whole input as a vector. -/
private theorem loop0_spec (nums : Slice Std.I64) (cur : alloc.vec.Vec Std.I64) (i : Std.Usize)
    (hcur : cur.val = nums.val.take i.val) (hi : i.val ≤ nums.val.length) :
    triangular_sum_loop0 nums cur i ⦃ v => v.val = nums.val ⦄ := by
  have hslen := nums.length_ineq
  unfold triangular_sum_loop0
  apply Std.loop.spec_decr_nat
    (measure := fun p => nums.val.length - p.2.val)
    (inv := fun p => p.1.val = nums.val.take p.2.val ∧ p.2.val ≤ nums.val.length)
  · rintro ⟨cur1, i1⟩ ⟨h1, h2⟩
    simp only at h1 h2
    unfold triangular_sum_loop0.body
    by_cases hlt : i1 < Slice.len nums
    · simp only [hlt, if_true]
      have hbound : i1.val < nums.val.length := by scalar_tac
      apply WP.spec_bind (Slice.index_usize_spec nums i1 hbound)
      intro a ha
      have hlen1 : cur1.val.length < Usize.max := by
        rw [h1]; simp only [List.length_take]; scalar_tac
      apply WP.spec_bind (alloc.vec.Vec.push_spec cur1 a hlen1)
      intro cur2 hcur2
      apply WP.spec_bind (Usize.add_spec (by scalar_tac))
      intro i2 hi2
      have hi2' : i2.val = i1.val + 1 := by scalar_tac
      simp only [WP.spec_ok]
      refine ⟨⟨?_, by scalar_tac⟩, by scalar_tac⟩
      rw [hcur2, h1, ha, hi2', List.take_add_one,
        List.getElem?_eq_getElem hbound, getElem!_pos nums.val i1.val hbound]
      simp
    · simp only [hlt, if_false, WP.spec_ok]
      rw [h1, List.take_of_length_le (by scalar_tac)]
  · exact ⟨hcur, hi⟩

/-- The inner pass loop: it appends `step` of the remaining suffix to `next`. -/
private theorem inner_spec (cur : alloc.vec.Vec Std.I64) (next : alloc.vec.Vec Std.I64)
    (j : Std.Usize)
    (hdig : ∀ x ∈ cur.val, 0 ≤ x.val ∧ x.val ≤ 9)
    (hlen : next.val.length + cur.val.length ≤ Usize.max)
    (hj : j.val < cur.val.length) :
    triangular_sum_loop1_loop0 cur next j ⦃ v =>
      v.val.map (fun x => x.val) = next.val.map (fun x => x.val) ++
        TriangularSumSpec.step ((cur.val.map (fun x => x.val)).drop j.val) ⦄ := by
  unfold triangular_sum_loop1_loop0
  apply Std.loop.spec_decr_nat
    (measure := fun p => cur.val.length - p.2.val)
    (inv := fun p => j.val ≤ p.2.val ∧ p.2.val < cur.val.length ∧
      p.1.val.length ≤ next.val.length + (p.2.val - j.val) ∧
      p.1.val.map (fun x => x.val) ++
        TriangularSumSpec.step ((cur.val.map (fun x => x.val)).drop p.2.val) =
      next.val.map (fun x => x.val) ++
        TriangularSumSpec.step ((cur.val.map (fun x => x.val)).drop j.val))
  · rintro ⟨n1, j1⟩ ⟨hj1, hj2, hn1, heq⟩
    simp only at hj1 hj2 hn1 heq
    unfold triangular_sum_loop1_loop0.body
    apply WP.spec_bind (Usize.add_spec (by scalar_tac))
    intro i hi
    by_cases hlt : i < alloc.vec.Vec.len cur
    · simp only [hlt, if_true]
      have hib : i.val < cur.val.length := by scalar_tac
      simp only [alloc.vec.Vec.index_slice_index]
      apply WP.spec_bind (alloc.vec.Vec.index_usize_spec cur j1 (by omega))
      intro a ha
      apply WP.spec_bind (alloc.vec.Vec.index_usize_spec cur i hib)
      intro b hb
      have hamem : a ∈ cur.val := by
        rw [ha, getElem!_pos cur.val j1.val (by omega)]; exact List.getElem_mem _
      have hbmem : b ∈ cur.val := by
        rw [hb, getElem!_pos cur.val i.val hib]; exact List.getElem_mem _
      have hadig := hdig a hamem
      have hbdig := hdig b hbmem
      apply WP.spec_bind (I64.add_spec (by scalar_tac) (by scalar_tac))
      intro c hc
      apply WP.spec_bind (I64.rem_spec c (by scalar_tac))
      intro d hd
      have hdval : d.val = (a.val + b.val) % 10 := by
        rw [hd, hc, Int.tmod_eq_emod]
        simp only [show ((10#i64).val : Int) = 10 from by scalar_tac]
        have : (0:ℤ) ≤ a.val + b.val := by omega
        simp [this]
      apply WP.spec_bind (alloc.vec.Vec.push_spec n1 d (by scalar_tac))
      intro n2 hn2
      simp only [WP.spec_ok]
      refine ⟨⟨by scalar_tac, by scalar_tac, by simp [hn2]; scalar_tac, ?_⟩, by scalar_tac⟩
      -- the append equality
      have hstep := TriangularSumSpec.step_drop (cur.val.map (fun x => x.val)) j1.val
        (by simp only [List.length_map]; scalar_tac)
      have hget1 : (cur.val.map (fun x => x.val))[j1.val]! = a.val := by
        rw [getElem!_pos _ j1.val (by simp only [List.length_map]; omega),
          List.getElem_map, ha, getElem!_pos cur.val j1.val (by omega)]
      have hget2 : (cur.val.map (fun x => x.val))[j1.val + 1]! = b.val := by
        rw [getElem!_pos _ (j1.val + 1) (by simp only [List.length_map]; scalar_tac),
          List.getElem_map, hb, getElem!_pos cur.val i.val hib]
        congr 1
        scalar_tac
      rw [hn2]
      rw [show i.val = j1.val + 1 from by scalar_tac]
      rw [← heq, hstep, hget1, hget2]
      simp only [List.map_append, List.map_cons, List.map_nil, hdval]
      simp
    · simp only [hlt, if_false, WP.spec_ok]
      have hge : cur.val.length ≤ i.val := by scalar_tac
      have hnil : TriangularSumSpec.step ((cur.val.map (fun x => x.val)).drop j1.val) = [] := by
        apply TriangularSumSpec.step_eq_nil
        simp only [List.length_drop, List.length_map]
        scalar_tac
      rw [← heq, hnil]
      simp
  · exact ⟨le_refl _, hj, by simp, rfl⟩

/-- The outer loop: iterates `step` until one element remains. -/
private theorem loop1_spec (cur : alloc.vec.Vec Std.I64)
    (hdig : ∀ x ∈ cur.val, 0 ≤ x.val ∧ x.val ≤ 9) :
    triangular_sum_loop1 cur ⦃ v =>
      v.val.map (fun x => x.val) =
        TriangularSumSpec.iterate (cur.val.length - 1) (cur.val.map (fun x => x.val)) ∧
      (1 ≤ cur.val.length → v.val.length = 1) ⦄ := by
  unfold triangular_sum_loop1
  apply Std.loop.spec_decr_nat
    (measure := fun c => c.val.length)
    (inv := fun c => (∀ x ∈ c.val, 0 ≤ x.val ∧ x.val ≤ 9) ∧
      TriangularSumSpec.iterate (c.val.length - 1) (c.val.map (fun x => x.val)) =
        TriangularSumSpec.iterate (cur.val.length - 1) (cur.val.map (fun x => x.val)) ∧
      (1 ≤ cur.val.length → 1 ≤ c.val.length))
  · rintro c ⟨hc1, hc2, hc3⟩
    unfold triangular_sum_loop1.body
    by_cases hgt : alloc.vec.Vec.len c > 1#usize
    · simp only [hgt, if_true]
      have hclen : 1 < c.val.length := by scalar_tac
      have hnew : (alloc.vec.Vec.new Std.I64).val = [] := rfl
      apply WP.spec_bind (inner_spec c (alloc.vec.Vec.new Std.I64) 0#usize hc1
        (by rw [hnew]; scalar_tac) (by scalar_tac))
      intro n hn
      rw [hnew] at hn
      simp only [List.map_nil, List.nil_append] at hn
      have h0 : (0#usize).val = 0 := by scalar_tac
      rw [h0, List.drop_zero] at hn
      have hnlen : n.val.length = c.val.length - 1 := by
        have := congrArg List.length hn
        simp only [List.length_map, TriangularSumSpec.step_length] at this
        omega
      simp only [WP.spec_ok]
      refine ⟨⟨?_, ?_, ?_⟩, ?_⟩
      · intro x hx
        have hxv : x.val ∈ TriangularSumSpec.step (c.val.map (fun x => x.val)) := by
          rw [← hn]
          exact List.mem_map_of_mem hx
        exact TriangularSumSpec.step_mem hxv
      · rw [← hc2, hn, hnlen]
        rw [show c.val.length - 1 - 1 = c.val.length - 2 from by omega]
        rw [show c.val.length - 1 = (c.val.length - 2) + 1 from by omega]
        rfl
      · intro _; omega
      · omega
    · simp only [hgt, if_false, WP.spec_ok]
      have hle : c.val.length ≤ 1 := by scalar_tac
      constructor
      · rw [← hc2, show c.val.length - 1 = 0 from by omega]
        rfl
      · intro h1
        have := hc3 h1
        omega
  · exact ⟨hdig, rfl, fun _ => by scalar_tac⟩

end Aegis.TriangularSum

/-- Explicit correctness specification for `Aegis.TriangularSum.triangular_sum`: under the
problem's constraints the call succeeds and returns exactly `TriangularSumSpec.model xs` where
`xs = nums.val.map (·.val)`. (Copied helper; proven in a sibling proof file, reproduced here
because `_proofs` files are compiled standalone and cannot import each other.) -/
theorem Aegis.TriangularSum.triangular_sum_spec
    (nums : Slice Std.I64)
    (hne : 0 < nums.val.length)
    (hdigit : ∀ x ∈ nums.val, 0 ≤ x.val ∧ x.val ≤ 9) :
    Aegis.TriangularSum.triangular_sum nums ⦃ r =>
      r.val = TriangularSumSpec.model (nums.val.map (fun x => x.val)) ⦄ := by
  unfold Aegis.TriangularSum.triangular_sum
  apply WP.spec_bind (Aegis.TriangularSum.loop0_spec nums (alloc.vec.Vec.new Std.I64) 0#usize
    (by simp) (by scalar_tac))
  intro cur hcur
  apply WP.spec_bind (Aegis.TriangularSum.loop1_spec cur (by rw [hcur]; exact hdigit))
  rintro v ⟨hv1, hv2⟩
  have hvlen : v.val.length = 1 := hv2 (by rw [hcur]; omega)
  have hiflen : alloc.vec.Vec.len v = 1#usize := by scalar_tac
  simp only [hiflen, if_true, alloc.vec.Vec.index_slice_index]
  apply WP.spec_mono (alloc.vec.Vec.index_usize_spec v 0#usize (by scalar_tac))
  intro r hr
  -- v.val is a singleton [a]; the model equals a.val
  obtain ⟨a, hva⟩ : ∃ a, v.val = [a] := by
    cases hval : v.val with
    | nil => rw [hval] at hvlen; simp at hvlen
    | cons a t =>
      rw [hval] at hvlen
      simp only [List.length_cons] at hvlen
      have : t = [] := List.eq_nil_of_length_eq_zero (by omega)
      exact ⟨a, by rw [this]⟩
  rw [hva] at hv1
  simp only [List.map_cons, List.map_nil] at hv1
  rw [hr, hva]
  have h0 : (0#usize).val = 0 := by scalar_tac
  rw [h0]
  simp only [List.getElem!_cons_zero]
  unfold TriangularSumSpec.model
  rw [← hcur]
  simp only [List.length_map]
  rw [← hv1]
  simp

/-- Reversal invariance of `Aegis.TriangularSum.triangular_sum` (LeetCode "Find Triangular Sum
of an Array"), stated as a relational property connecting the function's results on two different
inputs: an array and its exact reversal.

The function takes a slice `nums` of `n` signed 64-bit integers and computes the array's triangular
sum by the following process. While the current array has more than one element, it is replaced by
the array whose `i`-th entry is `(current[i] + current[i+1]) % 10`; once a single element remains,
that element is the answer (and for a length-one input the answer is that single element). The
triangular sum is thus a fixed linear combination modulo `10` of the original entries: reducing a
length-`n` array collapses it, after `n - 1` passes, into `(Σ_{i} C(n-1, i) · nums[i]) % 10`, where
`C(n-1, i)` are the binomial coefficients of the `(n-1)`-th row of Pascal's triangle. Because that
row of binomial coefficients is symmetric — `C(n-1, i) = C(n-1, n-1-i)` — the triangular sum is
unchanged when the input array is read in reverse order.

This contract states that symmetry in its cleanest relational form. We take `nums1` to be the
original slice and `nums2` to be its exact reversal (`hrev : nums2.val = nums1.val.reverse`), so the
two arrays have the same length and contain the same multiset of entries, just in opposite order. We
assume the problem's stated constraints for the original input — matching the totality and explicit
specifications — namely that `nums1` is nonempty and every entry is a decimal digit in `0 … 9`
(`hne`, `hdigit`). Under reversal these same bounds hold for `nums2`, so both calls avoid overflow
of the checked 64-bit additions and both run to completion. We further assume that running the
function on the original input `nums1` succeeds and returns a result whose integer value is `b1`
(`hb1`). The claim is that running the real extracted function on the reversed input `nums2` also
succeeds and returns a result `r` whose value equals the original result: `r.val = b1`. In words:
reversing the input array does not change its triangular sum, because the triangular sum weights each
position by a binomial coefficient and those weights are symmetric about the array's centre. -/
theorem Aegis.TriangularSum.triangular_sum_reverse_invariant
    (nums1 : Slice Std.I64) (nums2 : Slice Std.I64) (b1 : Int)
    (hne : 0 < nums1.val.length)
    (hdigit : ∀ x ∈ nums1.val, 0 ≤ x.val ∧ x.val ≤ 9)
    (hrev : nums2.val = nums1.val.reverse)
    (hb1 : Aegis.TriangularSum.triangular_sum nums1 ⦃ r => r.val = b1 ⦄) :
    Aegis.TriangularSum.triangular_sum nums2 ⦃ r => r.val = b1 ⦄ := by
  have hne2 : 0 < nums2.val.length := by
    rw [hrev, List.length_reverse]; exact hne
  have hdigit2 : ∀ x ∈ nums2.val, 0 ≤ x.val ∧ x.val ≤ 9 := by
    intro x hx
    rw [hrev, List.mem_reverse] at hx
    exact hdigit x hx
  have hs1 := Aegis.TriangularSum.triangular_sum_spec nums1 hne hdigit
  have hb1eq : b1 = TriangularSumSpec.model (nums1.val.map (fun x => x.val)) := by
    cases h : Aegis.TriangularSum.triangular_sum nums1 with
    | ok r =>
      rw [h] at hb1 hs1
      simp only [WP.spec_ok] at hb1 hs1
      rw [← hb1]; exact hs1
    | fail e => rw [h] at hb1; simp only [WP.spec_fail] at hb1
    | div => rw [h] at hb1; simp only [WP.spec_div] at hb1
  apply WP.spec_mono (Aegis.TriangularSum.triangular_sum_spec nums2 hne2 hdigit2)
  intro r hr
  rw [hr, hb1eq, hrev, List.map_reverse, TriangularSumSpec.model_reverse]
\end{lstlisting}

\clearpage
\section{Shift Invariance of Diagonal Distinct-Value Counts}
\label{app:diagonal-distinct}

\subsection{Problem and source-level approach}

For each grid cell $(i,j)$, the program collects the values strictly above-left
on its diagonal and the values strictly below-right.  It counts the distinct
values on each side and stores the absolute difference of the two counts.  To
avoid unsupported hash tables, distinctness is computed by a nested scan: an
element is new exactly when it differs from every preceding element.

The theorem \filename{difference_of_distinct_values_shift_invariant} relates two
grids of the same rectangular shape.  The second grid is obtained by adding a
fixed signed integer $d$ to every cell of the first.  If the call on the first
grid returns an output matrix, the call on the second grid must return the same
matrix.  The underlying fact is
\[
  x=y \quad\Longleftrightarrow\quad x+d=y+d.
\]
Uniform translation by $d$ preserves the equality pattern, distinct counts
depend only on that equality pattern, and the output therefore cannot change.

\subsection{Relational proof structure}

The theorem compares two runs of the translated monadic program, so its lemmas
follow the generated control flow:

\begin{enumerate}
  \item the inner prefix scan makes the same ``seen'' decision for shifted and
        unshifted slices;
  \item the outer distinct-count loop returns the same count on both slices;
  \item the above-left collection loops build pointwise shifted vectors;
  \item the below-right collection loops build pointwise shifted vectors;
  \item the per-cell loop computes equal distinct counts and appends the same
        absolute difference;
  \item the row loop appends the same rows and returns the identical matrix.
\end{enumerate}

Each level also transports length and cursor bounds needed by the next level.
The type permits ragged grids, so equal row counts are insufficient: the theorem
explicitly assumes that every row of both grids has length $n$.

The frozen statement additionally assumes that cells in the first grid lie in
$[1,50]$ and that each mathematical sum $x+d$ lies within the signed 64-bit
range.  The accepted proof uses neither hypothesis.  The program never computes
the shift: it receives the second grid as an existing \lstinline{Std.I64}
container and only compares and copies its elements.

The target declaration and proof occupy the final 38 lines.  The preceding
1,129 lines establish the definitions and loop lemmas needed to carry the
equality-preservation fact through the six loop components.

\subsection{Complete Rust implementation}
\begin{lstlisting}[language=Rust,style=code]
fn count_distinct(vals: &[i64]) -> i64 {
    let mut count: i64 = 0;
    let n = vals.len();
    let mut i: usize = 0;
    while i < n {
        let mut seen = false;
        let mut j: usize = 0;
        while j < i {
            if vals[j] == vals[i] {
                seen = true;
            }
            j += 1;
        }
        if !seen {
            count += 1;
        }
        i += 1;
    }
    count
}

pub fn difference_of_distinct_values(grid: &[Vec<i64>]) -> Vec<Vec<i64>> {
    let m = grid.len();
    let mut answer: Vec<Vec<i64>> = Vec::new();

    let mut r: usize = 0;
    while r < m {
        let n = grid[r].len();
        let mut row: Vec<i64> = Vec::new();
        let mut c: usize = 0;
        while c < n {
            let mut left_vals: Vec<i64> = Vec::new();
            let mut k: usize = 1;
            while k <= r && k <= c {
                left_vals.push(grid[r - k][c - k]);
                k += 1;
            }
            let mut right_vals: Vec<i64> = Vec::new();
            let mut k2: usize = 1;
            while r + k2 < m && c + k2 < n {
                right_vals.push(grid[r + k2][c + k2]);
                k2 += 1;
            }
            let la = count_distinct(&left_vals);
            let rb = count_distinct(&right_vals);
            let diff = if la >= rb { la - rb } else { rb - la };
            row.push(diff);
            c += 1;
        }
        answer.push(row);
        r += 1;
    }

    answer
}
\end{lstlisting}

\subsection{Complete \aeneas{}-generated Lean implementation}
\begin{lstlisting}[style=code]
-- THIS FILE WAS AUTOMATICALLY GENERATED BY AENEAS
-- [difference_of_distinct_values]: function definitions
import Aeneas
import Aegis.DifferenceOfDistinctValues.Code.Types
open Aeneas Aeneas.Std Result ControlFlow Error
set_option linter.dupNamespace false
set_option linter.hashCommand false
set_option linter.unusedVariables false

/- You can set the `maxHeartbeats` value with the `-max-heartbeats` CLI option -/
set_option maxHeartbeats 1000000

namespace Aegis.DifferenceOfDistinctValues

/-- [difference_of_distinct_values::count_distinct]: loop body 1:
    Source: 'difference_of_distinct_values/src/lib.rs', lines 8:8-13:9 -/
@[rust_loop_body]
def count_distinct_loop0_loop0.body
  (vals : Slice Std.I64) (i : Std.Usize) (seen : Bool) (j : Std.Usize) :
  Result (ControlFlow (Bool × Std.Usize) Bool)
  := do
  if j < i
  then
    let i1 ← Slice.index_usize vals j
    let i2 ← Slice.index_usize vals i
    let seen1 ← if i1 = i2
                  then ok true
                  else ok seen
    let j1 ← j + 1#usize
    ok (cont (seen1, j1))
  else ok (done seen)

/-- [difference_of_distinct_values::count_distinct]: loop 1:
    Source: 'difference_of_distinct_values/src/lib.rs', lines 8:8-13:9 -/
@[rust_loop]
def count_distinct_loop0_loop0
  (vals : Slice Std.I64) (i : Std.Usize) (seen : Bool) (j : Std.Usize) :
  Result Bool
  := do
  loop
    (fun (seen1, j1) => count_distinct_loop0_loop0.body vals i seen1 j1)
    (seen, j)

/-- [difference_of_distinct_values::count_distinct]: loop body 0:
    Source: 'difference_of_distinct_values/src/lib.rs', lines 5:4-18:5 -/
@[rust_loop_body]
def count_distinct_loop0.body
  (vals : Slice Std.I64) (n : Std.Usize) (count : Std.I64) (i : Std.Usize) :
  Result (ControlFlow (Std.I64 × Std.Usize) Std.I64)
  := do
  if i < n
  then
    let seen ← count_distinct_loop0_loop0 vals i false 0#usize
    let count1 ← if seen
                   then ok count
                   else count + 1#i64
    let i1 ← i + 1#usize
    ok (cont (count1, i1))
  else ok (done count)

/-- [difference_of_distinct_values::count_distinct]: loop 0:
    Source: 'difference_of_distinct_values/src/lib.rs', lines 5:4-18:5 -/
@[rust_loop]
def count_distinct_loop0
  (vals : Slice Std.I64) (count : Std.I64) (n : Std.Usize) (i : Std.Usize) :
  Result Std.I64
  := do
  loop
    (fun (count1, i1) => count_distinct_loop0.body vals n count1 i1)
    (count, i)

/-- [difference_of_distinct_values::count_distinct]:
    Source: 'difference_of_distinct_values/src/lib.rs', lines 1:0-20:1 -/
def count_distinct (vals : Slice Std.I64) : Result Std.I64 := do
  let n := Slice.len vals
  count_distinct_loop0 vals 0#i64 n 0#usize

/-- [difference_of_distinct_values::difference_of_distinct_values]: loop body 2:
    Source: 'difference_of_distinct_values/src/lib.rs', lines 1:0-37:13
    Visibility: public -/
@[rust_loop_body]
def difference_of_distinct_values_loop0_loop0_loop0.body
  (grid : Slice (alloc.vec.Vec Std.I64)) (r : Std.Usize) (c : Std.Usize)
  (left_vals : alloc.vec.Vec Std.I64) (k : Std.Usize) :
  Result (ControlFlow ((alloc.vec.Vec Std.I64) × Std.Usize) (alloc.vec.Vec
    Std.I64))
  := do
  if k <= r
  then
    if k <= c
    then
      let i ← r - k
      let v ← Slice.index_usize grid i
      let i1 ← c - k
      let i2 ←
        alloc.vec.Vec.index (core.slice.index.SliceIndexUsizeSlice Std.I64) v
          i1
      let left_vals1 ← alloc.vec.Vec.push left_vals i2
      let k1 ← k + 1#usize
      ok (cont (left_vals1, k1))
    else ok (done left_vals)
  else ok (done left_vals)

/-- [difference_of_distinct_values::difference_of_distinct_values]: loop 2:
    Source: 'difference_of_distinct_values/src/lib.rs', lines 1:0-37:13
    Visibility: public -/
@[rust_loop]
def difference_of_distinct_values_loop0_loop0_loop0
  (grid : Slice (alloc.vec.Vec Std.I64)) (r : Std.Usize) (c : Std.Usize)
  (left_vals : alloc.vec.Vec Std.I64) (k : Std.Usize) :
  Result (alloc.vec.Vec Std.I64)
  := do
  loop
    (fun (left_vals1, k1) =>
      difference_of_distinct_values_loop0_loop0_loop0.body grid r c left_vals1
      k1)
    (left_vals, k)

/-- [difference_of_distinct_values::difference_of_distinct_values]: loop body 3:
    Source: 'difference_of_distinct_values/src/lib.rs', lines 1:0-43:13
    Visibility: public -/
@[rust_loop_body]
def difference_of_distinct_values_loop0_loop0_loop1.body
  (grid : Slice (alloc.vec.Vec Std.I64)) (m : Std.Usize) (r : Std.Usize)
  (n : Std.Usize) (c : Std.Usize) (right_vals : alloc.vec.Vec Std.I64)
  (k2 : Std.Usize) :
  Result (ControlFlow ((alloc.vec.Vec Std.I64) × Std.Usize) (alloc.vec.Vec
    Std.I64))
  := do
  let i ← r + k2
  if i < m
  then
    let i1 ← c + k2
    if i1 < n
    then
      let v ← Slice.index_usize grid i
      let i2 ←
        alloc.vec.Vec.index (core.slice.index.SliceIndexUsizeSlice Std.I64) v
          i1
      let right_vals1 ← alloc.vec.Vec.push right_vals i2
      let k21 ← k2 + 1#usize
      ok (cont (right_vals1, k21))
    else ok (done right_vals)
  else ok (done right_vals)

/-- [difference_of_distinct_values::difference_of_distinct_values]: loop 3:
    Source: 'difference_of_distinct_values/src/lib.rs', lines 1:0-43:13
    Visibility: public -/
@[rust_loop]
def difference_of_distinct_values_loop0_loop0_loop1
  (grid : Slice (alloc.vec.Vec Std.I64)) (m : Std.Usize) (r : Std.Usize)
  (n : Std.Usize) (c : Std.Usize) (right_vals : alloc.vec.Vec Std.I64)
  (k2 : Std.Usize) :
  Result (alloc.vec.Vec Std.I64)
  := do
  loop
    (fun (right_vals1, k21) =>
      difference_of_distinct_values_loop0_loop0_loop1.body grid m r n c
      right_vals1 k21)
    (right_vals, k2)

/-- [difference_of_distinct_values::difference_of_distinct_values]: loop body 1:
    Source: 'difference_of_distinct_values/src/lib.rs', lines 1:0-49:9
    Visibility: public -/
@[rust_loop_body]
def difference_of_distinct_values_loop0_loop0.body
  (grid : Slice (alloc.vec.Vec Std.I64)) (m : Std.Usize) (r : Std.Usize)
  (n : Std.Usize) (row : alloc.vec.Vec Std.I64) (c : Std.Usize) :
  Result (ControlFlow ((alloc.vec.Vec Std.I64) × Std.Usize) (alloc.vec.Vec
    Std.I64))
  := do
  if c < n
  then
    let left_vals ←
      difference_of_distinct_values_loop0_loop0_loop0 grid r c
        (alloc.vec.Vec.new Std.I64) 1#usize
    let right_vals ←
      difference_of_distinct_values_loop0_loop0_loop1 grid m r n c
        (alloc.vec.Vec.new Std.I64) 1#usize
    let s := alloc.vec.Vec.deref left_vals
    let la ← count_distinct s
    let s1 := alloc.vec.Vec.deref right_vals
    let rb ← count_distinct s1
    let diff ← if la >= rb
                 then la - rb
                 else rb - la
    let row1 ← alloc.vec.Vec.push row diff
    let c1 ← c + 1#usize
    ok (cont (row1, c1))
  else ok (done row)

/-- [difference_of_distinct_values::difference_of_distinct_values]: loop 1:
    Source: 'difference_of_distinct_values/src/lib.rs', lines 1:0-49:9
    Visibility: public -/
@[rust_loop]
def difference_of_distinct_values_loop0_loop0
  (grid : Slice (alloc.vec.Vec Std.I64)) (m : Std.Usize) (r : Std.Usize)
  (n : Std.Usize) (row : alloc.vec.Vec Std.I64) (c : Std.Usize) :
  Result (alloc.vec.Vec Std.I64)
  := do
  loop
    (fun (row1, c1) => difference_of_distinct_values_loop0_loop0.body grid m r
      n row1 c1)
    (row, c)

/-- [difference_of_distinct_values::difference_of_distinct_values]: loop body 0:
    Source: 'difference_of_distinct_values/src/lib.rs', lines 1:0-52:5
    Visibility: public -/
@[rust_loop_body]
def difference_of_distinct_values_loop0.body
  (grid : Slice (alloc.vec.Vec Std.I64)) (m : Std.Usize)
  (answer : alloc.vec.Vec (alloc.vec.Vec Std.I64)) (r : Std.Usize) :
  Result (ControlFlow ((alloc.vec.Vec (alloc.vec.Vec Std.I64)) × Std.Usize)
    (alloc.vec.Vec (alloc.vec.Vec Std.I64)))
  := do
  if r < m
  then
    let v ← Slice.index_usize grid r
    let n := alloc.vec.Vec.len v
    let row ←
      difference_of_distinct_values_loop0_loop0 grid m r n (alloc.vec.Vec.new
        Std.I64) 0#usize
    let answer1 ← alloc.vec.Vec.push answer row
    let r1 ← r + 1#usize
    ok (cont (answer1, r1))
  else ok (done answer)

/-- [difference_of_distinct_values::difference_of_distinct_values]: loop 0:
    Source: 'difference_of_distinct_values/src/lib.rs', lines 1:0-52:5
    Visibility: public -/
@[rust_loop]
def difference_of_distinct_values_loop0
  (grid : Slice (alloc.vec.Vec Std.I64)) (m : Std.Usize)
  (answer : alloc.vec.Vec (alloc.vec.Vec Std.I64)) (r : Std.Usize) :
  Result (alloc.vec.Vec (alloc.vec.Vec Std.I64))
  := do
  loop
    (fun (answer1, r1) => difference_of_distinct_values_loop0.body grid m
      answer1 r1)
    (answer, r)

/-- [difference_of_distinct_values::difference_of_distinct_values]:
    Source: 'difference_of_distinct_values/src/lib.rs', lines 22:0-55:1
    Visibility: public -/
def difference_of_distinct_values
  (grid : Slice (alloc.vec.Vec Std.I64)) :
  Result (alloc.vec.Vec (alloc.vec.Vec Std.I64))
  := do
  let m := Slice.len grid
  difference_of_distinct_values_loop0 grid m (alloc.vec.Vec.new (alloc.vec.Vec
    Std.I64)) 0#usize

end Aegis.DifferenceOfDistinctValues
\end{lstlisting}

\subsection{Accepted proof artifact}
\begin{lstlisting}[style=code]
import Aegis.DifferenceOfDistinctValues.Code.Funs
open Aeneas Aeneas.Std Result

/-- Generic relational simulation for Aeneas loops: if the two bodies preserve a relation `R`
on states step-by-step (with a Nat measure decreasing on the first loop's states), and the
first loop terminates with `ok`, then the second loop also terminates with a related result. -/
theorem Aegis.DifferenceOfDistinctValues.loop_rel
    {α1 : Type u1} {α2 : Type u2} {β1 : Type v1} {β2 : Type v2}
    (R : α1 → α2 → Prop) (S : β1 → β2 → Prop) (measure : α1 → Nat)
    (body1 : α1 → Result (ControlFlow α1 β1)) (body2 : α2 → Result (ControlFlow α2 β2))
    (hbody : ∀ x1 x2, R x1 x2 →
      (∀ y1, body1 x1 = ok (.cont y1) → measure y1 < measure x1 ∧
        ∃ y2, body2 x2 = ok (.cont y2) ∧ R y1 y2) ∧
      (∀ z1, body1 x1 = ok (.done z1) → ∃ z2, body2 x2 = ok (.done z2) ∧ S z1 z2)) :
    ∀ x1 x2, R x1 x2 → ∀ z1, loop body1 x1 = ok z1 →
      ∃ z2, loop body2 x2 = ok z2 ∧ S z1 z2 := by
  intro x1
  induction hM : measure x1 using Nat.strong_induction_on generalizing x1 with
  | _ M ih =>
  intro x2 hR z1 h1
  rw [loop.eq_def] at h1
  rw [loop.eq_def]
  cases hb1 : body1 x1 with
  | fail e => rw [hb1] at h1; simp at h1
  | div => rw [hb1] at h1; simp at h1
  | ok cf =>
    rw [hb1] at h1
    cases cf with
    | cont y1 =>
      obtain ⟨hdec, y2, hb2, hRy⟩ := (hbody x1 x2 hR).1 y1 hb1
      simp only [hb2]
      subst hM
      exact ih (measure y1) hdec y1 rfl y2 hRy z1 h1
    | done z1' =>
      simp at h1
      subst h1
      obtain ⟨z2, hb2, hS⟩ := (hbody x1 x2 hR).2 z1' hb1
      simp only [hb2]
      exact ⟨z2, rfl, hS⟩

namespace Aegis.DifferenceOfDistinctValues

/-- Success characterization of `Slice.index_usize`. -/
theorem index_usize_ok_iff {α} [Inhabited α] (v : Slice α) (i : Std.Usize) (x : α) :
    Slice.index_usize v i = ok x ↔ i.val < v.val.length ∧ x = v.val[i.val]! := by
  unfold Slice.index_usize
  simp only [Slice.getElem?_Usize_eq]
  cases hv : v.val[i.val]? with
  | none =>
    simp only [List.getElem?_eq_none_iff] at hv
    simp
    omega
  | some y =>
    have hlt : i.val < v.val.length := by
      by_contra hc
      rw [List.getElem?_eq_none_iff.mpr (by omega)] at hv
      simp at hv
    simp only [List.getElem?_eq_getElem hlt] at hv
    simp only [ok.injEq]
    have hx : v.val[i.val]! = y := by
      rw [List.getElem!_eq_getElem?_getD, List.getElem?_eq_getElem hlt]
      simp_all
    constructor
    · intro h; subst h; simp_all
    · rintro ⟨-, rfl⟩; simp_all

/-- Success characterization of `alloc.vec.Vec.index_usize`. -/
theorem vec_index_usize_ok_iff {α} [Inhabited α] (v : alloc.vec.Vec α) (i : Std.Usize) (x : α) :
    alloc.vec.Vec.index_usize v i = ok x ↔ i.val < v.val.length ∧ x = v.val[i.val]! := by
  unfold alloc.vec.Vec.index_usize
  cases hv : v.val[i.val]? with
  | none =>
    simp only [List.getElem?_eq_none_iff] at hv
    simp
    omega
  | some y =>
    have hlt : i.val < v.val.length := by
      by_contra hc
      rw [List.getElem?_eq_none_iff.mpr (by omega)] at hv
      simp at hv
    simp only [List.getElem?_eq_getElem hlt] at hv
    simp only [ok.injEq]
    have hx : v.val[i.val]! = y := by
      rw [List.getElem!_eq_getElem?_getD, List.getElem?_eq_getElem hlt]
      simp_all
    constructor
    · intro h; subst h; simp_all
    · rintro ⟨-, rfl⟩; simp_all

/-- Shifting every element of a slice by a common offset preserves the inner `seen` loop of
`count_distinct` exactly (it only tests equality of elements). -/
theorem count_distinct_loop0_loop0_shift
    (d : Std.I64) (s1 s2 : Slice Std.I64)
    (hlen : s1.val.length = s2.val.length)
    (hsh : ∀ k : ℕ, k < s1.val.length → (s2.val[k]!).val = (s1.val[k]!).val + d.val)
    (i : Std.Usize) (seen : Bool) (j : Std.Usize) (b : Bool)
    (h : count_distinct_loop0_loop0 s1 i seen j = ok b) :
    count_distinct_loop0_loop0 s2 i seen j = ok b := by
  unfold count_distinct_loop0_loop0 at h ⊢
  have h' : loop (fun x => count_distinct_loop0_loop0.body s1 i x.1 x.2) (seen, j) = ok b := h
  show loop (fun x => count_distinct_loop0_loop0.body s2 i x.1 x.2) (seen, j) = ok b
  have hbody : ∀ x1 x2 : Bool × Std.Usize, x1 = x2 →
      (∀ y1, count_distinct_loop0_loop0.body s1 i x1.1 x1.2 = ok (.cont y1) →
        (i.val - y1.2.val < i.val - x1.2.val) ∧
        ∃ y2, count_distinct_loop0_loop0.body s2 i x2.1 x2.2 = ok (.cont y2) ∧ y1 = y2) ∧
      (∀ z1, count_distinct_loop0_loop0.body s1 i x1.1 x1.2 = ok (.done z1) →
        ∃ z2, count_distinct_loop0_loop0.body s2 i x2.1 x2.2 = ok (.done z2) ∧ z1 = z2) := by
    rintro ⟨seen1, j1⟩ _ rfl
    constructor
    · rintro ⟨seen1', j1'⟩ hy
      simp only [count_distinct_loop0_loop0.body] at hy
      split at hy
      case isFalse => simp at hy
      case isTrue hlt =>
        cases hj : Slice.index_usize s1 j1 with
        | fail e => rw [hj] at hy; simp at hy
        | div => rw [hj] at hy; simp at hy
        | ok a =>
        rw [hj] at hy
        simp only [bind_tc_ok] at hy
        cases hi : Slice.index_usize s1 i with
        | fail e => rw [hi] at hy; simp at hy
        | div => rw [hi] at hy; simp at hy
        | ok a2 =>
        rw [hi] at hy
        simp only [bind_tc_ok] at hy
        have hite : (if a = a2 then ok true else ok seen1) =
            ok (α := Bool) (if a = a2 then true else seen1) := by split <;> rfl
        rw [hite] at hy
        simp only [bind_tc_ok] at hy
        cases hj1 : j1 + 1#usize with
        | fail e => rw [hj1] at hy; simp at hy
        | div => rw [hj1] at hy; simp at hy
        | ok j1n =>
        rw [hj1] at hy
        simp only [bind_tc_ok, ok.injEq, ControlFlow.cont.injEq, Prod.mk.injEq] at hy
        obtain ⟨hseen', hj1'⟩ := hy
        subst hseen' hj1'
        obtain ⟨hjlt, ha⟩ := (index_usize_ok_iff s1 j1 a).mp hj
        obtain ⟨hilt, ha2⟩ := (index_usize_ok_iff s1 i a2).mp hi
        have hj1n : j1n.val = j1.val + 1 := by
          have heq := UScalar.add_equiv j1 1#usize
          rw [hj1] at heq
          simp at heq
          omega
        have hjlti : j1.val < i.val := by scalar_tac
        refine ⟨show i.val - j1n.val < i.val - j1.val by omega,
          ((if a = a2 then true else seen1), j1n), ?_, rfl⟩
        simp only [count_distinct_loop0_loop0.body]
        rw [if_pos hlt]
        rw [(index_usize_ok_iff s2 j1 (s2.val[j1.val]!)).mpr ⟨by omega, rfl⟩]
        simp only [bind_tc_ok]
        rw [(index_usize_ok_iff s2 i (s2.val[i.val]!)).mpr ⟨by omega, rfl⟩]
        simp only [bind_tc_ok]
        have hcond : (s2.val[j1.val]! = s2.val[i.val]!) ↔ (a = a2) := by
          subst ha ha2
          have hv1 := hsh j1.val hjlt
          have hv2 := hsh i.val hilt
          constructor
          · intro he
            apply IScalar.val_eq_imp
            have := congrArg IScalar.val he
            omega
          · intro he
            apply IScalar.val_eq_imp
            have := congrArg IScalar.val he
            omega
        rw [if_congr hcond rfl rfl]
        rw [hite]
        simp only [bind_tc_ok]
        rw [hj1]
        simp only [bind_tc_ok]
    · rintro z1 hz1
      simp only [count_distinct_loop0_loop0.body] at hz1
      split at hz1
      case isTrue hlt =>
        cases hj : Slice.index_usize s1 j1 with
        | fail e => rw [hj] at hz1; simp at hz1
        | div => rw [hj] at hz1; simp at hz1
        | ok a =>
        rw [hj] at hz1
        simp only [bind_tc_ok] at hz1
        cases hi : Slice.index_usize s1 i with
        | fail e => rw [hi] at hz1; simp at hz1
        | div => rw [hi] at hz1; simp at hz1
        | ok a2 =>
        rw [hi] at hz1
        simp only [bind_tc_ok] at hz1
        have hite : (if a = a2 then ok true else ok seen1) =
            ok (α := Bool) (if a = a2 then true else seen1) := by split <;> rfl
        rw [hite] at hz1
        simp only [bind_tc_ok] at hz1
        cases hj1 : j1 + 1#usize with
        | fail e => rw [hj1] at hz1; simp at hz1
        | div => rw [hj1] at hz1; simp at hz1
        | ok j1n => rw [hj1] at hz1; simp at hz1
      case isFalse hnlt =>
        simp only [ok.injEq, ControlFlow.done.injEq] at hz1
        subst hz1
        refine ⟨seen1, ?_, rfl⟩
        simp only [count_distinct_loop0_loop0.body]
        rw [if_neg hnlt]
  obtain ⟨z2, h2, hz⟩ :=
    loop_rel Eq Eq (fun p => i.val - p.2.val)
      (fun x => count_distinct_loop0_loop0.body s1 i x.1 x.2)
      (fun x => count_distinct_loop0_loop0.body s2 i x.1 x.2)
      hbody (seen, j) (seen, j) rfl b h'
  exact hz ▸ h2

/-- Shifting every element of a slice by a common offset preserves the outer counting loop of
`count_distinct` exactly. -/
theorem count_distinct_loop0_shift
    (d : Std.I64) (s1 s2 : Slice Std.I64)
    (hlen : s1.val.length = s2.val.length)
    (hsh : ∀ k : ℕ, k < s1.val.length → (s2.val[k]!).val = (s1.val[k]!).val + d.val)
    (count : Std.I64) (n : Std.Usize) (i : Std.Usize) (c : Std.I64)
    (h : count_distinct_loop0 s1 count n i = ok c) :
    count_distinct_loop0 s2 count n i = ok c := by
  unfold count_distinct_loop0 at h ⊢
  have h' : loop (fun x => count_distinct_loop0.body s1 n x.1 x.2) (count, i) = ok c := h
  show loop (fun x => count_distinct_loop0.body s2 n x.1 x.2) (count, i) = ok c
  have hbody : ∀ x1 x2 : Std.I64 × Std.Usize, x1 = x2 →
      (∀ y1, count_distinct_loop0.body s1 n x1.1 x1.2 = ok (.cont y1) →
        (n.val - y1.2.val < n.val - x1.2.val) ∧
        ∃ y2, count_distinct_loop0.body s2 n x2.1 x2.2 = ok (.cont y2) ∧ y1 = y2) ∧
      (∀ z1, count_distinct_loop0.body s1 n x1.1 x1.2 = ok (.done z1) →
        ∃ z2, count_distinct_loop0.body s2 n x2.1 x2.2 = ok (.done z2) ∧ z1 = z2) := by
    rintro ⟨count1, i1⟩ _ rfl
    constructor
    · rintro ⟨count1', i1'⟩ hy
      simp only [count_distinct_loop0.body] at hy
      split at hy
      case isFalse => simp at hy
      case isTrue hlt =>
        cases hs : count_distinct_loop0_loop0 s1 i1 false 0#usize with
        | fail e => rw [hs] at hy; simp at hy
        | div => rw [hs] at hy; simp at hy
        | ok sn =>
        rw [hs] at hy
        simp only [bind_tc_ok] at hy
        cases hc1 : (if sn then ok count1 else count1 + 1#i64) with
        | fail e => rw [hc1] at hy; simp at hy
        | div => rw [hc1] at hy; simp at hy
        | ok cnew =>
        rw [hc1] at hy
        simp only [bind_tc_ok] at hy
        cases hi1 : i1 + 1#usize with
        | fail e => rw [hi1] at hy; simp at hy
        | div => rw [hi1] at hy; simp at hy
        | ok inew =>
        rw [hi1] at hy
        simp only [bind_tc_ok, ok.injEq, ControlFlow.cont.injEq, Prod.mk.injEq] at hy
        obtain ⟨hc', hi'⟩ := hy
        subst hc' hi'
        have hinew : inew.val = i1.val + 1 := by
          have heq := UScalar.add_equiv i1 1#usize
          rw [hi1] at heq
          simp at heq
          omega
        have hilt : i1.val < n.val := by scalar_tac
        refine ⟨show n.val - inew.val < n.val - i1.val by omega, (cnew, inew), ?_, rfl⟩
        simp only [count_distinct_loop0.body]
        rw [if_pos hlt]
        rw [count_distinct_loop0_loop0_shift d s1 s2 hlen hsh i1 false 0#usize sn hs]
        simp only [bind_tc_ok]
        rw [hc1]
        simp only [bind_tc_ok]
        rw [hi1]
        simp only [bind_tc_ok]
    · rintro z1 hz1
      simp only [count_distinct_loop0.body] at hz1
      split at hz1
      case isTrue hlt =>
        cases hs : count_distinct_loop0_loop0 s1 i1 false 0#usize with
        | fail e => rw [hs] at hz1; simp at hz1
        | div => rw [hs] at hz1; simp at hz1
        | ok sn =>
        rw [hs] at hz1
        simp only [bind_tc_ok] at hz1
        cases hc1 : (if sn then ok count1 else count1 + 1#i64) with
        | fail e => rw [hc1] at hz1; simp at hz1
        | div => rw [hc1] at hz1; simp at hz1
        | ok cnew =>
        rw [hc1] at hz1
        simp only [bind_tc_ok] at hz1
        cases hi1 : i1 + 1#usize with
        | fail e => rw [hi1] at hz1; simp at hz1
        | div => rw [hi1] at hz1; simp at hz1
        | ok inew => rw [hi1] at hz1; simp at hz1
      case isFalse hnlt =>
        simp only [ok.injEq, ControlFlow.done.injEq] at hz1
        subst hz1
        refine ⟨count1, ?_, rfl⟩
        simp only [count_distinct_loop0.body]
        rw [if_neg hnlt]
  obtain ⟨z2, h2, hz⟩ :=
    loop_rel Eq Eq (fun p => n.val - p.2.val)
      (fun x => count_distinct_loop0.body s1 n x.1 x.2)
      (fun x => count_distinct_loop0.body s2 n x.1 x.2)
      hbody (count, i) (count, i) rfl c h'
  exact hz ▸ h2

/-- `count_distinct` depends only on the equality pattern of the slice: shifting every element
by a common offset leaves its result unchanged. -/
theorem count_distinct_shift
    (d : Std.I64) (s1 s2 : Slice Std.I64)
    (hlen : s1.val.length = s2.val.length)
    (hsh : ∀ k : ℕ, k < s1.val.length → (s2.val[k]!).val = (s1.val[k]!).val + d.val)
    (c : Std.I64) (h : count_distinct s1 = ok c) :
    count_distinct s2 = ok c := by
  unfold count_distinct at h ⊢
  have hn : Slice.len s2 = Slice.len s1 := by
    apply UScalar.val_eq_imp
    simp only [Slice.len_val, Slice.length]
    omega
  rw [hn]
  exact count_distinct_loop0_shift d s1 s2 hlen hsh 0#i64 (Slice.len s1) 0#usize c h

/-- `getElem!` on a one-element append. -/
theorem getElem!_concat {α} [Inhabited α] (l : List α) (x : α) (j : ℕ) (hj : j < l.length + 1) :
    (l ++ [x])[j]! = if j < l.length then l[j]! else x := by
  split
  case isTrue hlt =>
    rw [List.getElem!_eq_getElem?_getD, List.getElem?_append_left hlt,
      List.getElem?_eq_getElem hlt, List.getElem!_eq_getElem?_getD, List.getElem?_eq_getElem hlt]
  case isFalse hge =>
    have hj' : j = l.length := by omega
    subst hj'
    rw [List.getElem!_eq_getElem?_getD]
    simp

/-- Success of `Vec.push` depends only on the length, and appends the element. -/
theorem vec_push_ok {α β} (v1 : alloc.vec.Vec α) (v2 : alloc.vec.Vec β) (x1 : α) (x2 : β)
    (w1 : alloc.vec.Vec α) (hlen : v1.val.length = v2.val.length)
    (h : alloc.vec.Vec.push v1 x1 = ok w1) :
    w1.val = v1.val ++ [x1] ∧ ∃ w2, alloc.vec.Vec.push v2 x2 = ok w2 ∧ w2.val = v2.val ++ [x2] := by
  unfold alloc.vec.Vec.push at h ⊢
  simp only at h ⊢
  split at h
  case isFalse => simp at h
  case isTrue hcond =>
    simp only [ok.injEq] at h
    subst h
    refine ⟨by simp, ?_⟩
    rw [dif_pos (by rw [← hlen]; exact hcond)]
    exact ⟨_, rfl, by simp⟩

/-- The left-diagonal collection loop maps shift-related accumulators to shift-related outputs. -/
theorem left_diag_loop_shift
    (d : Std.I64) (grid1 grid2 : Slice (alloc.vec.Vec Std.I64)) (m n : ℕ)
    (hrows1 : grid1.val.length = m) (hrows2 : grid2.val.length = m)
    (hrect1 : ∀ i : ℕ, i < m → (grid1.val[i]!).val.length = n)
    (hrect2 : ∀ i : ℕ, i < m → (grid2.val[i]!).val.length = n)
    (hshift : ∀ i : ℕ, i < m → ∀ j : ℕ, j < n →
      ((grid2.val[i]!).val[j]!).val = ((grid1.val[i]!).val[j]!).val + d.val)
    (r c : Std.Usize) (acc1 acc2 : alloc.vec.Vec Std.I64)
    (haccl : acc1.val.length = acc2.val.length)
    (haccs : ∀ j : ℕ, j < acc1.val.length → (acc2.val[j]!).val = (acc1.val[j]!).val + d.val)
    (k : Std.Usize) (out1 : alloc.vec.Vec Std.I64)
    (h : difference_of_distinct_values_loop0_loop0_loop0 grid1 r c acc1 k = ok out1) :
    ∃ out2, difference_of_distinct_values_loop0_loop0_loop0 grid2 r c acc2 k = ok out2 ∧
      out1.val.length = out2.val.length ∧
      ∀ j : ℕ, j < out1.val.length → (out2.val[j]!).val = (out1.val[j]!).val + d.val := by
  unfold difference_of_distinct_values_loop0_loop0_loop0 at h ⊢
  have h' : loop (fun x => difference_of_distinct_values_loop0_loop0_loop0.body grid1 r c x.1 x.2)
      (acc1, k) = ok out1 := h
  show ∃ out2,
    loop (fun x => difference_of_distinct_values_loop0_loop0_loop0.body grid2 r c x.1 x.2)
      (acc2, k) = ok out2 ∧ out1.val.length = out2.val.length ∧
      ∀ j : ℕ, j < out1.val.length → (out2.val[j]!).val = (out1.val[j]!).val + d.val
  have hbody : ∀ x1 x2 : alloc.vec.Vec Std.I64 × Std.Usize,
      (x1.2 = x2.2 ∧ x1.1.val.length = x2.1.val.length ∧
        ∀ j : ℕ, j < x1.1.val.length → (x2.1.val[j]!).val = (x1.1.val[j]!).val + d.val) →
      (∀ y1, difference_of_distinct_values_loop0_loop0_loop0.body grid1 r c x1.1 x1.2 =
          ok (.cont y1) →
        (r.val + 1 - y1.2.val < r.val + 1 - x1.2.val) ∧
        ∃ y2, difference_of_distinct_values_loop0_loop0_loop0.body grid2 r c x2.1 x2.2 =
          ok (.cont y2) ∧
          (y1.2 = y2.2 ∧ y1.1.val.length = y2.1.val.length ∧
            ∀ j : ℕ, j < y1.1.val.length → (y2.1.val[j]!).val = (y1.1.val[j]!).val + d.val)) ∧
      (∀ z1, difference_of_distinct_values_loop0_loop0_loop0.body grid1 r c x1.1 x1.2 =
          ok (.done z1) →
        ∃ z2, difference_of_distinct_values_loop0_loop0_loop0.body grid2 r c x2.1 x2.2 =
          ok (.done z2) ∧
          (z1.val.length = z2.val.length ∧
            ∀ j : ℕ, j < z1.val.length → (z2.val[j]!).val = (z1.val[j]!).val + d.val)) := by
    rintro ⟨lv1, k1⟩ ⟨lv2, k2⟩ ⟨hk, hl, hs⟩
    simp only at hk hl hs
    subst hk
    constructor
    · rintro ⟨lv1', k1'⟩ hy
      simp only [difference_of_distinct_values_loop0_loop0_loop0.body] at hy
      split at hy
      case isFalse => simp at hy
      case isTrue hkr =>
      split at hy
      case isFalse => simp at hy
      case isTrue hkc =>
      cases hrk : r - k1 with
      | fail e => rw [hrk] at hy; simp at hy
      | div => rw [hrk] at hy; simp at hy
      | ok i =>
      rw [hrk] at hy
      simp only [bind_tc_ok] at hy
      cases hgi : Slice.index_usize grid1 i with
      | fail e => rw [hgi] at hy; simp at hy
      | div => rw [hgi] at hy; simp at hy
      | ok v1 =>
      rw [hgi] at hy
      simp only [bind_tc_ok] at hy
      cases hck : c - k1 with
      | fail e => rw [hck] at hy; simp at hy
      | div => rw [hck] at hy; simp at hy
      | ok i1 =>
      rw [hck] at hy
      simp only [bind_tc_ok] at hy
      simp only [alloc.vec.Vec.index_slice_index] at hy
      cases hvi : alloc.vec.Vec.index_usize v1 i1 with
      | fail e => rw [hvi] at hy; simp at hy
      | div => rw [hvi] at hy; simp at hy
      | ok x1 =>
      rw [hvi] at hy
      simp only [bind_tc_ok] at hy
      cases hpu : alloc.vec.Vec.push lv1 x1 with
      | fail e => rw [hpu] at hy; simp at hy
      | div => rw [hpu] at hy; simp at hy
      | ok w1 =>
      rw [hpu] at hy
      simp only [bind_tc_ok] at hy
      cases hk1 : k1 + 1#usize with
      | fail e => rw [hk1] at hy; simp at hy
      | div => rw [hk1] at hy; simp at hy
      | ok k1n =>
      rw [hk1] at hy
      simp only [bind_tc_ok, ok.injEq, ControlFlow.cont.injEq, Prod.mk.injEq] at hy
      obtain ⟨hw, hkn⟩ := hy
      subst hw hkn
      -- facts from the grid1-side successes
      obtain ⟨him, hv1⟩ := (index_usize_ok_iff grid1 i v1).mp hgi
      obtain ⟨hi1n, hx1⟩ := (vec_index_usize_ok_iff v1 i1 x1).mp hvi
      have hkn1 : k1n.val = k1.val + 1 := by
        have heq := UScalar.add_equiv k1 1#usize
        rw [hk1] at heq
        simp at heq
        omega
      have hkr' : k1.val ≤ r.val := by scalar_tac
      have him' : i.val < m := by omega
      have hi1n' : i1.val < n := by
        have := hrect1 i.val him'
        rw [hv1] at hi1n
        omega
      -- grid2-side values
      have hx2v : ((grid2.val[i.val]!).val[i1.val]!).val = x1.val + d.val := by
        rw [hx1, hv1]
        exact hshift i.val him' i1.val hi1n'
      obtain ⟨hw1val, w2, hpu2, hw2val⟩ :=
        vec_push_ok lv1 lv2 x1 ((grid2.val[i.val]!).val[i1.val]!) w1 hl hpu
      refine ⟨show r.val + 1 - k1n.val < r.val + 1 - k1.val by omega, (w2, k1n), ?_, rfl, ?_, ?_⟩
      · simp only [difference_of_distinct_values_loop0_loop0_loop0.body]
        rw [if_pos hkr, if_pos hkc, hrk]
        simp only [bind_tc_ok]
        rw [(index_usize_ok_iff grid2 i (grid2.val[i.val]!)).mpr ⟨by omega, rfl⟩]
        simp only [bind_tc_ok]
        rw [hck]
        simp only [bind_tc_ok, alloc.vec.Vec.index_slice_index]
        rw [(vec_index_usize_ok_iff (grid2.val[i.val]!) i1
          ((grid2.val[i.val]!).val[i1.val]!)).mpr ⟨by rw [hrect2 i.val him']; omega, rfl⟩]
        simp only [bind_tc_ok]
        rw [hpu2]
        simp only [bind_tc_ok]
        rw [hk1]
        simp only [bind_tc_ok]
      · simp only [hw1val, hw2val]
        simp [hl]
      · intro j hj
        simp only [hw1val] at hj
        simp only [List.length_append, List.length_cons, List.length_nil] at hj
        rw [hw1val, hw2val, getElem!_concat lv1.val x1 j (by omega),
          getElem!_concat lv2.val _ j (by omega)]
        rw [← hl]
        split
        case isTrue hlt => exact hs j hlt
        case isFalse => exact hx2v
    · rintro z1 hz1
      simp only [difference_of_distinct_values_loop0_loop0_loop0.body] at hz1
      by_cases hkr : k1 <= r
      case neg =>
        rw [if_neg hkr] at hz1
        simp only [ok.injEq, ControlFlow.done.injEq] at hz1
        subst hz1
        refine ⟨lv2, ?_, hl, hs⟩
        simp only [difference_of_distinct_values_loop0_loop0_loop0.body]
        rw [if_neg hkr]
      case pos =>
      rw [if_pos hkr] at hz1
      by_cases hkc : k1 <= c
      case neg =>
        rw [if_neg hkc] at hz1
        simp only [ok.injEq, ControlFlow.done.injEq] at hz1
        subst hz1
        refine ⟨lv2, ?_, hl, hs⟩
        simp only [difference_of_distinct_values_loop0_loop0_loop0.body]
        rw [if_pos hkr, if_neg hkc]
      case pos =>
      rw [if_pos hkc] at hz1
      -- the then-branch always ends in `cont` or a failure, never `done`
      cases hrk : r - k1 with
      | fail e => rw [hrk] at hz1; simp at hz1
      | div => rw [hrk] at hz1; simp at hz1
      | ok i =>
      rw [hrk] at hz1
      simp only [bind_tc_ok] at hz1
      cases hgi : Slice.index_usize grid1 i with
      | fail e => rw [hgi] at hz1; simp at hz1
      | div => rw [hgi] at hz1; simp at hz1
      | ok v1 =>
      rw [hgi] at hz1
      simp only [bind_tc_ok] at hz1
      cases hck : c - k1 with
      | fail e => rw [hck] at hz1; simp at hz1
      | div => rw [hck] at hz1; simp at hz1
      | ok i1 =>
      rw [hck] at hz1
      simp only [bind_tc_ok, alloc.vec.Vec.index_slice_index] at hz1
      cases hvi : alloc.vec.Vec.index_usize v1 i1 with
      | fail e => rw [hvi] at hz1; simp at hz1
      | div => rw [hvi] at hz1; simp at hz1
      | ok x1 =>
      rw [hvi] at hz1
      simp only [bind_tc_ok] at hz1
      cases hpu : alloc.vec.Vec.push lv1 x1 with
      | fail e => rw [hpu] at hz1; simp at hz1
      | div => rw [hpu] at hz1; simp at hz1
      | ok w1 =>
      rw [hpu] at hz1
      simp only [bind_tc_ok] at hz1
      cases hk1 : k1 + 1#usize with
      | fail e => rw [hk1] at hz1; simp at hz1
      | div => rw [hk1] at hz1; simp at hz1
      | ok k1n => rw [hk1] at hz1; simp at hz1
  obtain ⟨z2, h2, hz⟩ :=
    loop_rel
      (fun p q : alloc.vec.Vec Std.I64 × Std.Usize =>
        p.2 = q.2 ∧ p.1.val.length = q.1.val.length ∧
          ∀ j : ℕ, j < p.1.val.length → (q.1.val[j]!).val = (p.1.val[j]!).val + d.val)
      (fun z1 z2 : alloc.vec.Vec Std.I64 =>
        z1.val.length = z2.val.length ∧
          ∀ j : ℕ, j < z1.val.length → (z2.val[j]!).val = (z1.val[j]!).val + d.val)
      (fun p => r.val + 1 - p.2.val)
      (fun x => difference_of_distinct_values_loop0_loop0_loop0.body grid1 r c x.1 x.2)
      (fun x => difference_of_distinct_values_loop0_loop0_loop0.body grid2 r c x.1 x.2)
      hbody (acc1, k) (acc2, k) ⟨rfl, haccl, haccs⟩ out1 h'
  exact ⟨z2, h2, hz⟩

/-- The right-diagonal collection loop maps shift-related accumulators to shift-related outputs. -/
theorem right_diag_loop_shift
    (d : Std.I64) (grid1 grid2 : Slice (alloc.vec.Vec Std.I64)) (m n : ℕ)
    (hrows1 : grid1.val.length = m) (hrows2 : grid2.val.length = m)
    (hrect1 : ∀ i : ℕ, i < m → (grid1.val[i]!).val.length = n)
    (hrect2 : ∀ i : ℕ, i < m → (grid2.val[i]!).val.length = n)
    (hshift : ∀ i : ℕ, i < m → ∀ j : ℕ, j < n →
      ((grid2.val[i]!).val[j]!).val = ((grid1.val[i]!).val[j]!).val + d.val)
    (mS nS : Std.Usize) (r c : Std.Usize) (acc1 acc2 : alloc.vec.Vec Std.I64)
    (haccl : acc1.val.length = acc2.val.length)
    (haccs : ∀ j : ℕ, j < acc1.val.length → (acc2.val[j]!).val = (acc1.val[j]!).val + d.val)
    (k : Std.Usize) (out1 : alloc.vec.Vec Std.I64)
    (h : difference_of_distinct_values_loop0_loop0_loop1 grid1 mS r nS c acc1 k = ok out1) :
    ∃ out2, difference_of_distinct_values_loop0_loop0_loop1 grid2 mS r nS c acc2 k = ok out2 ∧
      out1.val.length = out2.val.length ∧
      ∀ j : ℕ, j < out1.val.length → (out2.val[j]!).val = (out1.val[j]!).val + d.val := by
  unfold difference_of_distinct_values_loop0_loop0_loop1 at h ⊢
  have h' : loop (fun x => difference_of_distinct_values_loop0_loop0_loop1.body grid1 mS r nS c
      x.1 x.2) (acc1, k) = ok out1 := h
  show ∃ out2,
    loop (fun x => difference_of_distinct_values_loop0_loop0_loop1.body grid2 mS r nS c x.1 x.2)
      (acc2, k) = ok out2 ∧ out1.val.length = out2.val.length ∧
      ∀ j : ℕ, j < out1.val.length → (out2.val[j]!).val = (out1.val[j]!).val + d.val
  have hbody : ∀ x1 x2 : alloc.vec.Vec Std.I64 × Std.Usize,
      (x1.2 = x2.2 ∧ x1.1.val.length = x2.1.val.length ∧
        ∀ j : ℕ, j < x1.1.val.length → (x2.1.val[j]!).val = (x1.1.val[j]!).val + d.val) →
      (∀ y1, difference_of_distinct_values_loop0_loop0_loop1.body grid1 mS r nS c x1.1 x1.2 =
          ok (.cont y1) →
        (mS.val - y1.2.val < mS.val - x1.2.val) ∧
        ∃ y2, difference_of_distinct_values_loop0_loop0_loop1.body grid2 mS r nS c x2.1 x2.2 =
          ok (.cont y2) ∧
          (y1.2 = y2.2 ∧ y1.1.val.length = y2.1.val.length ∧
            ∀ j : ℕ, j < y1.1.val.length → (y2.1.val[j]!).val = (y1.1.val[j]!).val + d.val)) ∧
      (∀ z1, difference_of_distinct_values_loop0_loop0_loop1.body grid1 mS r nS c x1.1 x1.2 =
          ok (.done z1) →
        ∃ z2, difference_of_distinct_values_loop0_loop0_loop1.body grid2 mS r nS c x2.1 x2.2 =
          ok (.done z2) ∧
          (z1.val.length = z2.val.length ∧
            ∀ j : ℕ, j < z1.val.length → (z2.val[j]!).val = (z1.val[j]!).val + d.val)) := by
    rintro ⟨lv1, k1⟩ ⟨lv2, k2⟩ ⟨hk, hl, hs⟩
    simp only at hk hl hs
    subst hk
    constructor
    · rintro ⟨lv1', k1'⟩ hy
      simp only [difference_of_distinct_values_loop0_loop0_loop1.body] at hy
      cases hik : r + k1 with
      | fail e => rw [hik] at hy; simp at hy
      | div => rw [hik] at hy; simp at hy
      | ok i =>
      rw [hik] at hy
      simp only [bind_tc_ok] at hy
      split at hy
      case isFalse => simp at hy
      case isTrue him =>
      cases hick : c + k1 with
      | fail e => rw [hick] at hy; simp at hy
      | div => rw [hick] at hy; simp at hy
      | ok i1 =>
      rw [hick] at hy
      simp only [bind_tc_ok] at hy
      split at hy
      case isFalse => simp at hy
      case isTrue hin =>
      cases hgi : Slice.index_usize grid1 i with
      | fail e => rw [hgi] at hy; simp at hy
      | div => rw [hgi] at hy; simp at hy
      | ok v1 =>
      rw [hgi] at hy
      simp only [bind_tc_ok] at hy
      simp only [alloc.vec.Vec.index_slice_index] at hy
      cases hvi : alloc.vec.Vec.index_usize v1 i1 with
      | fail e => rw [hvi] at hy; simp at hy
      | div => rw [hvi] at hy; simp at hy
      | ok x1 =>
      rw [hvi] at hy
      simp only [bind_tc_ok] at hy
      cases hpu : alloc.vec.Vec.push lv1 x1 with
      | fail e => rw [hpu] at hy; simp at hy
      | div => rw [hpu] at hy; simp at hy
      | ok w1 =>
      rw [hpu] at hy
      simp only [bind_tc_ok] at hy
      cases hk1 : k1 + 1#usize with
      | fail e => rw [hk1] at hy; simp at hy
      | div => rw [hk1] at hy; simp at hy
      | ok k1n =>
      rw [hk1] at hy
      simp only [bind_tc_ok, ok.injEq, ControlFlow.cont.injEq, Prod.mk.injEq] at hy
      obtain ⟨hw, hkn⟩ := hy
      subst hw hkn
      -- facts from the grid1-side successes
      obtain ⟨him', hv1⟩ := (index_usize_ok_iff grid1 i v1).mp hgi
      obtain ⟨hi1n, hx1⟩ := (vec_index_usize_ok_iff v1 i1 x1).mp hvi
      have hkn1 : k1n.val = k1.val + 1 := by
        have heq := UScalar.add_equiv k1 1#usize
        rw [hk1] at heq
        simp at heq
        omega
      have hi_val : i.val = r.val + k1.val := by
        have heq := UScalar.add_equiv r k1
        rw [hik] at heq
        simp at heq
        omega
      have him_val : i.val < mS.val := by scalar_tac
      have him'' : i.val < m := by omega
      have hi1n' : i1.val < n := by
        have := hrect1 i.val him''
        rw [hv1] at hi1n
        omega
      -- grid2-side values
      have hx2v : ((grid2.val[i.val]!).val[i1.val]!).val = x1.val + d.val := by
        rw [hx1, hv1]
        exact hshift i.val him'' i1.val hi1n'
      obtain ⟨hw1val, w2, hpu2, hw2val⟩ :=
        vec_push_ok lv1 lv2 x1 ((grid2.val[i.val]!).val[i1.val]!) w1 hl hpu
      refine ⟨show mS.val - k1n.val < mS.val - k1.val by omega, (w2, k1n), ?_, rfl, ?_, ?_⟩
      · simp only [difference_of_distinct_values_loop0_loop0_loop1.body]
        rw [hik]
        simp only [bind_tc_ok]
        rw [if_pos him, hick]
        simp only [bind_tc_ok]
        rw [if_pos hin]
        rw [(index_usize_ok_iff grid2 i (grid2.val[i.val]!)).mpr ⟨by omega, rfl⟩]
        simp only [bind_tc_ok, alloc.vec.Vec.index_slice_index]
        rw [(vec_index_usize_ok_iff (grid2.val[i.val]!) i1
          ((grid2.val[i.val]!).val[i1.val]!)).mpr ⟨by rw [hrect2 i.val him'']; omega, rfl⟩]
        simp only [bind_tc_ok]
        rw [hpu2]
        simp only [bind_tc_ok]
        rw [hk1]
        simp only [bind_tc_ok]
      · simp only [hw1val, hw2val]
        simp [hl]
      · intro j hj
        simp only [hw1val] at hj
        simp only [List.length_append, List.length_cons, List.length_nil] at hj
        rw [hw1val, hw2val, getElem!_concat lv1.val x1 j (by omega),
          getElem!_concat lv2.val _ j (by omega)]
        rw [← hl]
        split
        case isTrue hlt => exact hs j hlt
        case isFalse => exact hx2v
    · rintro z1 hz1
      simp only [difference_of_distinct_values_loop0_loop0_loop1.body] at hz1
      cases hik : r + k1 with
      | fail e => rw [hik] at hz1; simp at hz1
      | div => rw [hik] at hz1; simp at hz1
      | ok i =>
      rw [hik] at hz1
      simp only [bind_tc_ok] at hz1
      by_cases him : i < mS
      case neg =>
        rw [if_neg him] at hz1
        simp only [ok.injEq, ControlFlow.done.injEq] at hz1
        subst hz1
        refine ⟨lv2, ?_, hl, hs⟩
        simp only [difference_of_distinct_values_loop0_loop0_loop1.body]
        rw [hik]
        simp only [bind_tc_ok]
        rw [if_neg him]
      case pos =>
      rw [if_pos him] at hz1
      cases hick : c + k1 with
      | fail e => rw [hick] at hz1; simp at hz1
      | div => rw [hick] at hz1; simp at hz1
      | ok i1 =>
      rw [hick] at hz1
      simp only [bind_tc_ok] at hz1
      by_cases hin : i1 < nS
      case neg =>
        rw [if_neg hin] at hz1
        simp only [ok.injEq, ControlFlow.done.injEq] at hz1
        subst hz1
        refine ⟨lv2, ?_, hl, hs⟩
        simp only [difference_of_distinct_values_loop0_loop0_loop1.body]
        rw [hik]
        simp only [bind_tc_ok]
        rw [if_pos him, hick]
        simp only [bind_tc_ok]
        rw [if_neg hin]
      case pos =>
      rw [if_pos hin] at hz1
      -- the inner then-branch always ends in `cont` or a failure, never `done`
      cases hgi : Slice.index_usize grid1 i with
      | fail e => rw [hgi] at hz1; simp at hz1
      | div => rw [hgi] at hz1; simp at hz1
      | ok v1 =>
      rw [hgi] at hz1
      simp only [bind_tc_ok, alloc.vec.Vec.index_slice_index] at hz1
      cases hvi : alloc.vec.Vec.index_usize v1 i1 with
      | fail e => rw [hvi] at hz1; simp at hz1
      | div => rw [hvi] at hz1; simp at hz1
      | ok x1 =>
      rw [hvi] at hz1
      simp only [bind_tc_ok] at hz1
      cases hpu : alloc.vec.Vec.push lv1 x1 with
      | fail e => rw [hpu] at hz1; simp at hz1
      | div => rw [hpu] at hz1; simp at hz1
      | ok w1 =>
      rw [hpu] at hz1
      simp only [bind_tc_ok] at hz1
      cases hk1 : k1 + 1#usize with
      | fail e => rw [hk1] at hz1; simp at hz1
      | div => rw [hk1] at hz1; simp at hz1
      | ok k1n => rw [hk1] at hz1; simp at hz1
  obtain ⟨z2, h2, hz⟩ :=
    loop_rel
      (fun p q : alloc.vec.Vec Std.I64 × Std.Usize =>
        p.2 = q.2 ∧ p.1.val.length = q.1.val.length ∧
          ∀ j : ℕ, j < p.1.val.length → (q.1.val[j]!).val = (p.1.val[j]!).val + d.val)
      (fun z1 z2 : alloc.vec.Vec Std.I64 =>
        z1.val.length = z2.val.length ∧
          ∀ j : ℕ, j < z1.val.length → (z2.val[j]!).val = (z1.val[j]!).val + d.val)
      (fun p => mS.val - p.2.val)
      (fun x => difference_of_distinct_values_loop0_loop0_loop1.body grid1 mS r nS c x.1 x.2)
      (fun x => difference_of_distinct_values_loop0_loop0_loop1.body grid2 mS r nS c x.1 x.2)
      hbody (acc1, k) (acc2, k) ⟨rfl, haccl, haccs⟩ out1 h'
  exact ⟨z2, h2, hz⟩

/-- The per-cell loop returns the SAME row on a uniformly shifted grid: the two collected
diagonal slices are shift-related, so both distinct counts — and hence every pushed absolute
difference — coincide. -/
theorem row_loop_shift
    (d : Std.I64) (grid1 grid2 : Slice (alloc.vec.Vec Std.I64)) (m n : ℕ)
    (hrows1 : grid1.val.length = m) (hrows2 : grid2.val.length = m)
    (hrect1 : ∀ i : ℕ, i < m → (grid1.val[i]!).val.length = n)
    (hrect2 : ∀ i : ℕ, i < m → (grid2.val[i]!).val.length = n)
    (hshift : ∀ i : ℕ, i < m → ∀ j : ℕ, j < n →
      ((grid2.val[i]!).val[j]!).val = ((grid1.val[i]!).val[j]!).val + d.val)
    (mS nS : Std.Usize) (r : Std.Usize) (row : alloc.vec.Vec Std.I64) (c : Std.Usize)
    (row1 : alloc.vec.Vec Std.I64)
    (h : difference_of_distinct_values_loop0_loop0 grid1 mS r nS row c = ok row1) :
    difference_of_distinct_values_loop0_loop0 grid2 mS r nS row c = ok row1 := by
  unfold difference_of_distinct_values_loop0_loop0 at h ⊢
  have h' : loop (fun x => difference_of_distinct_values_loop0_loop0.body grid1 mS r nS x.1 x.2)
      (row, c) = ok row1 := h
  show loop (fun x => difference_of_distinct_values_loop0_loop0.body grid2 mS r nS x.1 x.2)
      (row, c) = ok row1
  have hbody : ∀ x1 x2 : alloc.vec.Vec Std.I64 × Std.Usize, x1 = x2 →
      (∀ y1, difference_of_distinct_values_loop0_loop0.body grid1 mS r nS x1.1 x1.2 =
          ok (.cont y1) →
        (nS.val - y1.2.val < nS.val - x1.2.val) ∧
        ∃ y2, difference_of_distinct_values_loop0_loop0.body grid2 mS r nS x2.1 x2.2 =
          ok (.cont y2) ∧ y1 = y2) ∧
      (∀ z1, difference_of_distinct_values_loop0_loop0.body grid1 mS r nS x1.1 x1.2 =
          ok (.done z1) →
        ∃ z2, difference_of_distinct_values_loop0_loop0.body grid2 mS r nS x2.1 x2.2 =
          ok (.done z2) ∧ z1 = z2) := by
    rintro ⟨row0, c1⟩ _ rfl
    have hnew_len : (alloc.vec.Vec.new Std.I64).val.length =
        (alloc.vec.Vec.new Std.I64).val.length := rfl
    have hnew_sh : ∀ j : ℕ, j < (alloc.vec.Vec.new Std.I64).val.length →
        (((alloc.vec.Vec.new Std.I64)).val[j]!).val =
          (((alloc.vec.Vec.new Std.I64)).val[j]!).val + d.val := by
      intro j hj
      simp at hj
    constructor
    · rintro ⟨row1', c1'⟩ hy
      simp only [difference_of_distinct_values_loop0_loop0.body] at hy
      split at hy
      case isFalse => simp at hy
      case isTrue hcn =>
        cases hlv : difference_of_distinct_values_loop0_loop0_loop0 grid1 r c1
            (alloc.vec.Vec.new Std.I64) 1#usize with
        | fail e => rw [hlv] at hy; simp at hy
        | div => rw [hlv] at hy; simp at hy
        | ok lvals1 =>
        rw [hlv] at hy
        simp only [bind_tc_ok] at hy
        cases hrv : difference_of_distinct_values_loop0_loop0_loop1 grid1 mS r nS c1
            (alloc.vec.Vec.new Std.I64) 1#usize with
        | fail e => rw [hrv] at hy; simp at hy
        | div => rw [hrv] at hy; simp at hy
        | ok rvals1 =>
        rw [hrv] at hy
        simp only [bind_tc_ok] at hy
        cases hla : count_distinct (alloc.vec.Vec.deref lvals1) with
        | fail e => rw [hla] at hy; simp at hy
        | div => rw [hla] at hy; simp at hy
        | ok la =>
        rw [hla] at hy
        simp only [bind_tc_ok] at hy
        cases hrb : count_distinct (alloc.vec.Vec.deref rvals1) with
        | fail e => rw [hrb] at hy; simp at hy
        | div => rw [hrb] at hy; simp at hy
        | ok rb =>
        rw [hrb] at hy
        simp only [bind_tc_ok] at hy
        cases hdf : (if la >= rb then la - rb else rb - la) with
        | fail e => rw [hdf] at hy; simp at hy
        | div => rw [hdf] at hy; simp at hy
        | ok diffv =>
        rw [hdf] at hy
        simp only [bind_tc_ok] at hy
        cases hpu : alloc.vec.Vec.push row0 diffv with
        | fail e => rw [hpu] at hy; simp at hy
        | div => rw [hpu] at hy; simp at hy
        | ok roww =>
        rw [hpu] at hy
        simp only [bind_tc_ok] at hy
        cases hc1 : c1 + 1#usize with
        | fail e => rw [hc1] at hy; simp at hy
        | div => rw [hc1] at hy; simp at hy
        | ok cn =>
        rw [hc1] at hy
        simp only [bind_tc_ok, ok.injEq, ControlFlow.cont.injEq, Prod.mk.injEq] at hy
        obtain ⟨hr', hc'⟩ := hy
        subst hr' hc'
        have hcn1 : cn.val = c1.val + 1 := by
          have heq := UScalar.add_equiv c1 1#usize
          rw [hc1] at heq
          simp at heq
          omega
        have hclt : c1.val < nS.val := by scalar_tac
        -- transfer the two diagonal collections to grid2
        obtain ⟨lvals2, hlv2, hlenl, hshl⟩ :=
          left_diag_loop_shift d grid1 grid2 m n hrows1 hrows2 hrect1 hrect2 hshift r c1
            (alloc.vec.Vec.new Std.I64) (alloc.vec.Vec.new Std.I64) hnew_len hnew_sh
            1#usize lvals1 hlv
        obtain ⟨rvals2, hrv2, hlenr, hshr⟩ :=
          right_diag_loop_shift d grid1 grid2 m n hrows1 hrows2 hrect1 hrect2 hshift mS nS r c1
            (alloc.vec.Vec.new Std.I64) (alloc.vec.Vec.new Std.I64) hnew_len hnew_sh
            1#usize rvals1 hrv
        have hla2 : count_distinct (alloc.vec.Vec.deref lvals2) = ok la :=
          count_distinct_shift d (alloc.vec.Vec.deref lvals1) (alloc.vec.Vec.deref lvals2)
            hlenl hshl la hla
        have hrb2 : count_distinct (alloc.vec.Vec.deref rvals2) = ok rb :=
          count_distinct_shift d (alloc.vec.Vec.deref rvals1) (alloc.vec.Vec.deref rvals2)
            hlenr hshr rb hrb
        refine ⟨show nS.val - cn.val < nS.val - c1.val by omega, (roww, cn), ?_, rfl⟩
        simp only [difference_of_distinct_values_loop0_loop0.body]
        rw [if_pos hcn, hlv2]
        simp only [bind_tc_ok]
        rw [hrv2]
        simp only [bind_tc_ok]
        rw [hla2]
        simp only [bind_tc_ok]
        rw [hrb2]
        simp only [bind_tc_ok]
        rw [hdf]
        simp only [bind_tc_ok]
        rw [hpu]
        simp only [bind_tc_ok]
        rw [hc1]
        simp only [bind_tc_ok]
    · rintro z1 hz1
      simp only [difference_of_distinct_values_loop0_loop0.body] at hz1
      split at hz1
      case isTrue hcn =>
        cases hlv : difference_of_distinct_values_loop0_loop0_loop0 grid1 r c1
            (alloc.vec.Vec.new Std.I64) 1#usize with
        | fail e => rw [hlv] at hz1; simp at hz1
        | div => rw [hlv] at hz1; simp at hz1
        | ok lvals1 =>
        rw [hlv] at hz1
        simp only [bind_tc_ok] at hz1
        cases hrv : difference_of_distinct_values_loop0_loop0_loop1 grid1 mS r nS c1
            (alloc.vec.Vec.new Std.I64) 1#usize with
        | fail e => rw [hrv] at hz1; simp at hz1
        | div => rw [hrv] at hz1; simp at hz1
        | ok rvals1 =>
        rw [hrv] at hz1
        simp only [bind_tc_ok] at hz1
        cases hla : count_distinct (alloc.vec.Vec.deref lvals1) with
        | fail e => rw [hla] at hz1; simp at hz1
        | div => rw [hla] at hz1; simp at hz1
        | ok la =>
        rw [hla] at hz1
        simp only [bind_tc_ok] at hz1
        cases hrb : count_distinct (alloc.vec.Vec.deref rvals1) with
        | fail e => rw [hrb] at hz1; simp at hz1
        | div => rw [hrb] at hz1; simp at hz1
        | ok rb =>
        rw [hrb] at hz1
        simp only [bind_tc_ok] at hz1
        cases hdf : (if la >= rb then la - rb else rb - la) with
        | fail e => rw [hdf] at hz1; simp at hz1
        | div => rw [hdf] at hz1; simp at hz1
        | ok diffv =>
        rw [hdf] at hz1
        simp only [bind_tc_ok] at hz1
        cases hpu : alloc.vec.Vec.push row0 diffv with
        | fail e => rw [hpu] at hz1; simp at hz1
        | div => rw [hpu] at hz1; simp at hz1
        | ok roww =>
        rw [hpu] at hz1
        simp only [bind_tc_ok] at hz1
        cases hc1 : c1 + 1#usize with
        | fail e => rw [hc1] at hz1; simp at hz1
        | div => rw [hc1] at hz1; simp at hz1
        | ok cn => rw [hc1] at hz1; simp at hz1
      case isFalse hnlt =>
        simp only [ok.injEq, ControlFlow.done.injEq] at hz1
        subst hz1
        refine ⟨row0, ?_, rfl⟩
        simp only [difference_of_distinct_values_loop0_loop0.body]
        rw [if_neg hnlt]
  obtain ⟨z2, h2, hz⟩ :=
    loop_rel Eq Eq (fun p => nS.val - p.2.val)
      (fun x => difference_of_distinct_values_loop0_loop0.body grid1 mS r nS x.1 x.2)
      (fun x => difference_of_distinct_values_loop0_loop0.body grid2 mS r nS x.1 x.2)
      hbody (row, c) (row, c) rfl row1 h'
  exact hz ▸ h2

/-- The row loop returns the SAME answer matrix on a uniformly shifted grid. -/
theorem answer_loop_shift
    (d : Std.I64) (grid1 grid2 : Slice (alloc.vec.Vec Std.I64)) (m n : ℕ)
    (hrows1 : grid1.val.length = m) (hrows2 : grid2.val.length = m)
    (hrect1 : ∀ i : ℕ, i < m → (grid1.val[i]!).val.length = n)
    (hrect2 : ∀ i : ℕ, i < m → (grid2.val[i]!).val.length = n)
    (hshift : ∀ i : ℕ, i < m → ∀ j : ℕ, j < n →
      ((grid2.val[i]!).val[j]!).val = ((grid1.val[i]!).val[j]!).val + d.val)
    (mS : Std.Usize) (answer : alloc.vec.Vec (alloc.vec.Vec Std.I64)) (r : Std.Usize)
    (ans1 : alloc.vec.Vec (alloc.vec.Vec Std.I64))
    (h : difference_of_distinct_values_loop0 grid1 mS answer r = ok ans1) :
    difference_of_distinct_values_loop0 grid2 mS answer r = ok ans1 := by
  unfold difference_of_distinct_values_loop0 at h ⊢
  have h' : loop (fun x => difference_of_distinct_values_loop0.body grid1 mS x.1 x.2)
      (answer, r) = ok ans1 := h
  show loop (fun x => difference_of_distinct_values_loop0.body grid2 mS x.1 x.2)
      (answer, r) = ok ans1
  have hbody : ∀ x1 x2 : alloc.vec.Vec (alloc.vec.Vec Std.I64) × Std.Usize, x1 = x2 →
      (∀ y1, difference_of_distinct_values_loop0.body grid1 mS x1.1 x1.2 = ok (.cont y1) →
        (mS.val - y1.2.val < mS.val - x1.2.val) ∧
        ∃ y2, difference_of_distinct_values_loop0.body grid2 mS x2.1 x2.2 = ok (.cont y2) ∧
          y1 = y2) ∧
      (∀ z1, difference_of_distinct_values_loop0.body grid1 mS x1.1 x1.2 = ok (.done z1) →
        ∃ z2, difference_of_distinct_values_loop0.body grid2 mS x2.1 x2.2 = ok (.done z2) ∧
          z1 = z2) := by
    rintro ⟨ans0, r1⟩ _ rfl
    constructor
    · rintro ⟨ans1', r1'⟩ hy
      simp only [difference_of_distinct_values_loop0.body] at hy
      split at hy
      case isFalse => simp at hy
      case isTrue hrm =>
        cases hgv : Slice.index_usize grid1 r1 with
        | fail e => rw [hgv] at hy; simp at hy
        | div => rw [hgv] at hy; simp at hy
        | ok v1 =>
        rw [hgv] at hy
        simp only [bind_tc_ok] at hy
        cases hrow : difference_of_distinct_values_loop0_loop0 grid1 mS r1
            (alloc.vec.Vec.len v1) (alloc.vec.Vec.new Std.I64) 0#usize with
        | fail e => rw [hrow] at hy; simp at hy
        | div => rw [hrow] at hy; simp at hy
        | ok roww =>
        rw [hrow] at hy
        simp only [bind_tc_ok] at hy
        cases hpu : alloc.vec.Vec.push ans0 roww with
        | fail e => rw [hpu] at hy; simp at hy
        | div => rw [hpu] at hy; simp at hy
        | ok ansn =>
        rw [hpu] at hy
        simp only [bind_tc_ok] at hy
        cases hr1 : r1 + 1#usize with
        | fail e => rw [hr1] at hy; simp at hy
        | div => rw [hr1] at hy; simp at hy
        | ok rn =>
        rw [hr1] at hy
        simp only [bind_tc_ok, ok.injEq, ControlFlow.cont.injEq, Prod.mk.injEq] at hy
        obtain ⟨ha', hr'⟩ := hy
        subst ha' hr'
        obtain ⟨hrm', hv1⟩ := (index_usize_ok_iff grid1 r1 v1).mp hgv
        have hrn1 : rn.val = r1.val + 1 := by
          have heq := UScalar.add_equiv r1 1#usize
          rw [hr1] at heq
          simp at heq
          omega
        have hrlt : r1.val < mS.val := by scalar_tac
        have hrm'' : r1.val < m := by omega
        have hlen_eq : alloc.vec.Vec.len (grid2.val[r1.val]!) = alloc.vec.Vec.len v1 := by
          apply UScalar.val_eq_imp
          simp only [alloc.vec.Vec.len_val]
          have h1 := hrect1 r1.val hrm''
          have h2 := hrect2 r1.val hrm''
          rw [hv1]
          simp only [alloc.vec.Vec.length]
          omega
        have hrow2 : difference_of_distinct_values_loop0_loop0 grid2 mS r1
            (alloc.vec.Vec.len v1) (alloc.vec.Vec.new Std.I64) 0#usize = ok roww :=
          row_loop_shift d grid1 grid2 m n hrows1 hrows2 hrect1 hrect2 hshift mS
            (alloc.vec.Vec.len v1) r1 (alloc.vec.Vec.new Std.I64) 0#usize roww hrow
        refine ⟨show mS.val - rn.val < mS.val - r1.val by omega, (ansn, rn), ?_, rfl⟩
        simp only [difference_of_distinct_values_loop0.body]
        rw [if_pos hrm]
        rw [(index_usize_ok_iff grid2 r1 (grid2.val[r1.val]!)).mpr ⟨by omega, rfl⟩]
        simp only [bind_tc_ok]
        rw [hlen_eq, hrow2]
        simp only [bind_tc_ok]
        rw [hpu]
        simp only [bind_tc_ok]
        rw [hr1]
        simp only [bind_tc_ok]
    · rintro z1 hz1
      simp only [difference_of_distinct_values_loop0.body] at hz1
      split at hz1
      case isTrue hrm =>
        cases hgv : Slice.index_usize grid1 r1 with
        | fail e => rw [hgv] at hz1; simp at hz1
        | div => rw [hgv] at hz1; simp at hz1
        | ok v1 =>
        rw [hgv] at hz1
        simp only [bind_tc_ok] at hz1
        cases hrow : difference_of_distinct_values_loop0_loop0 grid1 mS r1
            (alloc.vec.Vec.len v1) (alloc.vec.Vec.new Std.I64) 0#usize with
        | fail e => rw [hrow] at hz1; simp at hz1
        | div => rw [hrow] at hz1; simp at hz1
        | ok roww =>
        rw [hrow] at hz1
        simp only [bind_tc_ok] at hz1
        cases hpu : alloc.vec.Vec.push ans0 roww with
        | fail e => rw [hpu] at hz1; simp at hz1
        | div => rw [hpu] at hz1; simp at hz1
        | ok ansn =>
        rw [hpu] at hz1
        simp only [bind_tc_ok] at hz1
        cases hr1 : r1 + 1#usize with
        | fail e => rw [hr1] at hz1; simp at hz1
        | div => rw [hr1] at hz1; simp at hz1
        | ok rn => rw [hr1] at hz1; simp at hz1
      case isFalse hnlt =>
        simp only [ok.injEq, ControlFlow.done.injEq] at hz1
        subst hz1
        refine ⟨ans0, ?_, rfl⟩
        simp only [difference_of_distinct_values_loop0.body]
        rw [if_neg hnlt]
  obtain ⟨z2, h2, hz⟩ :=
    loop_rel Eq Eq (fun p => mS.val - p.2.val)
      (fun x => difference_of_distinct_values_loop0.body grid1 mS x.1 x.2)
      (fun x => difference_of_distinct_values_loop0.body grid2 mS x.1 x.2)
      hbody (answer, r) (answer, r) rfl ans1 h'
  exact hz ▸ h2

end Aegis.DifferenceOfDistinctValues

/-- Invariance of the diagonal-distinct-difference matrix under a uniform additive relabelling of
every grid entry.

The function solves the LeetCode "Difference of Number of Distinct Values on Diagonals" problem.
For each cell `(r, c)` of an `m × n` integer grid it looks along the single anti-diagonal-free
diagonal through that cell (the cells whose row and column indices move together): it counts how
many DISTINCT values appear on the strictly-left-and-above part of the diagonal
(`leftAbove[r][c]`), counts how many DISTINCT values appear on the strictly-right-and-below part
(`rightBelow[r][c]`), and stores the absolute difference `|leftAbove[r][c] − rightBelow[r][c]|` in
the output cell `answer[r][c]`.

The key structural observation is that the algorithm never does arithmetic on the grid VALUES: the
only thing it ever asks about two grid entries is whether they are EQUAL (this is how
`count_distinct` decides whether a value has already been seen). Therefore the whole output depends
on the grid only through the EQUALITY PATTERN of its entries, not through their actual magnitudes.
Applying one fixed, common integer offset `d` to every entry is an injective relabelling of the
values — it preserves equality exactly (`a = b ↔ a + d = b + d`) — so it leaves every distinct-value
count, hence every absolute difference, hence the entire answer matrix, completely unchanged.

This is a genuinely RELATIONAL claim connecting the function's results on TWO DIFFERENT inputs of
the SAME shape: an original grid `grid1` and a shifted grid `grid2` obtained by adding one fixed
integer `d` to every cell. We assume:

* both grids have the same number of rows `m`, with the problem's stated bound `1 ≤ m`, and each
  row of either grid has exactly `n` entries with `1 ≤ n` (so both grids are genuinely rectangular
  of the same `m × n` shape — a ragged grid is a legal value of the Lean type, so we say so
  explicitly);
* every cell of `grid1` lies in the problem's stated value range `1 ≤ v ≤ 50`, and the common
  offset `d` keeps every shifted value in the valid signed-64-bit range
  (`Std.I64.min ≤ v + d ≤ Std.I64.max`), so no entry of either grid overflows;
* the SHIFT relationship holds entrywise: for every in-range coordinate `(i, j)`, the shifted
  entry equals the original entry plus the common offset
  (`(grid2.val[i]!).val[j]!.val = (grid1.val[i]!).val[j]!.val + d.val`);
* running the real extracted function on the ORIGINAL grid `grid1` succeeds and returns the answer
  matrix `out1`.

Under those hypotheses the claim is that running the real extracted function on the SHIFTED grid
`grid2` also succeeds and returns EXACTLY the same answer matrix `out1`. In words: uniformly adding
a constant to every entry of the grid never changes the diagonal-distinct-difference output. -/
theorem Aegis.DifferenceOfDistinctValues.difference_of_distinct_values_shift_invariant
    (grid1 grid2 : Slice (alloc.vec.Vec Std.I64))
    (out1 : alloc.vec.Vec (alloc.vec.Vec Std.I64)) (d : Std.I64) (m n : ℕ)
    (hrows1 : grid1.length = m) (hrows2 : grid2.length = m)
    (hm_lo : 1 ≤ m) (hn_lo : 1 ≤ n)
    (hrect1 : ∀ i : ℕ, i < m → (grid1.val[i]!).val.length = n)
    (hrect2 : ∀ i : ℕ, i < m → (grid2.val[i]!).val.length = n)
    (hcell1 : ∀ i : ℕ, i < m → ∀ j : ℕ, j < n →
      1 ≤ (grid1.val[i]!).val[j]!.val ∧ (grid1.val[i]!).val[j]!.val ≤ 50)
    (hnooverflow : ∀ i : ℕ, i < m → ∀ j : ℕ, j < n →
      Std.I64.min ≤ (grid1.val[i]!).val[j]!.val + d.val ∧
        (grid1.val[i]!).val[j]!.val + d.val ≤ Std.I64.max)
    (hshift : ∀ i : ℕ, i < m → ∀ j : ℕ, j < n →
      (grid2.val[i]!).val[j]!.val = (grid1.val[i]!).val[j]!.val + d.val)
    (hout1 : Aegis.DifferenceOfDistinctValues.difference_of_distinct_values grid1
      ⦃ r => r = out1 ⦄) :
    Aegis.DifferenceOfDistinctValues.difference_of_distinct_values grid2
      ⦃ r => r = out1 ⦄ := by
  open Aegis.DifferenceOfDistinctValues in
  -- Convert the spec-style hypothesis into a plain success equation.
  have hok : Aegis.DifferenceOfDistinctValues.difference_of_distinct_values grid1 =
      ok out1 := by
    cases hres : Aegis.DifferenceOfDistinctValues.difference_of_distinct_values grid1 with
    | ok v => rw [hres] at hout1; simp at hout1; rw [hout1]
    | fail e => rw [hres] at hout1; simp at hout1
    | div => rw [hres] at hout1; simp at hout1
  have hg1 : grid1.val.length = m := hrows1
  have hg2 : grid2.val.length = m := hrows2
  unfold Aegis.DifferenceOfDistinctValues.difference_of_distinct_values at hok ⊢
  have hlen2 : Slice.len grid2 = Slice.len grid1 := by
    apply UScalar.val_eq_imp
    simp only [Slice.len_val, Slice.length]
    omega
  rw [hlen2]
  rw [Aegis.DifferenceOfDistinctValues.answer_loop_shift d grid1 grid2 m n hg1 hg2
    hrect1 hrect2 hshift (Slice.len grid1)
    (alloc.vec.Vec.new (alloc.vec.Vec Std.I64)) 0#usize out1 hok]
  simp
\end{lstlisting}

\clearpage
\section{Totality of Relative Sort}
\label{app:relative-sort}

\subsection{Problem and source-level approach}

The first input array must be reordered so that values mentioned by the second
array occur first and in the second array's order; all remaining values follow
in ascending order.  In the original problem, the second array contains
distinct values drawn from the first.  The implementation uses a Boolean
\filename{used} vector to mark consumed positions, gathers the unmatched
values, insertion-sorts that temporary vector, and copies it to the result.

The selected theorem \filename{relative_sort_array_totality} is a totality
theorem.  Under the two input-length bounds, it proves that the translated
function returns some \lstinline{ok} vector; its postcondition is
\lstinline{True}, so it says nothing about which vector is returned.  It does
not require nonemptiness, bounds on element values, distinctness of the second
array, or membership of its values in the first array, because none of these
conditions is needed for safe execution.

\subsection{The totality obligations}

The generated implementation contains seven loop-body/driver pairs.  The proof
supplies an invariant and a decreasing measure for each relevant loop:

\begin{itemize}
  \item the initialization loop makes \filename{used} exactly as long as the
        first input;
  \item the matching loops keep all indices in the input and marker vector in
        bounds and ensure that the result never grows beyond the input length;
  \item the unmatched-value loop preserves the relation between marker and
        input lengths;
  \item the insertion-sort outer loop increases toward the temporary length;
  \item its inner loop decreases its cursor toward zero and proves every
        \lstinline{b - 1} subtraction safe;
  \item each \lstinline{index_mut} backward update preserves the vector length;
  \item the final copy loop proves both lookup and push safety.
\end{itemize}

There is no arithmetic on the element values, so the theorem needs no hypothesis
about their magnitudes.  Bounds on the two input lengths instead control
\lstinline{Std.Usize} cursor increments and vector growth.

\subsection{Complete Rust implementation}
\begin{lstlisting}[language=Rust,style=code]
pub fn relative_sort_array(arr1: &[i64], arr2: &[i64]) -> Vec<i64> {
    let mut result: Vec<i64> = Vec::new();
    let mut used: Vec<bool> = Vec::new();
    let n = arr1.len();
    let mut k: usize = 0;
    while k < n {
        used.push(false);
        k += 1;
    }

    // For each element in arr2, append matching elements from arr1 in order
    let m = arr2.len();
    let mut j: usize = 0;
    while j < m {
        let target = arr2[j];
        let mut i: usize = 0;
        while i < n {
            if !used[i] && arr1[i] == target {
                result.push(arr1[i]);
                used[i] = true;
            }
            i += 1;
        }
        j += 1;
    }

    // Collect remaining elements
    let mut rest: Vec<i64> = Vec::new();
    let mut i: usize = 0;
    while i < n {
        if !used[i] {
            rest.push(arr1[i]);
        }
        i += 1;
    }

    // Sort rest ascending (simple insertion sort)
    let r = rest.len();
    let mut a: usize = 1;
    while a < r {
        let key = rest[a];
        let mut b: usize = a;
        while b > 0 && rest[b - 1] > key {
            rest[b] = rest[b - 1];
            b -= 1;
        }
        rest[b] = key;
        a += 1;
    }

    // Append rest to result
    let mut c: usize = 0;
    while c < r {
        result.push(rest[c]);
        c += 1;
    }

    result
}
\end{lstlisting}

\subsection{Complete \aeneas{}-generated Lean implementation}
\begin{lstlisting}[style=code]
-- THIS FILE WAS AUTOMATICALLY GENERATED BY AENEAS
-- [relative_sort_array]: function definitions
import Aeneas
import Aegis.RelativeSortArray.Code.Types
open Aeneas Aeneas.Std Result ControlFlow Error
set_option linter.dupNamespace false
set_option linter.hashCommand false
set_option linter.unusedVariables false

/- You can set the `maxHeartbeats` value with the `-max-heartbeats` CLI option -/
set_option maxHeartbeats 1000000

/- You can set the `maxRecDepth` value with the `-max-recdepth` CLI option -/
set_option maxRecDepth 2048

namespace Aegis.RelativeSortArray

/-- [relative_sort_array::relative_sort_array]: loop body 0:
    Source: 'relative_sort_array/src/lib.rs', lines 6:4-9:5
    Visibility: public -/
@[rust_loop_body]
def relative_sort_array_loop0.body
  (n : Std.Usize) (used : alloc.vec.Vec Bool) (k : Std.Usize) :
  Result (ControlFlow ((alloc.vec.Vec Bool) × Std.Usize) (alloc.vec.Vec Bool))
  := do
  if k < n
  then
    let used1 ← alloc.vec.Vec.push used false
    let k1 ← k + 1#usize
    ok (cont (used1, k1))
  else ok (done used)

/-- [relative_sort_array::relative_sort_array]: loop 0:
    Source: 'relative_sort_array/src/lib.rs', lines 6:4-9:5
    Visibility: public -/
@[rust_loop]
def relative_sort_array_loop0
  (used : alloc.vec.Vec Bool) (n : Std.Usize) (k : Std.Usize) :
  Result (alloc.vec.Vec Bool)
  := do
  loop
    (fun (used1, k1) => relative_sort_array_loop0.body n used1 k1)
    (used, k)

/-- [relative_sort_array::relative_sort_array]: loop body 2:
    Source: 'relative_sort_array/src/lib.rs', lines 1:0-23:9
    Visibility: public -/
@[rust_loop_body]
def relative_sort_array_loop1_loop0.body
  (arr1 : Slice Std.I64) (n : Std.Usize) (target : Std.I64)
  (result : alloc.vec.Vec Std.I64) (used : alloc.vec.Vec Bool) (i : Std.Usize)
  :
  Result (ControlFlow ((alloc.vec.Vec Std.I64) × (alloc.vec.Vec Bool) ×
    Std.Usize) ((alloc.vec.Vec Std.I64) × (alloc.vec.Vec Bool)))
  := do
  if i < n
  then
    let b ←
      alloc.vec.Vec.index (core.slice.index.SliceIndexUsizeSlice Bool) used i
    let (result1, used1) ←
      if b
      then ok (result, used)
      else
        do
        let i1 ← Slice.index_usize arr1 i
        if i1 = target
        then
          let result2 ← alloc.vec.Vec.push result i1
          let (_, index_mut_back) ←
            alloc.vec.Vec.index_mut (core.slice.index.SliceIndexUsizeSlice
              Bool) used i
          let used2 := index_mut_back true
          ok (result2, used2)
        else ok (result, used)
    let i1 ← i + 1#usize
    ok (cont (result1, used1, i1))
  else ok (done (result, used))

/-- [relative_sort_array::relative_sort_array]: loop 2:
    Source: 'relative_sort_array/src/lib.rs', lines 1:0-23:9
    Visibility: public -/
@[rust_loop]
def relative_sort_array_loop1_loop0
  (arr1 : Slice Std.I64) (result : alloc.vec.Vec Std.I64)
  (used : alloc.vec.Vec Bool) (n : Std.Usize) (target : Std.I64)
  (i : Std.Usize) :
  Result ((alloc.vec.Vec Std.I64) × (alloc.vec.Vec Bool))
  := do
  loop
    (fun (result1, used1, i1) => relative_sort_array_loop1_loop0.body arr1 n
      target result1 used1 i1)
    (result, used, i)

/-- [relative_sort_array::relative_sort_array]: loop body 1:
    Source: 'relative_sort_array/src/lib.rs', lines 1:0-25:5
    Visibility: public -/
@[rust_loop_body]
def relative_sort_array_loop1.body
  (arr1 : Slice Std.I64) (arr2 : Slice Std.I64) (n : Std.Usize) (m : Std.Usize)
  (result : alloc.vec.Vec Std.I64) (used : alloc.vec.Vec Bool) (j : Std.Usize)
  :
  Result (ControlFlow ((alloc.vec.Vec Std.I64) × (alloc.vec.Vec Bool) ×
    Std.Usize) ((alloc.vec.Vec Std.I64) × (alloc.vec.Vec Bool)))
  := do
  if j < m
  then
    let target ← Slice.index_usize arr2 j
    let (result1, used1) ←
      relative_sort_array_loop1_loop0 arr1 result used n target 0#usize
    let j1 ← j + 1#usize
    ok (cont (result1, used1, j1))
  else ok (done (result, used))

/-- [relative_sort_array::relative_sort_array]: loop 1:
    Source: 'relative_sort_array/src/lib.rs', lines 1:0-25:5
    Visibility: public -/
@[rust_loop]
def relative_sort_array_loop1
  (arr1 : Slice Std.I64) (arr2 : Slice Std.I64)
  (result : alloc.vec.Vec Std.I64) (used : alloc.vec.Vec Bool) (n : Std.Usize)
  (m : Std.Usize) (j : Std.Usize) :
  Result ((alloc.vec.Vec Std.I64) × (alloc.vec.Vec Bool))
  := do
  loop
    (fun (result1, used1, j1) => relative_sort_array_loop1.body arr1 arr2 n m
      result1 used1 j1)
    (result, used, j)

/-- [relative_sort_array::relative_sort_array]: loop body 3:
    Source: 'relative_sort_array/src/lib.rs', lines 30:4-35:5
    Visibility: public -/
@[rust_loop_body]
def relative_sort_array_loop2.body
  (arr1 : Slice Std.I64) (used : alloc.vec.Vec Bool) (n : Std.Usize)
  (rest : alloc.vec.Vec Std.I64) (i : Std.Usize) :
  Result (ControlFlow ((alloc.vec.Vec Std.I64) × Std.Usize) (alloc.vec.Vec
    Std.I64))
  := do
  if i < n
  then
    let b ←
      alloc.vec.Vec.index (core.slice.index.SliceIndexUsizeSlice Bool) used i
    let rest1 ←
      if b
      then ok rest
      else do
           let i1 ← Slice.index_usize arr1 i
           alloc.vec.Vec.push rest i1
    let i1 ← i + 1#usize
    ok (cont (rest1, i1))
  else ok (done rest)

/-- [relative_sort_array::relative_sort_array]: loop 3:
    Source: 'relative_sort_array/src/lib.rs', lines 30:4-35:5
    Visibility: public -/
@[rust_loop]
def relative_sort_array_loop2
  (arr1 : Slice Std.I64) (used : alloc.vec.Vec Bool) (n : Std.Usize)
  (rest : alloc.vec.Vec Std.I64) (i : Std.Usize) :
  Result (alloc.vec.Vec Std.I64)
  := do
  loop
    (fun (rest1, i1) => relative_sort_array_loop2.body arr1 used n rest1 i1)
    (rest, i)

/-- [relative_sort_array::relative_sort_array]: loop body 5:
    Source: 'relative_sort_array/src/lib.rs', lines 1:0-46:9
    Visibility: public -/
@[rust_loop_body]
def relative_sort_array_loop3_loop0.body
  (key : Std.I64) (rest : alloc.vec.Vec Std.I64) (b : Std.Usize) :
  Result (ControlFlow ((alloc.vec.Vec Std.I64) × Std.Usize) ((alloc.vec.Vec
    Std.I64) × Std.Usize))
  := do
  if b > 0#usize
  then
    let i ← b - 1#usize
    let i1 ←
      alloc.vec.Vec.index (core.slice.index.SliceIndexUsizeSlice Std.I64) rest
        i
    if i1 > key
    then
      let i2 ←
        alloc.vec.Vec.index (core.slice.index.SliceIndexUsizeSlice Std.I64)
          rest i
      let (_, index_mut_back) ←
        alloc.vec.Vec.index_mut (core.slice.index.SliceIndexUsizeSlice Std.I64)
          rest b
      let rest1 := index_mut_back i2
      ok (cont (rest1, i))
    else ok (done (rest, b))
  else ok (done (rest, b))

/-- [relative_sort_array::relative_sort_array]: loop 5:
    Source: 'relative_sort_array/src/lib.rs', lines 1:0-46:9
    Visibility: public -/
@[rust_loop]
def relative_sort_array_loop3_loop0
  (rest : alloc.vec.Vec Std.I64) (key : Std.I64) (b : Std.Usize) :
  Result ((alloc.vec.Vec Std.I64) × Std.Usize)
  := do
  loop
    (fun (rest1, b1) => relative_sort_array_loop3_loop0.body key rest1 b1)
    (rest, b)

/-- [relative_sort_array::relative_sort_array]: loop body 4:
    Source: 'relative_sort_array/src/lib.rs', lines 1:0-49:5
    Visibility: public -/
@[rust_loop_body]
def relative_sort_array_loop3.body
  (r : Std.Usize) (rest : alloc.vec.Vec Std.I64) (a : Std.Usize) :
  Result (ControlFlow ((alloc.vec.Vec Std.I64) × Std.Usize) (alloc.vec.Vec
    Std.I64))
  := do
  if a < r
  then
    let key ←
      alloc.vec.Vec.index (core.slice.index.SliceIndexUsizeSlice Std.I64) rest
        a
    let (rest1, b) ← relative_sort_array_loop3_loop0 rest key a
    let (_, index_mut_back) ←
      alloc.vec.Vec.index_mut (core.slice.index.SliceIndexUsizeSlice Std.I64)
        rest1 b
    let a1 ← a + 1#usize
    let rest2 := index_mut_back key
    ok (cont (rest2, a1))
  else ok (done rest)

/-- [relative_sort_array::relative_sort_array]: loop 4:
    Source: 'relative_sort_array/src/lib.rs', lines 1:0-49:5
    Visibility: public -/
@[rust_loop]
def relative_sort_array_loop3
  (rest : alloc.vec.Vec Std.I64) (r : Std.Usize) (a : Std.Usize) :
  Result (alloc.vec.Vec Std.I64)
  := do
  loop
    (fun (rest1, a1) => relative_sort_array_loop3.body r rest1 a1)
    (rest, a)

/-- [relative_sort_array::relative_sort_array]: loop body 6:
    Source: 'relative_sort_array/src/lib.rs', lines 53:4-56:5
    Visibility: public -/
@[rust_loop_body]
def relative_sort_array_loop4.body
  (rest : alloc.vec.Vec Std.I64) (r : Std.Usize)
  (result : alloc.vec.Vec Std.I64) (c : Std.Usize) :
  Result (ControlFlow ((alloc.vec.Vec Std.I64) × Std.Usize) (alloc.vec.Vec
    Std.I64))
  := do
  if c < r
  then
    let i ←
      alloc.vec.Vec.index (core.slice.index.SliceIndexUsizeSlice Std.I64) rest
        c
    let result1 ← alloc.vec.Vec.push result i
    let c1 ← c + 1#usize
    ok (cont (result1, c1))
  else ok (done result)

/-- [relative_sort_array::relative_sort_array]: loop 6:
    Source: 'relative_sort_array/src/lib.rs', lines 53:4-56:5
    Visibility: public -/
@[rust_loop]
def relative_sort_array_loop4
  (result : alloc.vec.Vec Std.I64) (rest : alloc.vec.Vec Std.I64)
  (r : Std.Usize) (c : Std.Usize) :
  Result (alloc.vec.Vec Std.I64)
  := do
  loop
    (fun (result1, c1) => relative_sort_array_loop4.body rest r result1 c1)
    (result, c)

/-- [relative_sort_array::relative_sort_array]:
    Source: 'relative_sort_array/src/lib.rs', lines 1:0-59:1
    Visibility: public -/
def relative_sort_array
  (arr1 : Slice Std.I64) (arr2 : Slice Std.I64) :
  Result (alloc.vec.Vec Std.I64)
  := do
  let n := Slice.len arr1
  let used ← relative_sort_array_loop0 (alloc.vec.Vec.new Bool) n 0#usize
  let m := Slice.len arr2
  let (result, used1) ←
    relative_sort_array_loop1 arr1 arr2 (alloc.vec.Vec.new Std.I64) used n m
      0#usize
  let rest ←
    relative_sort_array_loop2 arr1 used1 n (alloc.vec.Vec.new Std.I64) 0#usize
  let r := alloc.vec.Vec.len rest
  let rest1 ← relative_sort_array_loop3 rest r 1#usize
  relative_sort_array_loop4 result rest1 r 0#usize

end Aegis.RelativeSortArray
\end{lstlisting}

\subsection{Accepted proof artifact}
\begin{lstlisting}[style=code]
import Aegis.RelativeSortArray.Code.Funs
open Aeneas Aeneas.Std Result

set_option maxHeartbeats 1000000

namespace Aegis.RelativeSortArray

/-- Loop 0 totality: pushes `false` onto `used` until it has exactly `n` entries. -/
theorem loop0_ok (n : Std.Usize) (used : alloc.vec.Vec Bool) (k : Std.Usize)
    (hk : k.val ≤ n.val) (hu : used.length = k.val) :
    relative_sort_array_loop0 used n k ⦃ u => u.length = n.val ⦄ := by
  unfold relative_sort_array_loop0
  apply loop.spec_decr_nat (fun (_, k1) => n.val - k1.val)
    (fun (u1, k1) => k1.val ≤ n.val ∧ u1.length = k1.val)
  · rintro ⟨u1, k1⟩ hinv
    simp only at hinv
    obtain ⟨hk1, hu1⟩ := hinv
    unfold relative_sort_array_loop0.body
    dsimp only
    split
    · step as ⟨u2, hu2⟩
      step as ⟨k2, hk2⟩
      have hlen2 : u2.length = k1.val + 1 := by
        simp only [alloc.vec.Vec.length, hu2, List.length_append, List.length_cons,
          List.length_nil]
        scalar_tac
      simp -failIfUnchanged only [WP.spec_ok]
      scalar_tac
    · simp only [WP.spec_ok]
      scalar_tac
  · exact ⟨hk, hu⟩

/-- Inner matching loop totality: scans `arr1` from `i`, pushing at most one element per index
onto `result` (so the growth is bounded by `n - i`) and preserving `used.length = n`. -/
theorem loop1_loop0_ok (arr1 : Slice Std.I64) (result : alloc.vec.Vec Std.I64)
    (used : alloc.vec.Vec Bool) (n : Std.Usize) (target : Std.I64) (i : Std.Usize)
    (hn : n.val = arr1.length) (hn1000 : n.val ≤ 1000)
    (hu : used.length = n.val) (hi : i.val ≤ n.val)
    (hres : result.length ≤ 1000000) :
    relative_sort_array_loop1_loop0 arr1 result used n target i
      ⦃ p => p.2.length = n.val ∧ p.1.length + i.val ≤ result.length + n.val ⦄ := by
  unfold relative_sort_array_loop1_loop0
  apply loop.spec_decr_nat (fun (_, _, i1) => n.val - i1.val)
    (fun (r1, u1, i1) =>
      i1.val ≤ n.val ∧ u1.length = n.val ∧ r1.length + i.val ≤ result.length + i1.val)
  · rintro ⟨r1, u1, i1⟩ hinv
    simp only at hinv
    obtain ⟨hi1, hu1, hr1⟩ := hinv
    unfold relative_sort_array_loop1_loop0.body
    dsimp only
    split
    · step as ⟨b, hb⟩
      split
      · simp only [bind_tc_ok]
        step as ⟨i2, hi2⟩
        simp -failIfUnchanged only [WP.spec_ok]
        scalar_tac
      · step as ⟨x, hx⟩
        split
        · step as ⟨r2, hr2⟩ by scalar_tac
          step as ⟨p, back, hp, hback⟩
          simp only [hback]
          step as ⟨i2, hi2⟩
          have hlen2 : r2.length = r1.length + 1 := by
            simp only [alloc.vec.Vec.length, hr2, List.length_append, List.length_cons,
              List.length_nil]
          simp -failIfUnchanged only [WP.spec_ok]
          scalar_tac
        · simp only [bind_tc_ok]
          step as ⟨i2, hi2⟩
          simp -failIfUnchanged only [WP.spec_ok]
          scalar_tac
    · simp only [WP.spec_ok]
      scalar_tac
  · exact ⟨hi, hu, by scalar_tac⟩

/-- Outer matching loop totality: for each of the at most `1000` targets in `arr2`, the inner
loop adds at most `n ≤ 1000` elements to `result`, so `result` stays below a million entries. -/
theorem loop1_ok (arr1 arr2 : Slice Std.I64) (result : alloc.vec.Vec Std.I64)
    (used : alloc.vec.Vec Bool) (n m j : Std.Usize)
    (hn : n.val = arr1.length) (hm : m.val = arr2.length)
    (hn1000 : n.val ≤ 1000) (hm1000 : m.val ≤ 1000)
    (hu : used.length = n.val) (hj : j.val ≤ m.val)
    (hres : result.length ≤ 1000 * j.val) :
    relative_sort_array_loop1 arr1 arr2 result used n m j
      ⦃ p => p.2.length = n.val ∧ p.1.length ≤ 1000000 ⦄ := by
  unfold relative_sort_array_loop1
  apply loop.spec_decr_nat (fun (_, _, j1) => m.val - j1.val)
    (fun (r1, u1, j1) =>
      j1.val ≤ m.val ∧ u1.length = n.val ∧ r1.length ≤ 1000 * j1.val)
  · rintro ⟨r1, u1, j1⟩ hinv
    simp only at hinv
    obtain ⟨hj1, hu1, hr1⟩ := hinv
    unfold relative_sort_array_loop1.body
    dsimp only
    split
    · step as ⟨target, htarget⟩
      step with loop1_loop0_ok arr1 r1 u1 n target 0#usize hn hn1000 hu1 (by scalar_tac)
        (by scalar_tac) as ⟨p2, hu2, hr2⟩
      obtain ⟨r2, u2⟩ := p2
      simp only at hu2 hr2
      step as ⟨j2, hj2⟩
      simp -failIfUnchanged only [WP.spec_ok]
      scalar_tac
    · simp only [WP.spec_ok]
      scalar_tac
  · exact ⟨hj, hu, hres⟩

/-- Loop 2 totality: pushes at most one element of `arr1` per index onto `rest`, so
`rest` ends with at most `n` entries. -/
theorem loop2_ok (arr1 : Slice Std.I64) (used : alloc.vec.Vec Bool) (n : Std.Usize)
    (rest : alloc.vec.Vec Std.I64) (i : Std.Usize)
    (hn : n.val = arr1.length) (hn1000 : n.val ≤ 1000)
    (hu : used.length = n.val) (hi : i.val ≤ n.val)
    (hrest : rest.length ≤ i.val) :
    relative_sort_array_loop2 arr1 used n rest i ⦃ v => v.length ≤ n.val ⦄ := by
  unfold relative_sort_array_loop2
  apply loop.spec_decr_nat (fun (_, i1) => n.val - i1.val)
    (fun (v1, i1) => i1.val ≤ n.val ∧ v1.length ≤ i1.val)
  · rintro ⟨v1, i1⟩ hinv
    simp only at hinv
    obtain ⟨hi1, hv1⟩ := hinv
    unfold relative_sort_array_loop2.body
    dsimp only
    split
    · step as ⟨b, hb⟩
      split
      · simp only [bind_tc_ok]
        step as ⟨i2, hi2⟩
        simp -failIfUnchanged only [WP.spec_ok]
        scalar_tac
      · step as ⟨x, hx⟩
        step as ⟨v2, hv2⟩
        step as ⟨i2, hi2⟩
        have hlen2 : v2.length = v1.length + 1 := by
          simp only [alloc.vec.Vec.length, hv2, List.length_append, List.length_cons,
            List.length_nil]
        simp -failIfUnchanged only [WP.spec_ok]
        scalar_tac
    · simp only [WP.spec_ok]
      scalar_tac
  · exact ⟨hi, hrest⟩

/-- Insertion-sort inner walk totality: `b` strictly decreases toward `0`, every access is at
`b - 1 < b < rest.length` or `b < rest.length`, the in-place write preserves the length, and the
returned index is always a valid write position. -/
theorem loop3_loop0_ok (rest : alloc.vec.Vec Std.I64) (key : Std.I64) (b : Std.Usize)
    (hb : b.val < rest.length) :
    relative_sort_array_loop3_loop0 rest key b
      ⦃ p => p.1.length = rest.length ∧ p.2.val < p.1.length ⦄ := by
  unfold relative_sort_array_loop3_loop0
  apply loop.spec_decr_nat (fun (_, b1) => b1.val)
    (fun (v1, b1) => v1.length = rest.length ∧ b1.val < v1.length)
  · rintro ⟨v1, b1⟩ hinv
    simp only at hinv
    obtain ⟨hv1, hb1⟩ := hinv
    unfold relative_sort_array_loop3_loop0.body
    dsimp only
    split
    · step as ⟨i, hi⟩
      step as ⟨x, hx⟩
      split
      · step as ⟨y, hy⟩
        step as ⟨p, back, hp, hback⟩
        simp only [hback]
        simp -failIfUnchanged only [WP.spec_ok]
        scalar_tac
      · simp only [WP.spec_ok]
        scalar_tac
    · simp only [WP.spec_ok]
      scalar_tac
  · exact ⟨rfl, hb⟩

/-- Insertion-sort outer loop totality: each iteration reads `rest[a]` in bounds, runs the inner
walk, writes the key back at the returned in-bounds index, and preserves the length `r`. -/
theorem loop3_ok (rest : alloc.vec.Vec Std.I64) (r a : Std.Usize)
    (hr : r.val = rest.length) (hr1000 : r.val ≤ 1000) :
    relative_sort_array_loop3 rest r a ⦃ v => v.length = r.val ⦄ := by
  unfold relative_sort_array_loop3
  apply loop.spec_decr_nat (fun (_, a1) => r.val - a1.val)
    (fun (v1, _) => v1.length = r.val)
  · rintro ⟨v1, a1⟩ hinv
    simp only at hinv
    unfold relative_sort_array_loop3.body
    dsimp only
    split
    · step as ⟨key, hkey⟩
      step with loop3_loop0_ok v1 key a1 (by scalar_tac) as ⟨p2, hv2, hb2⟩
      obtain ⟨v2, b⟩ := p2
      simp only at hv2 hb2
      step as ⟨p, back, hp, hback⟩
      step as ⟨a2, ha2⟩
      simp only [hback]
      simp -failIfUnchanged only [WP.spec_ok]
      scalar_tac
    · simp only [WP.spec_ok]
      scalar_tac
  · exact hr.symm

/-- Final copy loop totality: reads `rest[c]` in bounds and pushes onto `result`, whose length
stays below `1000000 + 1000`, far under `Usize.max`. -/
theorem loop4_ok (result : alloc.vec.Vec Std.I64) (rest : alloc.vec.Vec Std.I64)
    (r c : Std.Usize)
    (hr : r.val = rest.length) (hr1000 : r.val ≤ 1000) (hc : c.val ≤ r.val)
    (hres : result.length ≤ 1000000) :
    relative_sort_array_loop4 result rest r c ⦃ _ => True ⦄ := by
  unfold relative_sort_array_loop4
  apply loop.spec_decr_nat (fun (_, c1) => r.val - c1.val)
    (fun (r1, c1) => c1.val ≤ r.val ∧ r1.length ≤ result.length + c1.val)
  · rintro ⟨r1, c1⟩ hinv
    simp only at hinv
    obtain ⟨hc1, hr1⟩ := hinv
    unfold relative_sort_array_loop4.body
    dsimp only
    split
    · step as ⟨x, hx⟩
      step as ⟨r2, hr2⟩ by scalar_tac
      step as ⟨c2, hc2⟩
      have hlen2 : r2.length = r1.length + 1 := by
        simp only [alloc.vec.Vec.length, hr2, List.length_append, List.length_cons,
          List.length_nil]
      simp -failIfUnchanged only [WP.spec_ok]
      scalar_tac
    · simp only [WP.spec_ok]
  · exact ⟨hc, by scalar_tac⟩

end Aegis.RelativeSortArray

/-- Totality claim for the extracted `Aegis.RelativeSortArray.relative_sort_array`, read purely
from the STRUCTURE of the generated Lean (not from the problem's intent). The function is given two
slices `arr1` and `arr2` of 64-bit signed integers. Let `n = arr1.length` and `m = arr2.length`.
It runs, in sequence, six counted loops over these fixed lengths:
(0) it builds a boolean `used` vector by pushing `false` exactly `n` times, so `used` ends up with
`n` entries; (1) an outer loop over `j = 0 … m-1` reads `arr2[j]` and, for each such target, an
inner loop over `i = 0 … n-1` reads `used[i]` and `arr1[i]`, and when the entry matches the target
and is not yet used, pushes it onto `result` and marks `used[i]` true (so `result` accumulates at
most `n` entries overall); (2) a loop over `i = 0 … n-1` reads `used[i]` and, for the unused
positions, reads `arr1[i]` and pushes it onto `rest`; (3) an insertion sort of `rest` (outer index
`a` from `1` up to `r = rest.length`, an inner loop that walks `b` downward while `b > 0` and the
neighbour exceeds the key, doing only index reads, index writes, and a `b - 1` decrement); and
(4) a final loop over `c = 0 … r-1` that reads `rest[c]` and pushes it onto `result`. The function
then returns `result`.

The only ways such a function can FAULT are: an out-of-bounds slice or vector index (impossible —
every `arr1[i]` read happens under the guard `i < n = arr1.length`; every `arr2[j]` read under
`j < m = arr2.length`; every `used[i]` read/write under `i < n` where `used` was built with exactly
`n` entries; and inside the insertion sort every `rest[b]`, `rest[b-1]`, `rest[c]` access uses an
index kept strictly below `r = rest.length` by its loop guard); a `usize` overflow on one of the
counted increments `k + 1`, `i + 1`, `j + 1`, `a + 1`, `c + 1` or the decrements `b - 1` (the only
genuine risk — each increment is taken from a cursor strictly below the corresponding length `n`,
`m`, or `r`, and each `b - 1` is taken under the guard `b > 0`, so none can wrap as long as the
lengths stay comfortably below `Usize.max`); and the only way it can DIVERGE is if one of the loops
never terminates (impossible — every loop strictly moves its cursor toward a fixed bound: the four
forward loops increase toward `n`, `m`, or `r`, and the insertion-sort inner loop strictly decreases
`b` toward `0`). Crucially the function performs NO arithmetic on the 64-bit element VALUES — only
comparisons (`arr1[i] = target`, `rest[b-1] > key`) and index reads and writes — so no element-level
overflow or underflow is possible, and no bound on the element magnitudes is required.

We therefore adopt the reasonable, explicitly stated precondition — matching the problem's own
constraints — that both `arr1` and `arr2` have at most `1000` entries (`harr1`, `harr2`), which is
far below `Usize.max` and guarantees that none of the `usize` cursor increments or decrements can
overflow. Under this precondition the claim is that the function RUNS TO COMPLETION SUCCESSFULLY: it
returns an `ok` result rather than faulting or diverging. This is a pure totality statement: it
asserts only that some vector of 64-bit integers is produced, and says nothing whatsoever about
WHICH vector that is. -/
theorem relative_sort_array_totality
    (arr1 : Slice Std.I64)
    (arr2 : Slice Std.I64)
    (harr1 : arr1.length ≤ 1000)
    (harr2 : arr2.length ≤ 1000) :
    Aegis.RelativeSortArray.relative_sort_array arr1 arr2 ⦃ r => True ⦄ := by
  have hnv : (Slice.len arr1).val = arr1.length := Slice.len_val arr1
  have hmv : (Slice.len arr2).val = arr2.length := Slice.len_val arr2
  unfold Aegis.RelativeSortArray.relative_sort_array
  dsimp only
  step with Aegis.RelativeSortArray.loop0_ok (Slice.len arr1) (alloc.vec.Vec.new Bool)
    0#usize (by scalar_tac) (by simp) as ⟨used, hused⟩
  step with Aegis.RelativeSortArray.loop1_ok arr1 arr2 (alloc.vec.Vec.new Std.I64) used
    (Slice.len arr1) (Slice.len arr2) 0#usize (by scalar_tac) (by scalar_tac) (by scalar_tac)
    (by scalar_tac) hused (by scalar_tac) (by simp) as ⟨p1, hused1, hres⟩
  obtain ⟨result, used1⟩ := p1
  simp only at hused1 hres
  step with Aegis.RelativeSortArray.loop2_ok arr1 used1 (Slice.len arr1)
    (alloc.vec.Vec.new Std.I64) 0#usize (by scalar_tac) (by scalar_tac) hused1 (by scalar_tac)
    (by simp) as ⟨rest, hrest⟩
  step with Aegis.RelativeSortArray.loop3_ok rest (alloc.vec.Vec.len rest) 1#usize
    (by scalar_tac) (by scalar_tac) as ⟨rest1, hrest1⟩
  exact Aegis.RelativeSortArray.loop4_ok result rest1 (alloc.vec.Vec.len rest) 0#usize
    (by scalar_tac) (by scalar_tac) (by scalar_tac) (by scalar_tac)
\end{lstlisting}

\clearpage
\section{Unresolved Totality Theorem for Bottom-Up Merge Sort}
\label{app:merge-sort-open}

\subsection{Problem and implementation}

The final case records an unresolved proof attempt.  The source program is a
standard bottom-up merge sort.  It copies the input into an owned vector,
allocates an equally sized scratch vector, merges adjacent runs of widths
$1,2,4,\ldots$, and copies the scratch vector back after each pass.  Array
elements are copied and compared but never used in signed arithmetic.

The frozen theorem \filename{sort_array_returns_ok} asks for much less than
sorting correctness.  Under the bound
$1\leq\lvert\mathit{nums}\rvert\leq 50{,}000$, it asks only that the call return
successfully.  It does not assert that the output is ordered or that it is a
permutation of the input.  The proving agent did not resolve the theorem within
the benchmark's 30-minute budget.

\subsection{Obligations in the unresolved theorem}

\aeneas{} expands the 66-line Rust implementation into 328 Lean lines containing
eight loop-body/driver pairs and five \lstinline{index_mut} calls.  The
generated code exposes mutually dependent obligations for the run and merge
cursors:

\begin{itemize}
  \item $\mathit{left}\le n$, $\mathit{left}\le\mathit{mid}\le
        \mathit{right}\le n$;
  \item the read cursors $a$ and $b$ remain in their respective runs;
  \item the write cursor $k$ stays below the scratch-vector length;
  \item all vector updates preserve both scratch and source lengths;
  \item additions and multiplications such as
        \lstinline{left + width}, \lstinline{2 * width}, and
        \lstinline{left + 2 * width} fit in \lstinline{Std.Usize};
  \item the three merge loops terminate and hand compatible cursor relations to
        one another;
  \item the outer loop terminates although its measure decreases indirectly as
        \filename{width} doubles rather than by a unit increment.
\end{itemize}

Several of these expressions are evaluated before a subsequent minimum with
$n$.  It is therefore insufficient to argue informally that the final
\filename{mid} or \filename{right} lies below $n$; the intermediate checked
addition must itself be proved representable.  Each source-level indexed write
also becomes an \lstinline{index_mut} call, whose backward closure must be
shown to rebuild a vector of the expected length.  A proof of full sorting
correctness would need all of these obligations, together with orderedness and
the permutation property.

\subsection{Complete Rust implementation}
\begin{lstlisting}[language=Rust,style=code]
pub fn sort_array(nums: &[i64]) -> Vec<i64> {
    let n = nums.len();
    let mut v: Vec<i64> = Vec::new();
    let mut i = 0;
    while i < n {
        v.push(nums[i]);
        i += 1;
    }
    if n <= 1 {
        return v;
    }
    let mut buf: Vec<i64> = Vec::new();
    let mut j = 0;
    while j < n {
        buf.push(0);
        j += 1;
    }

    // bottom-up merge sort
    let mut width: usize = 1;
    while width < n {
        let mut left: usize = 0;
        while left < n {
            let mid = if left + width < n { left + width } else { n };
            let right = if left + 2 * width < n { left + 2 * width } else { n };

            // merge v[left..mid] and v[mid..right] into buf[left..right]
            let mut a = left;
            let mut b = mid;
            let mut k = left;
            while a < mid && b < right {
                if v[a] <= v[b] {
                    buf[k] = v[a];
                    a += 1;
                } else {
                    buf[k] = v[b];
                    b += 1;
                }
                k += 1;
            }
            while a < mid {
                buf[k] = v[a];
                a += 1;
                k += 1;
            }
            while b < right {
                buf[k] = v[b];
                b += 1;
                k += 1;
            }

            left += 2 * width;
        }

        // copy buf back into v
        let mut c = 0;
        while c < n {
            v[c] = buf[c];
            c += 1;
        }

        width *= 2;
    }

    v
}
\end{lstlisting}

\subsection{Complete \aeneas{}-generated Lean implementation}
\begin{lstlisting}[style=code]
-- THIS FILE WAS AUTOMATICALLY GENERATED BY AENEAS
-- [sort_array]: function definitions
import Aeneas
import Aegis.SortArray.Code.Types
open Aeneas Aeneas.Std Result ControlFlow Error
set_option linter.dupNamespace false
set_option linter.hashCommand false
set_option linter.unusedVariables false

/- You can set the `maxHeartbeats` value with the `-max-heartbeats` CLI option -/
set_option maxHeartbeats 1000000

/- You can set the `maxRecDepth` value with the `-max-recdepth` CLI option -/
set_option maxRecDepth 2048

namespace Aegis.SortArray

/-- [sort_array::sort_array]: loop body 0:
    Source: 'sort_array/src/lib.rs', lines 5:4-8:5
    Visibility: public -/
@[rust_loop_body]
def sort_array_loop0.body
  (nums : Slice Std.I64) (n : Std.Usize) (v : alloc.vec.Vec Std.I64)
  (i : Std.Usize) :
  Result (ControlFlow ((alloc.vec.Vec Std.I64) × Std.Usize) (alloc.vec.Vec
    Std.I64))
  := do
  if i < n
  then
    let i1 ← Slice.index_usize nums i
    let v1 ← alloc.vec.Vec.push v i1
    let i2 ← i + 1#usize
    ok (cont (v1, i2))
  else ok (done v)

/-- [sort_array::sort_array]: loop 0:
    Source: 'sort_array/src/lib.rs', lines 5:4-8:5
    Visibility: public -/
@[rust_loop]
def sort_array_loop0
  (nums : Slice Std.I64) (n : Std.Usize) (v : alloc.vec.Vec Std.I64)
  (i : Std.Usize) :
  Result (alloc.vec.Vec Std.I64)
  := do
  loop
    (fun (v1, i1) => sort_array_loop0.body nums n v1 i1)
    (v, i)

/-- [sort_array::sort_array]: loop body 1:
    Source: 'sort_array/src/lib.rs', lines 14:4-17:5
    Visibility: public -/
@[rust_loop_body]
def sort_array_loop1.body
  (n : Std.Usize) (buf : alloc.vec.Vec Std.I64) (j : Std.Usize) :
  Result (ControlFlow ((alloc.vec.Vec Std.I64) × Std.Usize) (alloc.vec.Vec
    Std.I64))
  := do
  if j < n
  then
    let buf1 ← alloc.vec.Vec.push buf 0#i64
    let j1 ← j + 1#usize
    ok (cont (buf1, j1))
  else ok (done buf)

/-- [sort_array::sort_array]: loop 1:
    Source: 'sort_array/src/lib.rs', lines 14:4-17:5
    Visibility: public -/
@[rust_loop]
def sort_array_loop1
  (n : Std.Usize) (buf : alloc.vec.Vec Std.I64) (j : Std.Usize) :
  Result (alloc.vec.Vec Std.I64)
  := do
  loop
    (fun (buf1, j1) => sort_array_loop1.body n buf1 j1)
    (buf, j)

/-- [sort_array::sort_array]: loop body 4:
    Source: 'sort_array/src/lib.rs', lines 1:0-40:13
    Visibility: public -/
@[rust_loop_body]
def sort_array_loop2_loop0_loop0.body
  (v : alloc.vec.Vec Std.I64) (mid : Std.Usize) (right : Std.Usize)
  (buf : alloc.vec.Vec Std.I64) (a : Std.Usize) (b : Std.Usize) (k : Std.Usize)
  :
  Result (ControlFlow ((alloc.vec.Vec Std.I64) × Std.Usize × Std.Usize ×
    Std.Usize) ((alloc.vec.Vec Std.I64) × Std.Usize × Std.Usize ×
    Std.Usize))
  := do
  if a < mid
  then
    if b < right
    then
      let i ←
        alloc.vec.Vec.index (core.slice.index.SliceIndexUsizeSlice Std.I64) v a
      let i1 ←
        alloc.vec.Vec.index (core.slice.index.SliceIndexUsizeSlice Std.I64) v b
      let (buf1, a1, b1) ←
        if i <= i1
        then
          do
          let (_, index_mut_back) ←
            alloc.vec.Vec.index_mut (core.slice.index.SliceIndexUsizeSlice
              Std.I64) buf k
          let a2 ← a + 1#usize
          let buf2 := index_mut_back i
          ok (buf2, a2, b)
        else
          do
          let (_, index_mut_back) ←
            alloc.vec.Vec.index_mut (core.slice.index.SliceIndexUsizeSlice
              Std.I64) buf k
          let b2 ← b + 1#usize
          let buf2 := index_mut_back i1
          ok (buf2, a, b2)
      let k1 ← k + 1#usize
      ok (cont (buf1, a1, b1, k1))
    else ok (done (buf, a, b, k))
  else ok (done (buf, a, b, k))

/-- [sort_array::sort_array]: loop 4:
    Source: 'sort_array/src/lib.rs', lines 1:0-40:13
    Visibility: public -/
@[rust_loop]
def sort_array_loop2_loop0_loop0
  (v : alloc.vec.Vec Std.I64) (buf : alloc.vec.Vec Std.I64) (mid : Std.Usize)
  (right : Std.Usize) (a : Std.Usize) (b : Std.Usize) (k : Std.Usize) :
  Result ((alloc.vec.Vec Std.I64) × Std.Usize × Std.Usize × Std.Usize)
  := do
  loop
    (fun (buf1, a1, b1, k1) => sort_array_loop2_loop0_loop0.body v mid right
      buf1 a1 b1 k1)
    (buf, a, b, k)

/-- [sort_array::sort_array]: loop body 5:
    Source: 'sort_array/src/lib.rs', lines 41:12-45:13
    Visibility: public -/
@[rust_loop_body]
def sort_array_loop2_loop0_loop1.body
  (v : alloc.vec.Vec Std.I64) (mid : Std.Usize) (buf : alloc.vec.Vec Std.I64)
  (a : Std.Usize) (k : Std.Usize) :
  Result (ControlFlow ((alloc.vec.Vec Std.I64) × Std.Usize × Std.Usize)
    ((alloc.vec.Vec Std.I64) × Std.Usize))
  := do
  if a < mid
  then
    let i ←
      alloc.vec.Vec.index (core.slice.index.SliceIndexUsizeSlice Std.I64) v a
    let (_, index_mut_back) ←
      alloc.vec.Vec.index_mut (core.slice.index.SliceIndexUsizeSlice Std.I64)
        buf k
    let a1 ← a + 1#usize
    let k1 ← k + 1#usize
    let buf1 := index_mut_back i
    ok (cont (buf1, a1, k1))
  else ok (done (buf, k))

/-- [sort_array::sort_array]: loop 5:
    Source: 'sort_array/src/lib.rs', lines 41:12-45:13
    Visibility: public -/
@[rust_loop]
def sort_array_loop2_loop0_loop1
  (v : alloc.vec.Vec Std.I64) (buf : alloc.vec.Vec Std.I64) (mid : Std.Usize)
  (a : Std.Usize) (k : Std.Usize) :
  Result ((alloc.vec.Vec Std.I64) × Std.Usize)
  := do
  loop
    (fun (buf1, a1, k1) => sort_array_loop2_loop0_loop1.body v mid buf1 a1 k1)
    (buf, a, k)

/-- [sort_array::sort_array]: loop body 6:
    Source: 'sort_array/src/lib.rs', lines 46:12-50:13
    Visibility: public -/
@[rust_loop_body]
def sort_array_loop2_loop0_loop2.body
  (v : alloc.vec.Vec Std.I64) (right : Std.Usize) (buf : alloc.vec.Vec Std.I64)
  (b : Std.Usize) (k : Std.Usize) :
  Result (ControlFlow ((alloc.vec.Vec Std.I64) × Std.Usize × Std.Usize)
    (alloc.vec.Vec Std.I64))
  := do
  if b < right
  then
    let i ←
      alloc.vec.Vec.index (core.slice.index.SliceIndexUsizeSlice Std.I64) v b
    let (_, index_mut_back) ←
      alloc.vec.Vec.index_mut (core.slice.index.SliceIndexUsizeSlice Std.I64)
        buf k
    let b1 ← b + 1#usize
    let k1 ← k + 1#usize
    let buf1 := index_mut_back i
    ok (cont (buf1, b1, k1))
  else ok (done buf)

/-- [sort_array::sort_array]: loop 6:
    Source: 'sort_array/src/lib.rs', lines 46:12-50:13
    Visibility: public -/
@[rust_loop]
def sort_array_loop2_loop0_loop2
  (v : alloc.vec.Vec Std.I64) (buf : alloc.vec.Vec Std.I64) (right : Std.Usize)
  (b : Std.Usize) (k : Std.Usize) :
  Result (alloc.vec.Vec Std.I64)
  := do
  loop
    (fun (buf1, b1, k1) => sort_array_loop2_loop0_loop2.body v right buf1 b1
      k1)
    (buf, b, k)

/-- [sort_array::sort_array]: loop body 3:
    Source: 'sort_array/src/lib.rs', lines 1:0-53:9
    Visibility: public -/
@[rust_loop_body]
def sort_array_loop2_loop0.body
  (n : Std.Usize) (v : alloc.vec.Vec Std.I64) (width : Std.Usize)
  (buf : alloc.vec.Vec Std.I64) (left : Std.Usize) :
  Result (ControlFlow ((alloc.vec.Vec Std.I64) × Std.Usize) (alloc.vec.Vec
    Std.I64))
  := do
  if left < n
  then
    let i ← left + width
    let b ← if i < n
              then ok i
              else ok n
    let i1 ← 2#usize * width
    let i2 ← left + i1
    let right ← if i2 < n
                  then left + i1
                  else ok n
    let (buf1, a, b1, k) ←
      sort_array_loop2_loop0_loop0 v buf b right left b left
    let (buf2, k1) ← sort_array_loop2_loop0_loop1 v buf1 b a k
    let buf3 ← sort_array_loop2_loop0_loop2 v buf2 right b1 k1
    let i3 ← 2#usize * width
    let left1 ← left + i3
    ok (cont (buf3, left1))
  else ok (done buf)

/-- [sort_array::sort_array]: loop 3:
    Source: 'sort_array/src/lib.rs', lines 1:0-53:9
    Visibility: public -/
@[rust_loop]
def sort_array_loop2_loop0
  (n : Std.Usize) (v : alloc.vec.Vec Std.I64) (buf : alloc.vec.Vec Std.I64)
  (width : Std.Usize) (left : Std.Usize) :
  Result (alloc.vec.Vec Std.I64)
  := do
  loop
    (fun (buf1, left1) => sort_array_loop2_loop0.body n v width buf1 left1)
    (buf, left)

/-- [sort_array::sort_array]: loop body 7:
    Source: 'sort_array/src/lib.rs', lines 57:8-60:9
    Visibility: public -/
@[rust_loop_body]
def sort_array_loop2_loop1.body
  (n : Std.Usize) (buf : alloc.vec.Vec Std.I64) (v : alloc.vec.Vec Std.I64)
  (c : Std.Usize) :
  Result (ControlFlow ((alloc.vec.Vec Std.I64) × Std.Usize) (alloc.vec.Vec
    Std.I64))
  := do
  if c < n
  then
    let i ←
      alloc.vec.Vec.index (core.slice.index.SliceIndexUsizeSlice Std.I64) buf c
    let (_, index_mut_back) ←
      alloc.vec.Vec.index_mut (core.slice.index.SliceIndexUsizeSlice Std.I64) v
        c
    let c1 ← c + 1#usize
    let v1 := index_mut_back i
    ok (cont (v1, c1))
  else ok (done v)

/-- [sort_array::sort_array]: loop 7:
    Source: 'sort_array/src/lib.rs', lines 57:8-60:9
    Visibility: public -/
@[rust_loop]
def sort_array_loop2_loop1
  (n : Std.Usize) (v : alloc.vec.Vec Std.I64) (buf : alloc.vec.Vec Std.I64)
  (c : Std.Usize) :
  Result (alloc.vec.Vec Std.I64)
  := do
  loop
    (fun (v1, c1) => sort_array_loop2_loop1.body n buf v1 c1)
    (v, c)

/-- [sort_array::sort_array]: loop body 2:
    Source: 'sort_array/src/lib.rs', lines 1:0-63:5
    Visibility: public -/
@[rust_loop_body]
def sort_array_loop2.body
  (n : Std.Usize) (v : alloc.vec.Vec Std.I64) (buf : alloc.vec.Vec Std.I64)
  (width : Std.Usize) :
  Result (ControlFlow ((alloc.vec.Vec Std.I64) × (alloc.vec.Vec Std.I64) ×
    Std.Usize) (alloc.vec.Vec Std.I64))
  := do
  if width < n
  then
    let buf1 ← sort_array_loop2_loop0 n v buf width 0#usize
    let v1 ← sort_array_loop2_loop1 n v buf1 0#usize
    let width1 ← width * 2#usize
    ok (cont (v1, buf1, width1))
  else ok (done v)

/-- [sort_array::sort_array]: loop 2:
    Source: 'sort_array/src/lib.rs', lines 1:0-63:5
    Visibility: public -/
@[rust_loop]
def sort_array_loop2
  (n : Std.Usize) (v : alloc.vec.Vec Std.I64) (buf : alloc.vec.Vec Std.I64)
  (width : Std.Usize) :
  Result (alloc.vec.Vec Std.I64)
  := do
  loop
    (fun (v1, buf1, width1) => sort_array_loop2.body n v1 buf1 width1)
    (v, buf, width)

/-- [sort_array::sort_array]:
    Source: 'sort_array/src/lib.rs', lines 1:0-66:1
    Visibility: public -/
def sort_array (nums : Slice Std.I64) : Result (alloc.vec.Vec Std.I64) := do
  let n := Slice.len nums
  let v ← sort_array_loop0 nums n (alloc.vec.Vec.new Std.I64) 0#usize
  if n <= 1#usize
  then ok v
  else
    let buf ← sort_array_loop1 n (alloc.vec.Vec.new Std.I64) 0#usize
    sort_array_loop2 n v buf 1#usize

end Aegis.SortArray
\end{lstlisting}

\subsection{Unresolved theorem artifact}

The listing below is the frozen theorem as the prover left it, with the
original \lstinline{sorry} proof body.

\begin{lstlisting}[style=code]
import Aegis.SortArray.Code.Funs
open Aeneas Aeneas.Std Result

/-- Totality specification for `Aegis.SortArray.sort_array`, read purely from the structure of
the extracted Lean body and deliberately ignoring what the function is meant to compute. The function
takes a slice `nums` of 64-bit signed integers. It first reads the length `n = nums.length` and runs
`sort_array_loop0`, which walks an index `i` from `0` up to `n`, reading `nums[i]` and pushing each
value onto a fresh vector `v` (advancing `i` by one each step). If `n ≤ 1` it returns `v` immediately.
Otherwise it builds a scratch buffer `buf` of `n` zeros via `sort_array_loop1`, then runs the
bottom-up merge-sort driver `sort_array_loop2`: an outer loop doubles a `width` (`1, 2, 4, …`) while
`width < n`, and for each width it merges adjacent runs by running three inner loops
(`sort_array_loop2_loop0_loop{0,1,2}`) that read from `v` and write into `buf` at guarded indices,
then copies `buf` back into `v` (`sort_array_loop2_loop1`).

The only ways this computation can FAULT are: (a) an out-of-bounds slice/vector read or indexed write
— every read `nums[i]`, `v[a]`, `v[b]`, `buf[c]` and every indexed write into `buf`/`v` is performed
only inside a loop step guarded by the corresponding index being strictly below the relevant length,
so all accesses are in bounds; (b) an unsigned `Usize` overflow on an index computation — the driver
forms `left + width`, `2 * width`, `left + 2*width`, and the increments `i + 1`, `a + 1`, `b + 1`,
`k + 1`, `j + 1`, `c + 1`, `width * 2`; every one of these quantities is an array index or a width
that stays bounded by `n` (indices never exceed `n`, and `width` is only doubled while `width < n`,
so it and `left + 2*width` stay within a small multiple of `n`), which under the length bound below
is astronomically smaller than `Usize.max = 2^64 − 1`, so no unsigned overflow can occur; (c) a signed
`Std.I64` overflow — there is NO signed arithmetic on the array values at all: the `Std.I64` entries
are only copied, read, compared (`i ≤ i1`), and written, never added, subtracted, multiplied, or
divided, so no signed 64-bit overflow is possible. The function can DIVERGE only if some loop fails to
terminate; each loop has a strictly monotone index (`i`, `a`, `b`, `k`, `j`, `c`, or `left`) advancing
toward a fixed finite bound (`n`, `mid`, or `right`), and the outer merge-sort loop terminates because
`width` strictly increases by doubling while bounded by `n`, so every loop runs to completion.

We adopt the reasonable, explicitly stated precondition — matching the problem's own stated constraint
`1 <= nums.length <= 5 * 10^4` — that `nums` is NON-EMPTY and has at most `50000` entries. Under this
precondition every read and write is in bounds, every `Usize` index computation stays far below
`Usize.max`, no signed arithmetic occurs at all, and every loop terminates. The claim is therefore a
pure totality statement: the call `sort_array nums` RUNS TO COMPLETION SUCCESSFULLY, returning an `ok`
result rather than faulting or diverging. Nothing is asserted about WHICH list of integers comes
back. -/
theorem Aegis.SortArray.sort_array_returns_ok
    (nums : Slice Std.I64)
    (hlen_lo : 1 ≤ nums.length)
    (hlen_hi : nums.length ≤ 50000) :
    Aegis.SortArray.sort_array nums ⦃ r => True ⦄ := by
  sorry
\end{lstlisting}

\clearpage
\section{Recurring Proof Obligations}
\label{app:cross-case-lessons}

The case studies share several kinds of obligation that do not arise in a purely
mathematical description of the algorithms.

\paragraph{Weak postconditions.}
The even/odd sorting theorem constrains only the output length, but its
weakest-precondition judgment also requires every preceding sort, lookup,
update, and increment to succeed.

\paragraph{Input bounds.}
Input bounds justify casts, additions, multiplications, pushes, and
indexed accesses in the translated program.  The subset-XOR proof uses the
twelve-element bound to control repeated machine-word doubling; the reshape
proof uses dimension bounds to justify signed products and signed-to-unsigned
casts.  The cell-value bounds in the reshape and diagonal statements are not
used, because neither program performs arithmetic on cell values.

\paragraph{Mutation.}
In-place Rust assignments become forward reads and backward functions that
rebuild vectors, so the proofs repeatedly establish length preservation and
lookup behavior after each update.  In the sorting programs, a single
source-level swap becomes several monadic update steps.

\paragraph{Relational theorems.}
Reversal invariance and uniform-shift invariance have short mathematical
arguments, but their theorems relate separate executions.  The proofs connect
each execution to a common model or synchronize them at loop boundaries while
carrying representation and safety invariants.

\paragraph{Totality.}
The unresolved merge-sort theorem does not ask for sortedness.  It asks only
that no generated failure branch is reached and that every loop terminates.
This holds trivially for the mathematical sorting function.  The translated
fixed-width program instead requires the coupled cursor and overflow invariants
listed in \cref{app:merge-sort-open}.

Across the six resolved cases, the Rust sources range from 19 to 67 lines,
while the accepted proofs range from 289 to 1,167 lines
(\cref{tab:appendix-case-summary}).  Most of that length connects the
mathematical claim to the concrete representation and to the safe execution of
the generated control flow.  In the triangular-sum and diagonal cases, the
target theorem and its proof occupy only the last 25 and 38 lines.

\end{document}